\documentclass[final]{cvpr}

\def\cvprPaperID{8742}
\def\confYear{CVPR 2021}

\usepackage{times}
\usepackage{float}
\usepackage{graphicx}
\graphicspath{ {images/} {data/} }
\usepackage{amsmath}
\usepackage{amssymb}
\usepackage{enumitem}
\usepackage{multirow}
\usepackage[export]{adjustbox}

\usepackage{pifont}

\usepackage{booktabs}

\usepackage{makecell}

\newcolumntype{H}{>{\setbox0=\hbox\bgroup}c<{\egroup}@{}}
\newcommand*{\statstabh}[2]{\rotatebox{0}{\makecell[c]{\textbf{#1}\\\textbf{#2}}}}

\usepackage{microtype}

\usepackage{comment}

\usepackage[symbol]{footmisc}

\usepackage[pagebackref=true,breaklinks=true,colorlinks,bookmarks=false]{hyperref}

\begin{document}

\title{Part-aware Panoptic Segmentation}

\author{Daan de Geus\textsuperscript{1}\footnotemark[1] \quad Panagiotis Meletis\textsuperscript{1}\footnotemark[1] \quad Chenyang Lu\textsuperscript{1} \quad Xiaoxiao Wen\textsuperscript{2} \quad Gijs Dubbelman\textsuperscript{1}\\
\textsuperscript{1}Eindhoven University of Technology \quad \textsuperscript{2}University of Amsterdam\\
{\{\tt\small d.c.d.geus}, {\tt\small p.c.meletis}\} {\tt\small@tue.nl}
}

\maketitle
\thispagestyle{empty}
\pagestyle{empty}

\footnotetext[1]{Both authors contributed equally.}

\begin{abstract}
In this work, we introduce the new scene understanding task of Part-aware Panoptic Segmentation (PPS), which aims to understand a scene at multiple levels of abstraction, and unifies the tasks of scene parsing and part parsing. For this novel task, we provide consistent annotations on two commonly used datasets: Cityscapes and Pascal VOC. Moreover, we present a single metric to evaluate PPS, called Part-aware Panoptic Quality (PartPQ). For this new task, using the metric and annotations, we set multiple baselines by merging results of existing state-of-the-art methods for panoptic segmentation and part segmentation. Finally, we conduct several experiments that evaluate the importance of the different levels of abstraction in this single task.
\end{abstract}

\vspace{-2pt}

\section{Introduction}

Humans perceive and understand a scene at multiple levels of abstraction. Concretely, when observing a scene, we do not only see a single semantic label for each visual entity, such as \textit{person} or \textit{car}. We also distinguish the parts of entities, such as \textit{person-leg} and \textit{car-wheel}, and we are able to group together the parts that belong to a single individual entity. Currently, there is no computer vision task that aims at simultaneously understanding a scene holistically on both of these levels of abstraction: \textit{scene parsing} and \textit{part parsing}.

Instead, most methods focus on solving a task at a single level of abstraction. On the one hand, scene parsing aims to recognize and semantically segment all foreground objects (\textit{things}) and background classes (\textit{stuff}) in an image. Recently, this task has been formalized as \textit{panoptic segmentation}~\cite{Kirillov2019PS}, for which the goal is to predict 1) a class label and 2) an instance \textit{id} for each pixel in an image. This formalization has resulted in a boost in research interest that advanced the state-of-the-art~\cite{Cheng2020PanopticDeepLab, Kirillov2019PanopticFPN, Li2020Unifying, mohan2020efficientps, porzi2019seamless, xiong2019upsnet}. On the other hand, part parsing takes over where scene parsing stops, as it aims to segment an image based on part-level semantics, \ie, the parts constituting the scene-level classes. For this level of abstraction, there is a wide range of different task definitions, and resulting methods. Most methods focus on a single object class and are instance-agnostic, while only a few are instance-aware~\cite{gong2018instance, li2017holistic, zhao2018understanding}, or focus on multiple object classes \cite{michieli2020gmnet, Zhao2019BSANet}. A more comprehensive overview of related work is provided in Section \ref{sec:related_work}. 

\begin{figure}
\centering
\begin{adjustbox}{width=1.\linewidth}
\begin{tabular}{c@{\extracolsep{2pt}}c}
\centering

\includegraphics[width=0.5\linewidth]{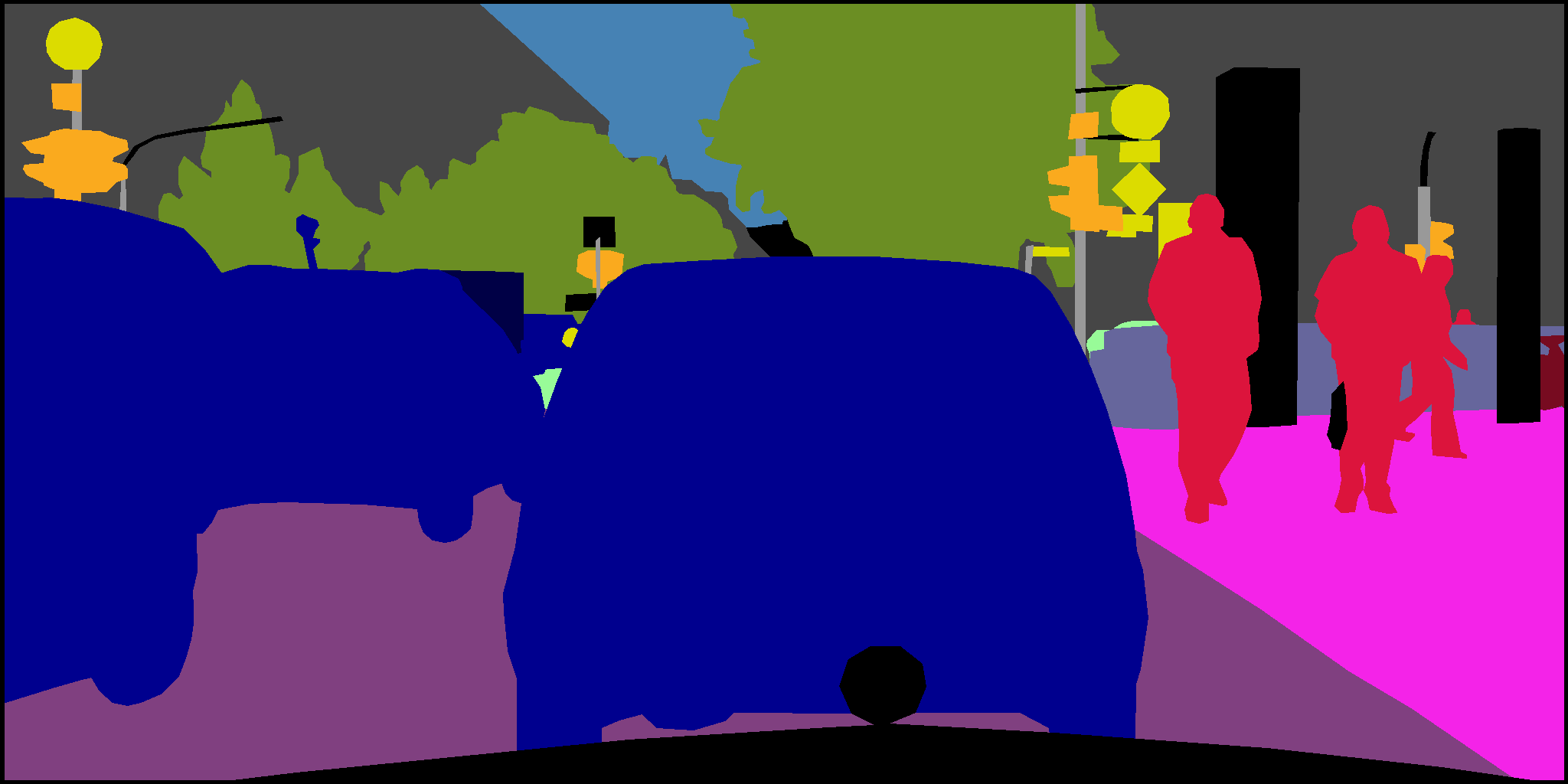} &
\includegraphics[width=0.5\linewidth]{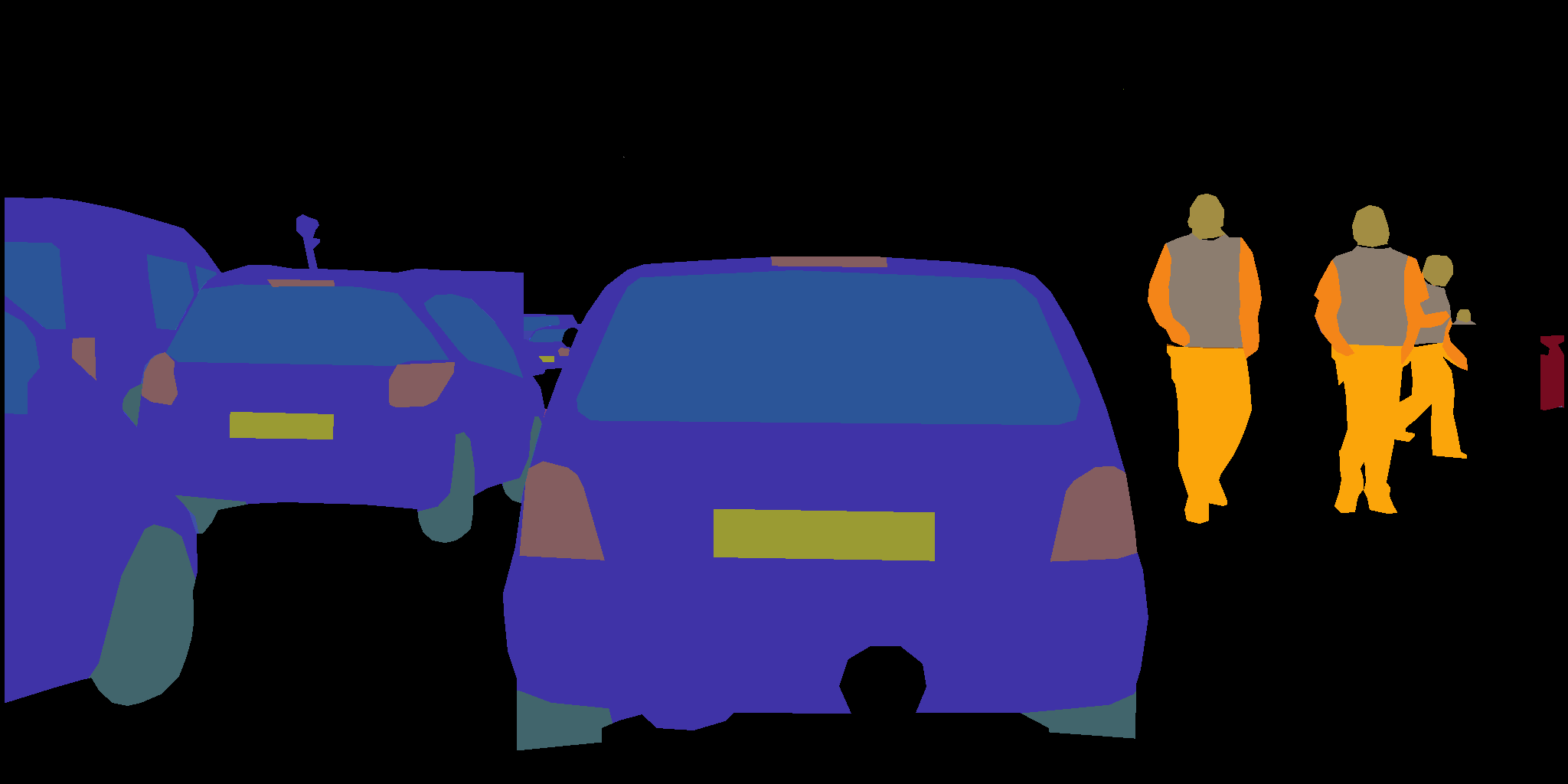} \vspace{-3pt}\\

{\footnotesize Semantic Segmentation} & {\footnotesize Part Segmentation} \\

\includegraphics[width=0.5\linewidth]{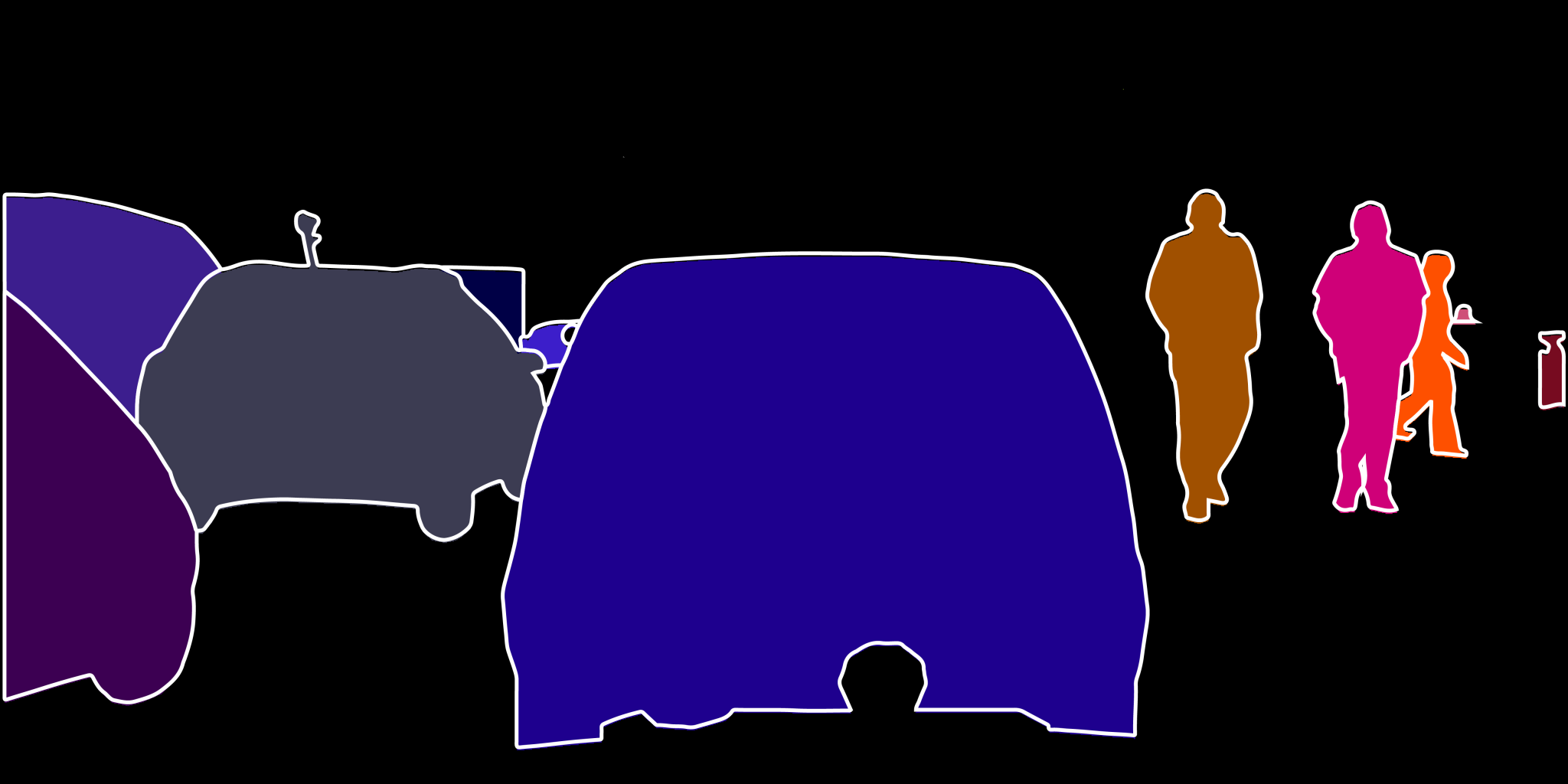} &
\includegraphics[width=0.5\linewidth]{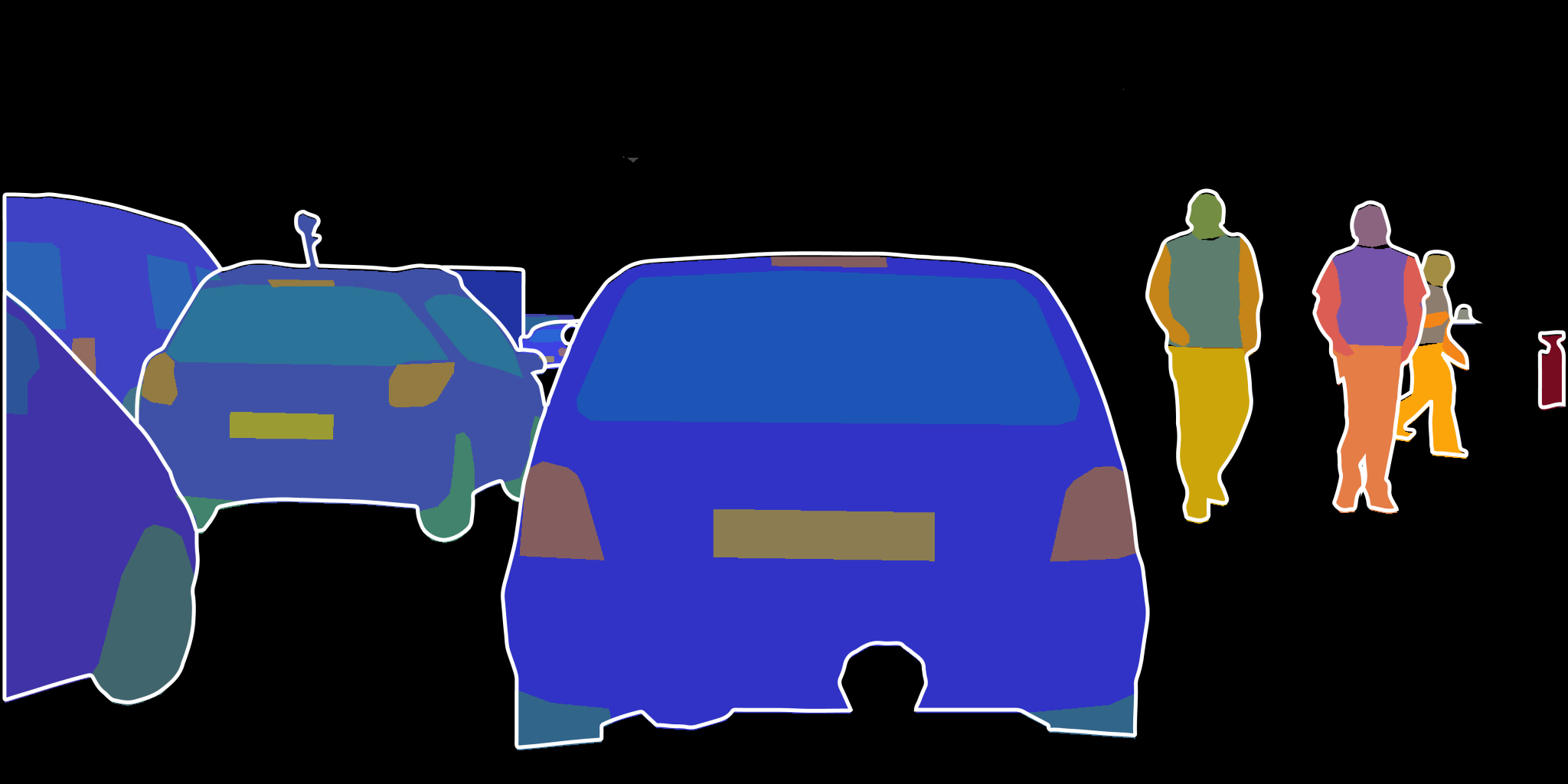} \vspace{-3pt} \\

{\footnotesize Instance Segmentation} & {\footnotesize Instance-aware Part Segmentation} \\

\includegraphics[width=0.5\linewidth]{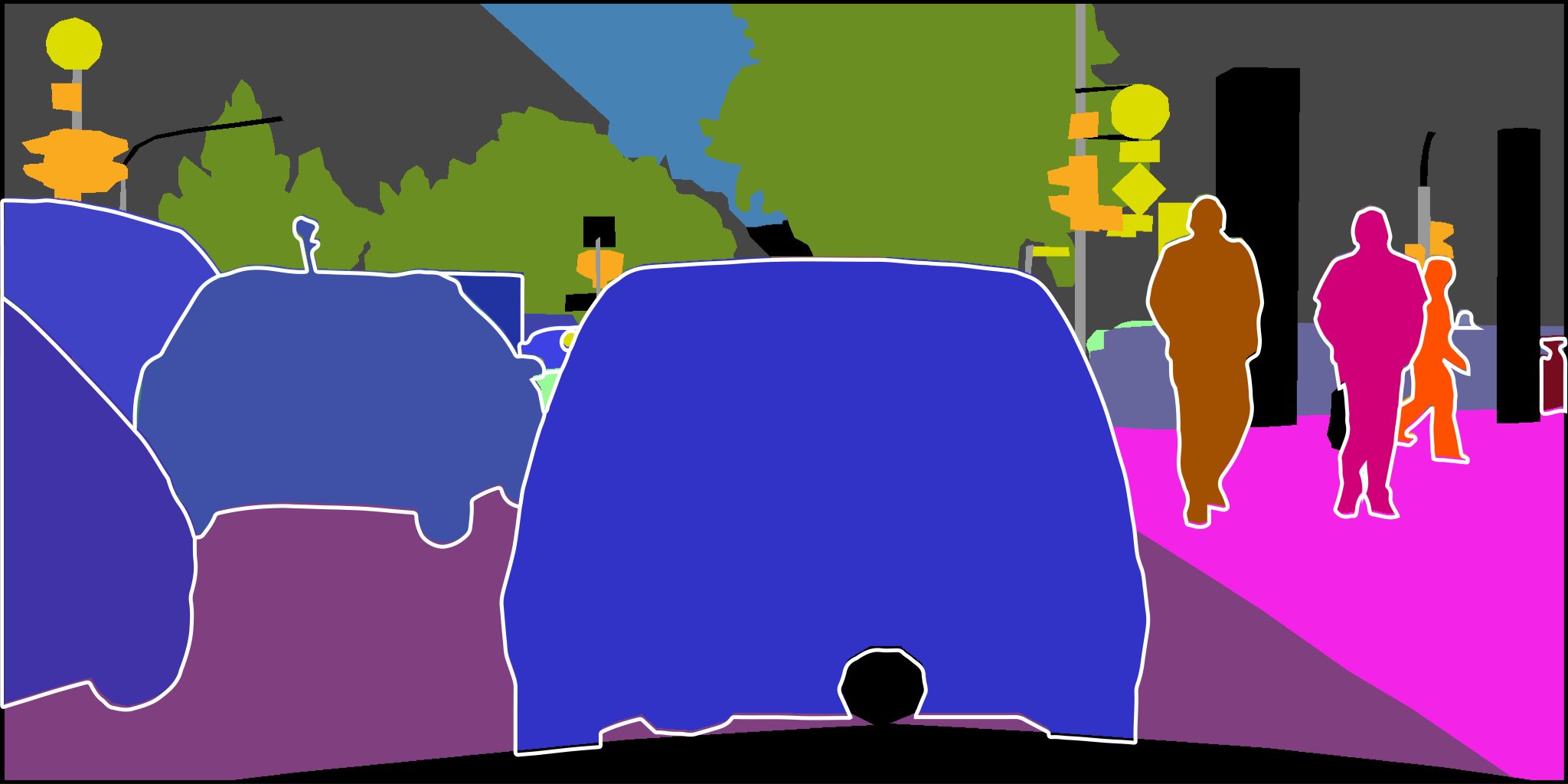} &
\includegraphics[width=0.5\linewidth]{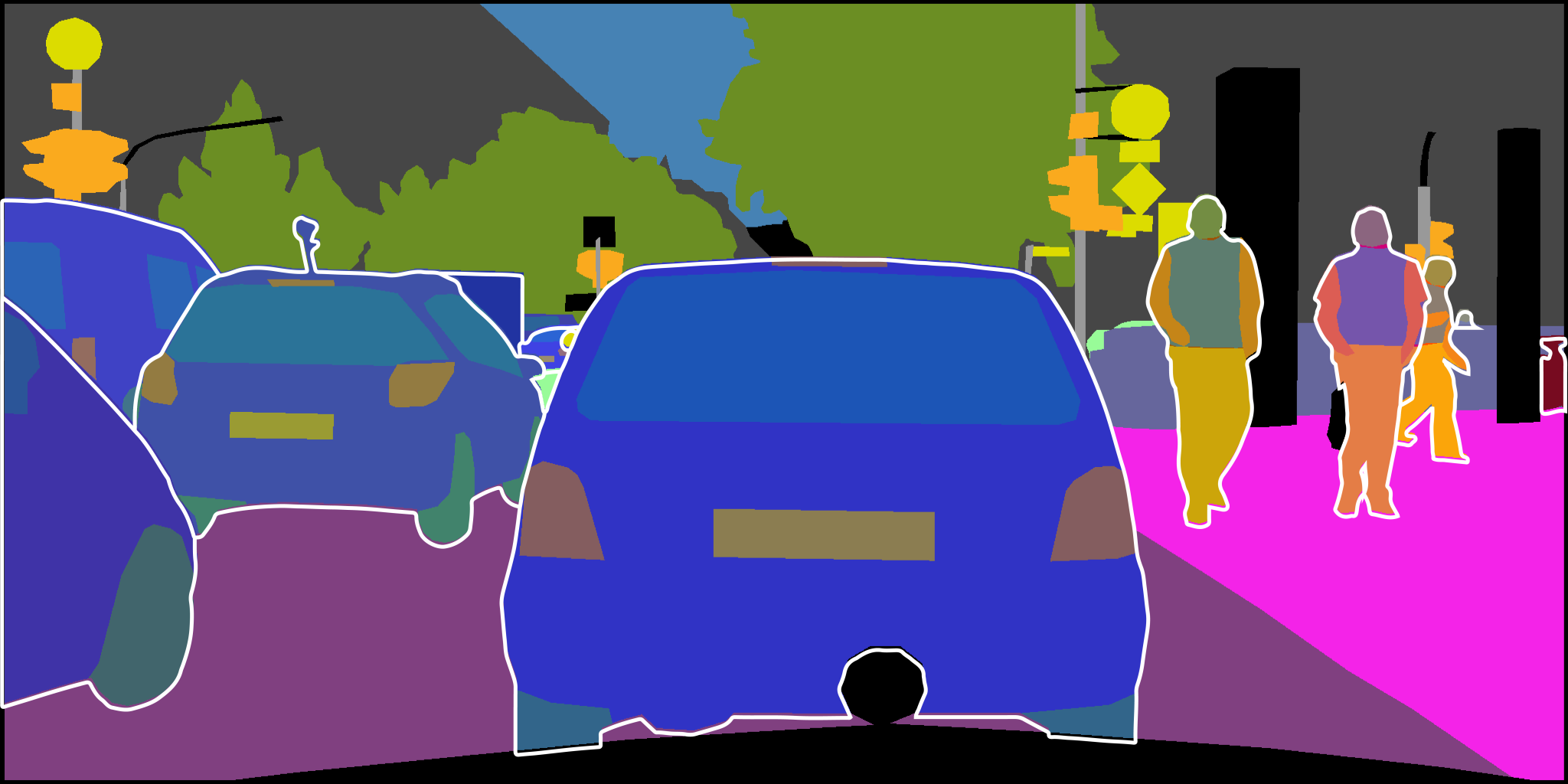} \vspace{-3pt} \\
{\footnotesize{Panoptic Segmentation}} & {\footnotesize{\textbf{Part-aware Panoptic Segmentation}}} \\
\end{tabular}
\end{adjustbox}

\vspace{3pt}
\caption{\textbf{Evolution of scene understanding tasks}: from semantic to panoptic (top to bottom) and from part-agnostic to part-aware (left to right). Colors indicate scene-level and part-level semantics. Instance-level boundaries are emphasized with a white contour.\vspace{-10pt}}
\label{fig:task_definition}
\end{figure}

To come closer to unified perception at multiple levels of abstraction, this work defines a task that combines scene parsing and part parsing in a single task. This task encompasses the ability to 1) apply per-pixel scene-level classification, 2) segment things classes into individual instances, and 3)~segment stuff classes or things instances into their respective parts. We call this task \textit{part-aware panoptic segmentation} (PPS); the conceptual differences with existing tasks are visualized in Figure \ref{fig:task_definition}. Together with this task, we also define a metric to evaluate it. This metric, \textit{part-aware panoptic quality} (PartPQ), extends the \textit{panoptic quality} metric \cite{Kirillov2019PS} to cover part segmentation performance per detected things instance or stuff class. More details on the task and metric can be found in Section~\ref{sec:task_definition}.

To allow for research on the new task of PPS, we introduce consistent part-aware panoptic annotations for two commonly used datasets. For Cityscapes \cite{Cordts2016Cityscapes}, we have labeled part classes for all 3.5k images of the train and validation set, which are consistent with the existing panoptic annotations. For Pascal VOC \cite{Everingham2010Pascal}, we have combined the existing datasets for semantic segmentation~\cite{mottaghi14pascalcontext} and instance-aware part segmentation~\cite{chen2014detect} to generate a consistent annotation set for PPS. In Section \ref{sec:datasets}, we provide further explanations and statistics on these datasets.

As there is no existing work on part-aware panoptic segmentation, we establish several benchmarks. We create baselines by generating state-of-the-art results on panoptic segmentation and part segmentation, and merging these to the PPS format using heuristics. As explained in Section \ref{sec:experiments}, there are several design choices that need to be taken into account, when combining predictions at multiple levels of abstraction. Specifically: should we opt for a top-down method, where we prioritize the scene-level predictions from panoptic segmentation, and complement these with part predictions, or is it better to use a bottom-up approach, where we combine parts to generate scene-level predictions? To evaluate this, we conduct experiments to research the benefits of both types of strategies. Both these experiments and the baselines provide a direction for future research on multi-task training of PPS architectures where the different subtasks can benefit from each other.

To summarize, this work contains the following contributions:
\begin{itemize}[noitemsep,topsep=0pt]
\item The introduction of the part-aware panoptic segmentation (PPS) task, unifying perception at multiple levels of abstraction.
\item The PartPQ metric to evaluate this task.
\item Coherent PPS annotations for two commonly used datasets, which are made available to the public.
\item Baselines for the PPS task on two datasets.
\item An analysis of the design choices for the new PPS task.
\end{itemize}
\vspace{2pt}
\noindent All annotations and the code are available at \url{https://github.com/tue-mps/panoptic_parts}.

\section{Related work}
\label{sec:related_work}
Research on visual scene understanding aims to extract all-encompassing information from images with the long-term goal to mimic human visual-cognitive capabilities. So far, research has primarily focused on approaching scene understanding at a single level of abstraction. In this work, we propose a single coherent task for multiple levels of abstraction, which unifies the tasks of scene parsing and part parsing. 

\subsection{Scene parsing}
We refer to scene parsing as the overall task to semantically understand an image at the class level, and to distinguish between individual things instances. Recently, this task has been formalized as panoptic segmentation \cite{Kirillov2019PS}, which is a unification of the typically distinct tasks of semantic segmentation and instance segmentation. In earlier forms, this task has been investigated in~\cite{tu2005image, yao2012describing}. 

Initially, most panoptic segmentation methods applied a multi-task network that trains and outputs instance segmentation and semantic segmentation in parallel, followed by a merging operation to generate panoptic segmentation results~\cite{degeus2019single, Kirillov2019PanopticFPN, li2019attention, mohan2020efficientps, porzi2019seamless}. Recently, more methods are introduced that focus on optimizing the process of merging to panoptic segmentation~\cite{Lazarow2020OCFusion, liu2019end, xiong2019upsnet, Yang2020SOGNet}, or try to solve the task more holistically or efficiently~\cite{Cheng2020PanopticDeepLab, degeus2020fpsnet, Hou2020Real, Li2020Unifying, yang2019deeperlab}.

Although panoptic segmentation allows for more holistic scene understanding than the earlier tasks of semantic segmentation and instance segmentation, it does not cover knowledge of part-level semantics of the identified segments. Such knowledge would provide a more comprehensive understanding of the scene, and would allow for more detailed downstream reasoning.

\subsection{Part parsing}
We refer to part parsing as the umbrella task of segmenting images based on part-level semantics. At a high level, we can distinguish two types of tasks: part segmentation and pose estimation. Part segmentation requires a pixel-level prediction for all identified parts, whereas pose estimation aims at detecting connecting keypoints between the parts for each object. Pose estimation is inherently instance-aware and is exhaustively researched, as is clear from the surveys in~\cite{dang2019deep, liu2015survey}. 

However, dense part-level segmentation remained for a long time instance-agnostic, as it is usually treated as a semantic segmentation problem ~\cite{gong2019graphonomy, jiang2018cnn, jiang2019cnn, li2017holistic, liang2015human, liu2018cross, luo2013pedestrian, luo2018trusted, michieli2020gmnet, Zhao2019BSANet}.  In the trend of coming to more holistic tasks, a dense pose task was introduced in~\cite{alp2018densepose} and a unification of pose estimation and part segmentation is provided in~\cite{dong2014towards}. Only recently, to the best of our knowledge, an instance-aware human part segmentation task was introduced and studied in~\cite{gong2018instance, li2017holistic, zhao2018understanding}. Most research has focused on part segmentation for humans~\cite{dong2013deformable, gong2018instance, ladicky2013human, li2018multi, li2019self, liang2016semantic, liang2018look, lin2019cross, ruan2019devil, yang2019parsing, zhao2018understanding}, but other parts have also received attention, \eg, facial parts~\cite{lin2019face}, and animal parts~\cite{chen2014detect, wang2015joint}. A limited amount of papers have addressed multi-class part segmentation~\cite{michieli2020gmnet, Zhao2019BSANet}, but so far these methods are not instance-aware. As a result, instance-aware part segmentation on a more general dataset, consisting of a wider range of classes and parts, remains unaddressed. 

Moreover, although work has shown that learning on multiple levels of abstraction can improve the performance of a part segmentation network~\cite{michieli2020gmnet, Zhao2019BSANet}, part parsing has not yet been merged with scene parsing into one holistic task, which can describe the image at multiple levels of abstraction. In our work, we aim to boost the interest in this area by providing a unified task for \textit{part-aware panoptic segmentation}, and accompanying metrics and annotations.

\subsection{Datasets}
In order to train and evaluate on the new PPS task, we need datasets that 1) have scene-level labels for panoptic segmentation and 2) have part-level labels for a set of those scene-level classes. Although a plethora of datasets exist for object detection and semantic segmentation, only few have labels compatible with the panoptic segmentation task (\eg, \cite{Cordts2016Cityscapes, Lin2014COCO}). For part-level segmentation, the datasets are even more scarce. LIP~\cite{liang2018look}, MHP~\cite{zhao2018understanding} and CIHP~\cite{gong2018instance} provide instance-aware, part-level annotations, but only for human parts. To the best of our knowledge, Pascal-Parts is the only dataset that has part-level annotations for a more general set of classes~\cite{chen2014detect}. However, these annotations do not contain any information on classes without parts. 

From this, we observe that there is no dataset that covers all the requirements for the PPS task. Therefore, we present consistent part-aware panoptic annotations on two datasets. For Cityscapes~\cite{Cordts2016Cityscapes}, a commonly used dataset for panoptic segmentation, we annotate parts for five different things classes. Moreover, we collect and arrange the different annotation sets for Pascal VOC~\cite{Everingham2010Pascal} to generate a complete and consistent annotation set for 10k Pascal VOC images.

\section{Part-aware Panoptic Segmentation}
\label{sec:task_definition}

\subsection{Task definition}

The task of \textit{Part-aware Panoptic Segmentation} (PPS) is an image understanding task that is designed to capture image understanding at multiple levels of abstraction. Specifically, it captures 1) scene-level semantics, 2) instance-level information, and 3) part-level semantics. To achieve this, we define PPS as a task that enriches panoptic segmentation \cite{Kirillov2019PS} with part-level semantics. 

A part-aware panoptic segmentation algorithm describes every pixel in an image with a set of semantic and instance-level information. This can be expressed for pixel $i$ in the form $(l, p, z)_i$, where $l$ represents the scene-level semantic class, $p$ the part-level semantic class, and $z \in \mathbb{N}$ the instance \textit{id}. The scene- and part-level semantic classes are predefined and usually correspond to the available semantic granularity of a dataset's labels, while the instance \textit{id} is an unbounded integer separating, per image, distinct instances of the same scene-level semantic class.

The scene-level semantic class $l$ is chosen from a predetermined set of $\mathcal{L} := \{l_1,\dots,l_{L}\}$ classes. For any of these classes a set of part-level semantic classes $\mathcal{P}_l = \{p_{l,1},\dots,p_{l,P_l}\}$ containing $P_l$ semantic parts may be defined. We denote the superset of all parts as $\mathfrak{P} = \cup_l \mathcal{P}_l,~l \in \mathcal{L}$. The set $\mathcal{L}$ can be separated into disjoint subsets in two different ways. \textit{Firstly}, $\mathcal{L} = \mathcal{L}^\text{St} \cup \mathcal{L}^\text{Th}$. The subset $\mathcal{L}^\text{St}$ consists of the stuff classes, \ie, uncountable entities with amorphous shape (\eg, sky, sea), and subset $\mathcal{L}^\text{Th}$ contains the things classes, which are classes for countable objects with well-defined shape (\eg, car, person). 
\textit{Secondly}, $\mathcal{L}$ can also be separated in a subset of scene-level classes that have parts (\eg, limbs, car parts), $\mathcal{L}^\text{parts}$, and scene-level classes that do not have parts, $\mathcal{L}^\text{no-parts}$. Here, $\mathcal{L} = \mathcal{L}^\text{parts} \cup \mathcal{L}^\text{no-parts}$. We require that both $\mathcal{L}^\text{St} \cap \mathcal{L}^\text{Th} = \emptyset$ and $\mathcal{L}^\text{parts} \cap \mathcal{L}^\text{no-parts} = \emptyset$. The selection of classes belonging to the four subsets $\mathcal{L}^\text{St}$, $\mathcal{L}^\text{Th}$, $\mathcal{L}^\text{parts}$, $\mathcal{L}^\text{no-parts}$ is a design choice that is typically determined based on the requirements of the application, or the purpose of a dataset, as for~\cite{Kirillov2019PS}.

A PPS algorithm makes a prediction that adheres to the following requirements: 1) a scene-level semantic class $\mathcal{L}$ must be assigned to all pixels, 2) a part-level semantic class must be assigned to -- and only to -- all pixels that are assigned a scene-level class from $\mathcal{L}^\text{parts}$, and 3) an instance-level \textit{id} is provided for -- and only for -- pixels that are assigned a scene-level class from $\mathcal{L}^\text{Th}$. In summary, a pixel can be labeled with one of following combinations, where ``$-$'' denotes that the specific abstraction level is irrelevant, as a:
\begin{itemize}[noitemsep,topsep=0pt,leftmargin=11pt]
\item Stuff class: $(l, -, -)$, $l \in \mathcal{L}^{St}$
\item Stuff class with parts: $(l, p, -)$, $l \in \mathcal{L}^{St} \cap \mathcal{L}^\text{parts}, p \in \mathcal{P}_l$
\item Things class: $(l, -, z)$, $l \in \mathcal{L}^{Th}, z \in \mathbb{N}$
\item Things class with parts: $(l, p, z)$, $l \in \mathcal{L}^{Th} \cap \mathcal{L}^\text{parts}, p \in \mathcal{P}_l$
\end{itemize}
Finally, the PPS format accepts a special \textit{void} label for scene-level and part-level semantics, which represents ambiguous pixels or concepts not included in any subset $\mathcal{L}$.

\vspace{2pt}
\noindent\textbf{Relationship to other tasks.} \enskip
\textit{Part-aware panoptic segmentation} (PPS) is related to and generalizes various per-pixel segmentation tasks. \textit{Part segmentation} is specialized semantic segmentation focusing on segmenting object parts, but it does not require separating parts according to the object instance they belong to. In the PPS format it can be described as $(l, p, -)_i$, $l \in \mathcal{L}^\text{parts}$, $p \in \mathfrak{P}$. \textit{Instance-aware part segmentation}, can be described as $(l, p, z)_i$, $l \in \mathcal{L}^\text{Th} \cap \mathcal{L}^\text{parts}$, $p \in \mathfrak{P}$, and pivots part parsing on an instance level, but treats any non-things pixel as background, losing environmental context. Finally, \textit{panoptic segmentation}, $(l, -, z)_i$, $l \in \mathcal{L}$, includes no notion of part semantics.

\begin{table*}[t]
    \small
    \setlength\tabcolsep{5.0pt}
	\centering
	\begin{tabular}{@{}lccccccHHccc@{}}
		\toprule
		\textbf{Dataset} & \statstabh{Instance}{aware} & \statstabh{Panoptic}{aware} & \statstabh{Parts}{aware} & \statstabh{Stuff}{classes} & \statstabh{Things}{classes} & \statstabh{Parts}{classes} & \statstabh{Human parts}{} & \statstabh{Vehicle parts}{} & \statstabh{\#Images}{train / val} & \statstabh{Average}{image size} & \statstabh{Average}{\#inst./img}\\
		\midrule
		PASCAL-Context~\cite{mottaghi14pascalcontext} & - & - & - & 459 (59) & - & - & - & - & 4998 / 5105 & 387 $\times$ 470 & -\\
		LIP~\cite{liang2018look} & - & - & \ding{51} & - & 1 & 20 & 20 & - & 30.5k / 10k & 325 $\times$ 240 & -\\
		CIHP~\cite{gong2018instance} & \ding{51} & - & \ding{51} & - & 1 & 20 & 20 & - & 28.3k / 5k & 484 $\times$ 578 & 3.4\\
		MHP v2.0~\cite{zhao2018understanding} & \ding{51} & - & \ding{51} & - & 1 & 59 & 59 & - & 15.4k / 5k & 644 $\times$ 718 & 3\\
		PASCAL-Person-Parts~\cite{chen2014detect} & \ding{51} & - & \ding{51} & - & 1 & 6 & 6 & - & 1716 / 1817 & 387 $\times$ 470 & 2.2\\
		PASCAL-Parts~\cite{chen2014detect} & \ding{51} & - & \ding{51} & - & 20 & 194 & 24 & 57 & 4998 / 5105 & 387 $\times$ 470 & 2.5\\
		Cityscapes~\cite{Cordts2016Cityscapes} & \ding{51} & \ding{51} & - & 23 & 8 & - & - & - & 2975 / 500 & 1024 $\times$ 2048 & 17.9\\
		\midrule
		\textbf{\textit{This work}}\\
		~~\textbf{PASCAL Panoptic Parts} & \ding{51} & \ding{51} & \ding{51} & 80 & 20 & 194 & 24 & 57 & 4998 / 5105 & 387 $\times$ 470 & 2.5\\
		~~\textbf{Cityscapes Panoptic Parts} & \ding{51} & \ding{51} & \ding{51} & 23 & 8 & 23 & 4 & 5 & 2975 / 500 & 1024 $\times$ 2048 & 17.9\\
		\bottomrule
	\end{tabular}
	\vspace{2pt}
	\caption{\textbf{Dataset statistics} for related (part) segmentation datasets and our proposed datasets.
	\textit{PASCAL-Context} has 459 semantic classes but only 59 of them are included in the official split.}
	\label{tab:related-datasets}
\end{table*}

\subsection{Part-aware Panoptic Quality}
\label{sec:task_definition:metric}

With the proposed PPS task, that unifies perception at multiple levels of abstraction, we aim to quantify the performance of the methods for this task using a \textit{single unified metric}. Inspired by the previous Panoptic Quality (PQ) metric \cite{Kirillov2019PS}, we propose \textit{Part-aware Panoptic Quality} (PartPQ). The proposed PartPQ is designed to capture 1) the ability to identify and classify panoptic segments, \ie, stuff regions and things instances, and 2) the part segmentation quality within the identified panoptic segments. 

The PartPQ per scene-level class $l$ is formalized as 
\begin{equation}
\textrm{PartPQ} = \frac{\sum_{(p,g) \in \textit{TP}}\textrm{IOU\textsubscript{p}}(p,g)}{|\textit{TP}| + \frac{1}{2}|\textit{FP}|+ \frac{1}{2}|\textit{FN}|}.
\label{equ:pq-parts}
\end{equation}

As in the original PQ, we assess the ability to identify panoptic segments by counting the amount of true positive, ${TP}$, false positive, ${FP}$, and false negative, ${FN}$, segments, based on the Intersection Over Union (IOU) between a predicted segment $p$ and a ground-truth segment $g$ for a class $l$. A prediction is a ${TP}$ if it has an overlap with a ground-truth segment with an $\textrm{IOU} > 0.5$. An ${FP}$ is a predicted segment that is not matched with the ground-truth, and an ${FN}$ is a ground-truth segment not matched with a prediction.

The part segmentation performance within matched segments is captured by the $\textrm{IOU\textsubscript{p}}(p,g)$ term in Equation \ref{equ:pq-parts}. To be compatible both with scene-level classes with parts ($\mathcal{L}^\text{parts}$), and without parts ($\mathcal{L}^\text{no-parts}$), we define two cases:
\begin{equation}
  \textrm{IOU\textsubscript{p}}(p,g) =
    \begin{cases}
      \textrm{mean IOU\textsubscript{part}}(p,g), & \textrm{$l \in \mathcal{L}^\text{parts}$}\\
      \textrm{IOU\textsubscript{inst}}(p,g), & \textrm{$l \in \mathcal{L}^\text{no-parts}$}
    \end{cases}   
\end{equation}

For the classes $\mathcal{L}^\text{parts}$, we calculate the mean Intersection Over Union for all part classes in the two matched panoptic segments. This is the multi-class mean IOU where the region outside the two segments is labeled background. When computing this score, we allow the prediction to contain pixels with a \textit{void} part label. In the mean IOU, those pixels will not be counted as false positives, but will be counted as false negatives (similar to scene-level \textit{void} labels in PQ~\cite{Kirillov2019PS}). For the subset of classes without parts, $\mathcal{L}^\text{no-parts}$, the instance-level IOU is computed as in the original PQ. 

In essence, the multi-class mean IOU\textsubscript{part} term captures the quality of both the mask of the panoptic segment, and the part segmentation within this segment. Both the quality of the panoptic mask and the part segmentation within the mask need to be high in order to get a high score.

The overall PartPQ is calculated by averaging over all per-class $\textrm{PartPQ}$ scores for scene-level classes $l \in \mathcal{L}$. In Section \ref{sec:experiments}, we evaluate the performance using PartPQ on two datasets. We show that this metric exhibits a reliable performance measure of different approaches, and is consistent with other metrics commonly used for the subtasks combined in part-aware panoptic segmentation.

\section{Datasets}\label{sec:datasets}
We accompany the PPS task with two new datasets, Cityscapes Panoptic Parts (CPP) and PASCAL Panoptic Parts (PPP), which are based on the established scene understanding datasets Cityscapes~\cite{Cordts2016Cityscapes} and PASCAL VOC~\cite{Everingham2010Pascal}, respectively. The introduced datasets include per-pixel annotations on multiple levels of visual abstraction: scene-level and part-level semantics, and instance-level information. As can be seen from Table~\ref{tab:related-datasets}, the existing datasets landscape is inadequate for PPS since no dataset features all of these levels of abstraction. If any combination of the existing datasets is used to achieve multi-level abstraction, conflicts would arise at the pixel level due to overlapping labels. Our datasets comprise a consistent set of annotations, which are free of such conflicts. 

\subsection{Cityscapes Panoptic Parts}
Cityscapes Panoptic Parts (CPP) extends with part-level semantics the popular Cityscapes dataset~\cite{Cordts2016Cityscapes} of urban scenes recorded in Germany and neighboring countries. We manually annotated with 23 part-level semantic classes the original publicly available 2975 training and 500 validation images. We employed a pipeline that takes advantage of original annotations to guide and hint annotators.

CPP is fully compatible with the original Cityscapes panoptic annotations and is, to the best of our knowledge, the first urban scenes dataset with annotations on scene-level, part-level and instance-level, on the same set of images.

\begin{figure}[t]
	\centering
	\includegraphics[width=.9\linewidth]{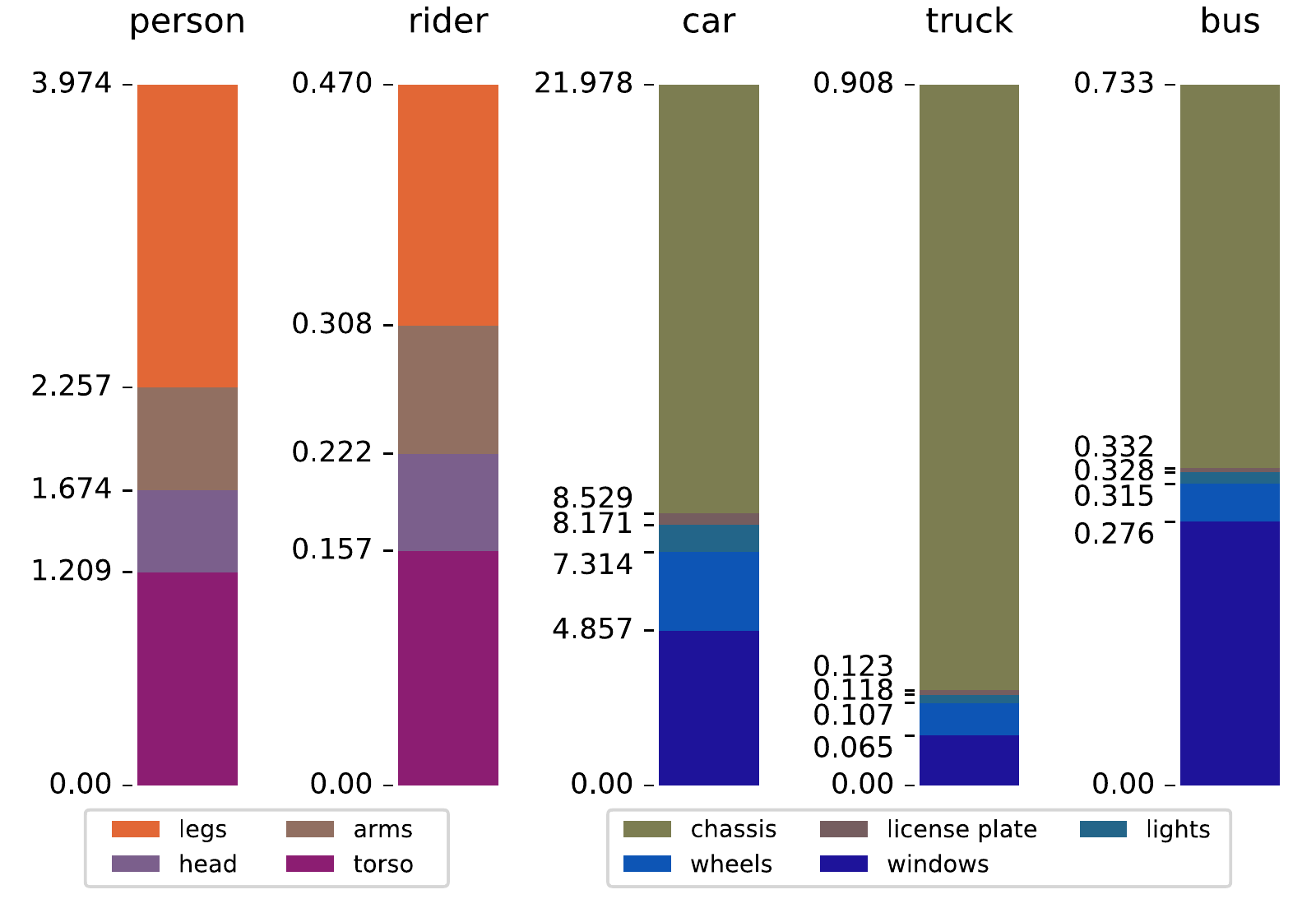}
	\caption{\textbf{Statistics CPP.} Absolute number of Cityscapes pixels ($\times 10^7$) that we annotated per scene-level semantic class.}
	\label{fig:cpp-parts-stats}
	\vspace{-10pt}
\end{figure}

Taking into consideration the complexity of scenes and the variety in number and pose of traffic participants we selected 5 scene-level semantic classes from the \textit{human} and \textit{vehicle} high-level categories to be annotated with parts,~\ie, $\mathcal{L^\text{parts} = \{\textit{person}, \textit{rider}, \textit{car}, \textit{truck}, \textit{bus}\}}$. The \textit{human} categories are annotated with $\mathcal{P}^\textit{human} = \{\textit{torso}, \textit{head}, \textit{arm}, \textit{leg}\}$ and the \textit{vehicle} categories with $\mathcal{P}^\textit{vehicle} = \{\textit{chassis}, \textit{window}, \textit{wheel}, \textit{light}, \textit{license plate}\}$. Statistics for CPP are presented in Table~\ref{tab:related-datasets} and in Figure~\ref{fig:cpp-parts-stats}.

\subsection{PASCAL Panoptic Parts}
PASCAL Panoptic Parts (PPP) extends the PASCAL VOC 2010 benchmark~\cite{Everingham2010Pascal} with part-level and scene-level semantics. The original PASCAL VOC dataset is labeled on scene-level semantics, and only partly on instance-level. A large number of subsequent extensions have been proposed with annotations over different levels of abstraction, leading to various inconsistencies between them at the pixel level. We created PPP by carefully merging PASCAL-Context~\cite{mottaghi14pascalcontext} and PASCAL-Parts~\cite{chen2014detect} to maintain high quality of annotations and solve any conflicts. As the PPP dataset solves conflicts between PASCAL-Context~\cite{mottaghi14pascalcontext} and PASCAL-Parts~\cite{chen2014detect}, evaluations on PPP are not consistent with those on the aforementioned datasets.

PPP preserves the original splitting into 4998 training and 5105 validation images. On the scene-level abstraction PPP contains $\left|\mathcal{L}^\text{Th}\right| = 20$ classes with instance-level annotations and $\left|\mathcal{L}^\text{St}\right| = 80$ classes without instances. On the  part-level abstraction it comprises $\left|\mathfrak{P}\right| = 194$ parts spanning $\left|\mathcal{L}^\text{parts}\right| = 16$ classes, and  $\left|\mathcal{L}^\text{Th} \cap \mathcal{L}^\text{parts}\right| = 16$. For easier comparison with related methods we provide mappings from PPP  to commonly used subsets: 7 parts for human part parsing on PASCAL-Person-Parts~\cite{chen2014detect} and 58 parts for the reduced set used in~\cite{michieli2020gmnet, Zhao2019BSANet}. More statistics can be found in Table~\ref{tab:related-datasets}.

For both CPP and PPP, part-level classes are only defined for scene-level things classes. We anticipate that, in future work, designers of datasets also opt for assigning part classes to stuff classes. If so, this is fully compatible with our task definition and metric, as they already support this.

\section{Experimental analysis}
\label{sec:experiments}
With the introduced task definition, annotations and metric, we now establish benchmarks for the part-aware panoptic segmentation task, and compare the PartPQ metric with existing metrics. The results are presented and explained in Section \ref{sec:experiments:baselines}, and can serve as references for future research. 

Secondly, to get insight into the difference in quality and relative importance of results on the different levels of abstraction in our unified task, and the design choices that play a role in this regard, we conduct several ablation experiments on these baselines in Section \ref{sec:experiments:ablations}.

\begin{table*}[t]
\begin{adjustbox}{width=1\textwidth}
\begin{tabular}{ll||cc|ccc|c||ccc}
& & \multicolumn{6}{l||}{\textit{Before merging}}  & \multicolumn{3}{l}{\textit{After merging}}  \\
 &  & \textbf{mIOU} & \textbf{AP} & \multicolumn{3}{c|}{\textbf{PQ}}  &\textbf{mIOU} & \multicolumn{3}{c}{\textbf{PartPQ}} \\ 
\textbf{Panoptic seg. method}  &  \textbf{Part seg. method} & SemS  & mask  & All & P & NP & PartS & All & P & NP \\ 
\midrule \midrule

\textit{\textbf{Cityscapes Panoptic Parts} val} \\
UPSNet \cite{xiong2019upsnet} & DeepLabv3+ \cite{Chen2018deeplabv3plus} &  75.2  &  33.3 & 59.1 & 57.3 & 59.7 &  75.6  & 55.1 & 42.3 & 59.7 \\
DeepLabv3+ \& Mask R-CNN* \cite{Chen2018deeplabv3plus, he2017mask} & DeepLabv3+ \cite{Chen2018deeplabv3plus} & 78.8 & 36.5 & 61.0 & 58.7 & 61.9 &  75.6 &  56.9 & 43.0 & 61.9  \\ \hline
EfficientPS \cite{mohan2020efficientps} &  BSANet \cite{Zhao2019BSANet} &  80.3 & 39.7  & 65.0 & 64.2 & 65.2  &   76.0  & 60.2 & 46.1 & 65.2  \\
HRNet-OCR \& PolyTransform* \cite{Yuan2020ocr, Liang2020polytransform} &  BSANet \cite{Zhao2019BSANet} &  81.6 & 44.6  & 66.2 & 64.2 & 67.0 &  76.0 & 61.4 & 45.8 & 67.0 \\

\midrule \midrule
\textit{\textbf{Pascal Panoptic Parts} validation} \\
DeepLabv3+ \& Mask R-CNN \cite{Chen2018deeplabv3plus, he2017mask} & DeepLabv3+ \cite{Chen2018deeplabv3plus} & 47.1 & 38.5 & 35.0 & 61.5 & 26.0  & 53.9 &  31.4 & 47.2 & 26.0  \\ \hline
DLv3-ResNeSt269 \& DetectoRS \cite{chen2017deeplabv3, zhang2020resnest, qiao2020detectors} &  BSANet \cite{Zhao2019BSANet} &  55.1 & 44.8 & 42.0 & 66.0 & 33.8 & 58.6 & 38.3 & 51.6 & 33.8 \\ \bottomrule

\end{tabular}
\end{adjustbox}
\vspace{0pt}
\caption{ \textbf{Baselines.} Part-aware panoptic segmentation results for the baselines on the \textit{Cityscapes Panoptic Parts} (CPP) and \textit{Pascal Panoptic Parts} (PPP) datasets, generated using results from commonly used (top), and state-of-the-art methods (bottom) for semantic segmentation, instance segmentation, panoptic segmentation and part segmentation. For the results on CPP, mIOU\textsubscript{PartS} indicates the mean IOU for part segmentation on grouped parts (see Section \ref{sec:experiments:grouping}). Metrics split into \textit{P} and \textit{NP} are evaluated on scene-level classes with and without parts, respectively (see Section \ref{sec:experiments:baselines}). * Indicates pretraining on the COCO dataset \cite{Lin2014COCO}. }
\label{tab:experiments:baselines}
\vspace{-10pt}
\end{table*}

\subsection{Benchmarking}
\label{sec:experiments:baselines}
Since the part-aware panoptic segmentation task and the PartPQ metric are new, there are no methods for this task yet, and hence no results. To fill this gap, we establish baselines for PPS by merging results of methods for the subtasks of panoptic and part segmentation. For this process, we select both state-of-the-art and commonly used methods. The results for these subtasks are mostly generated using publicly available code, or provided to us by the authors of the respective methods. Only in select cases, when trained models are not publicly available, we train an existing network on the concerned data. If so, we indicate this. 

All results are evaluated on the PartPQ metric introduced in Section \ref{sec:task_definition:metric}. We also report on the PartPQ separately for scene-level classes that have parts ($\mathcal{L}^\text{parts}$) with PartPQ\textsubscript{P} and those that do not have parts ($\mathcal{L}^\text{no-parts}$) with PartPQ\textsubscript{NP}. To show the performance of the subtask methods before merging, we also report the performance on the Panoptic Quality (PQ) for panoptic segmentation \cite{Kirillov2019PS} (which we also split in PQ\textsubscript{P} and PQ\textsubscript{NP}), Average Precision (AP) for instance segmentation, and mean Intersection Over Union (mIOU) for semantic segmentation and part segmentation.

\subsubsection{Merging procedure}
To get predictions that adhere to the PPS task defined in Section \ref{sec:task_definition}, we need to merge the results on the subtasks of panoptic segmentation and part segmentation. To achieve this, we maintain a straightforward top-down, rule-based merging approach. First, for scene-level semantic classes that do not have part classes ($l \in \mathcal{L}^\text{no-parts}$), no additional prediction is required, so we copy the predictions from panoptic segmentation. Secondly, for each segment in the panoptic segmentation prediction that does require an additional part label ($l \in \mathcal{L}^\text{parts}$), we identify and extract the part predictions for the pixels corresponding to this segment. If a part prediction contains a part class that does not correspond to the scene-level class (\eg, a \textit{head} pixel in a \textit{bus} segment), we set the part prediction for this pixel to the \textit{void} label.

In Section \ref{sec:experiments:merging}, we show that this top-down merging strategy works better than a strategy that requires the predictions for both part segmentation and panoptic segmentation to agree. It is likely that there is a better, more complex way to construct or possibly learn this merging strategy, but we leave this for future work to address.

\subsubsection{Cityscapes Panoptic Parts}
\label{sec:experiments:baselines:cpp}
\noindent\textbf{Methods.} \enskip
For the baselines on Cityscapes~\cite{Cordts2016Cityscapes}, we generate part-aware panoptic segmentation results using both single network methods for panoptic segmentation \cite{xiong2019upsnet, mohan2020efficientps}, and panoptic segmentation results generated from methods on semantic segmentation~\cite{Chen2018deeplabv3plus, Yuan2020ocr} and instance segmentation~\cite{he2017mask, Liang2020polytransform}. In the latter case, the panoptic segmentation results are created using the heuristic merging process described in \cite{Kirillov2019PS}. For part segmentation, we trained two networks \cite{Chen2018deeplabv3plus, Zhao2019BSANet} ourselves, since we are the first to introduce part labels on Cityscapes.

For all baselines, in order to have fair and consistent results, we use methods that are only trained on the Cityscapes \texttt{train} set without using the \textit{coarse} labels, with pre-training only on ImageNet \cite{Deng2009ImageNet}. The only exceptions are the instance segmentation methods \cite{he2017mask, Liang2020polytransform}, which are pre-trained on COCO~\cite{Lin2014COCO}, as has become common practice.

\begin{figure*}[t]
	\centering
    	\includegraphics[width=0.22\linewidth]{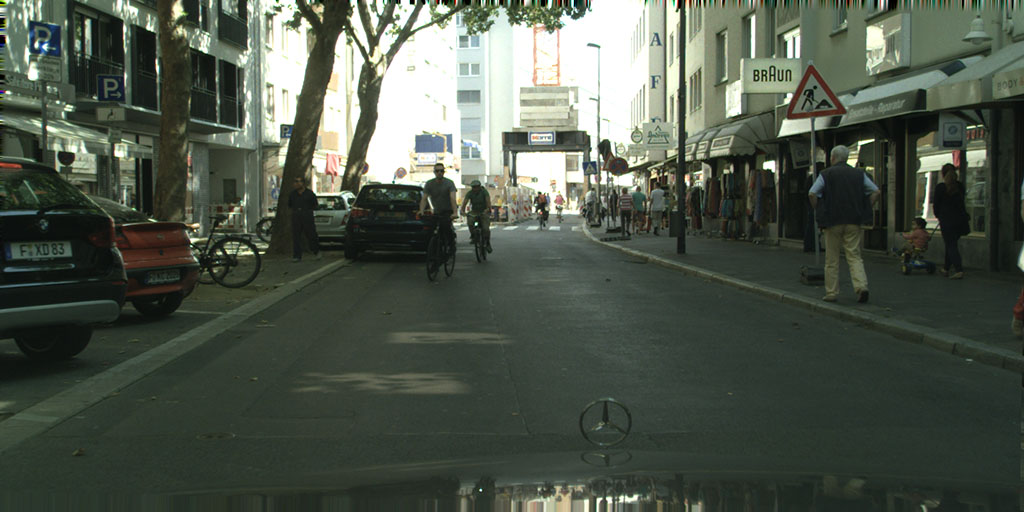}
    	\includegraphics[width=0.22\linewidth]{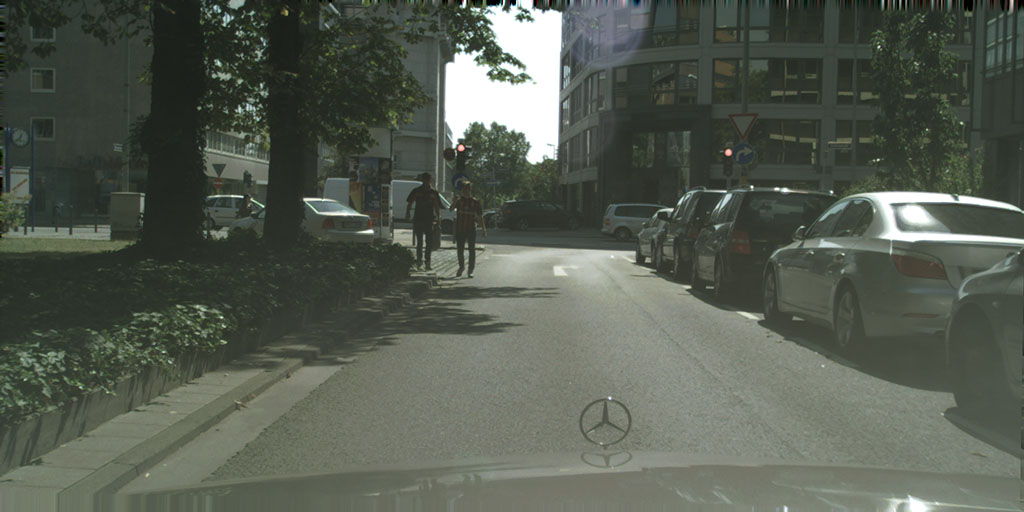}
    	\includegraphics[width=0.22\linewidth]{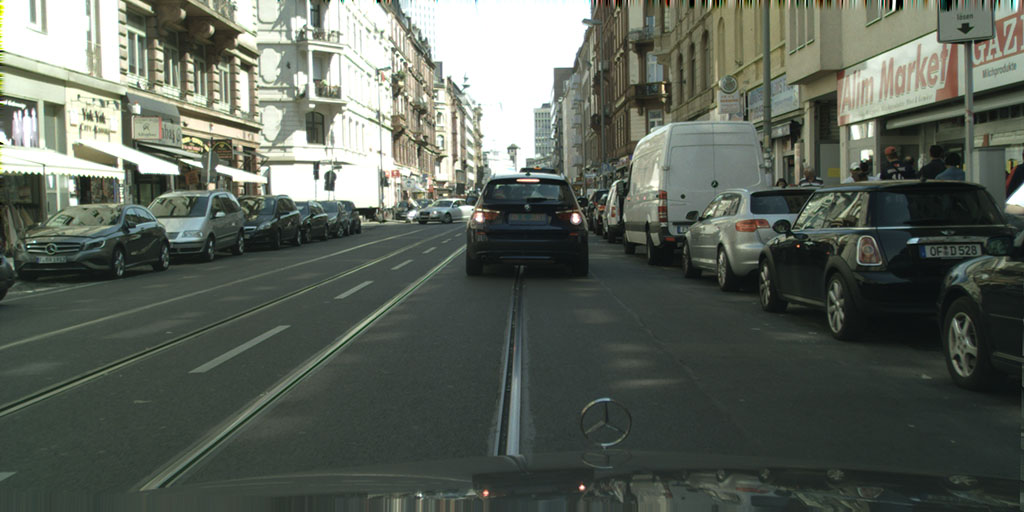}
    	\includegraphics[width=0.22\linewidth]{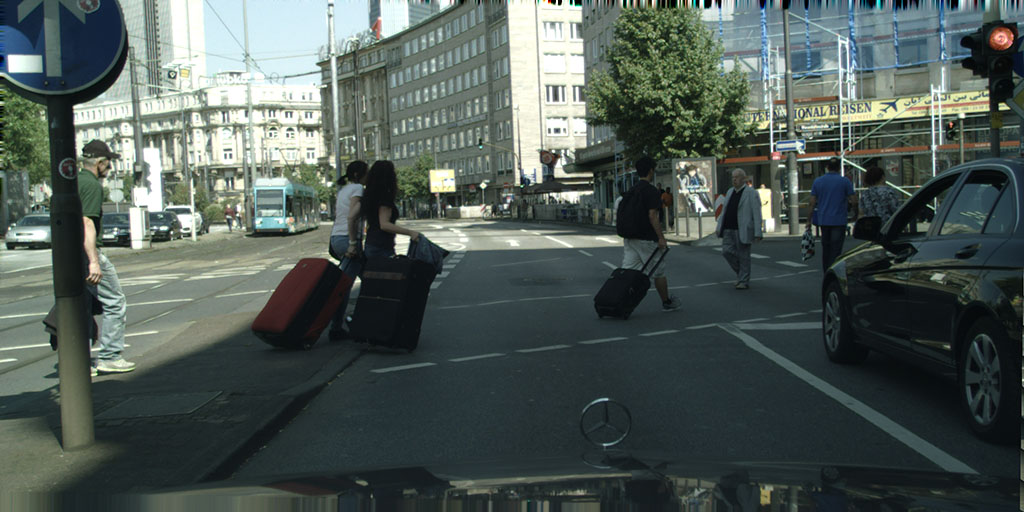}\\
		\includegraphics[width=0.22\linewidth]{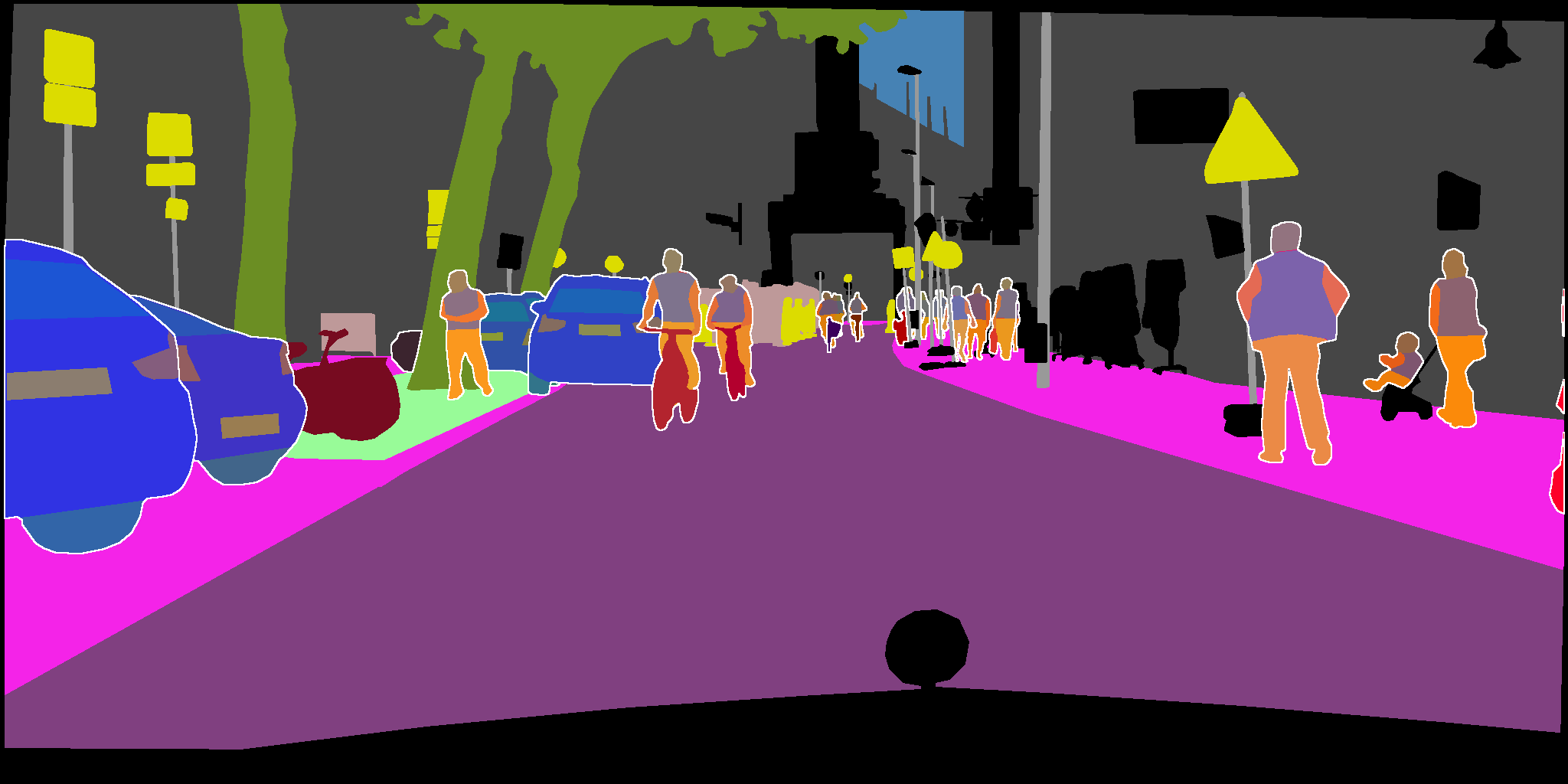}
		\includegraphics[width=0.22\linewidth]{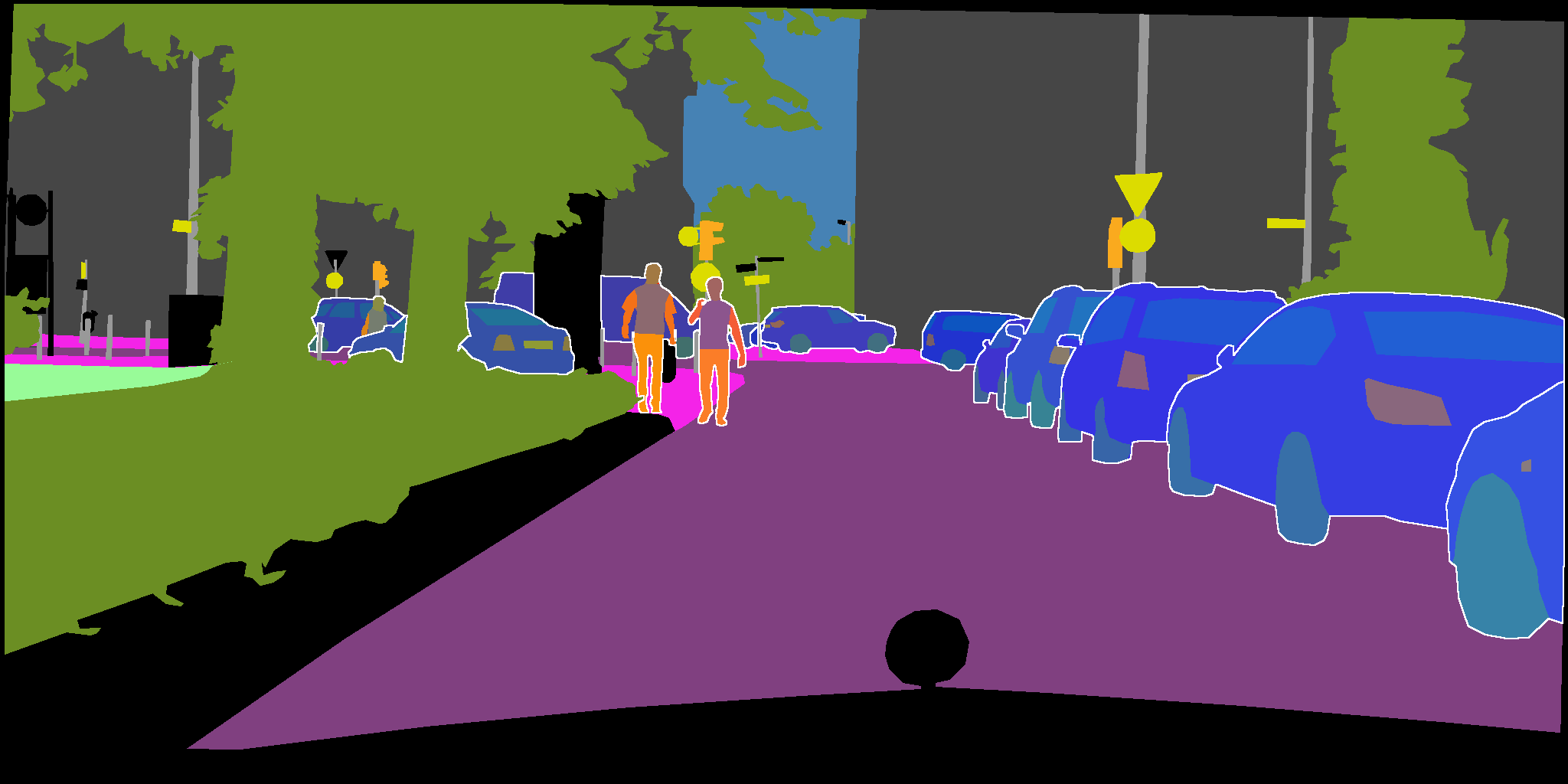}
		\includegraphics[width=0.22\linewidth]{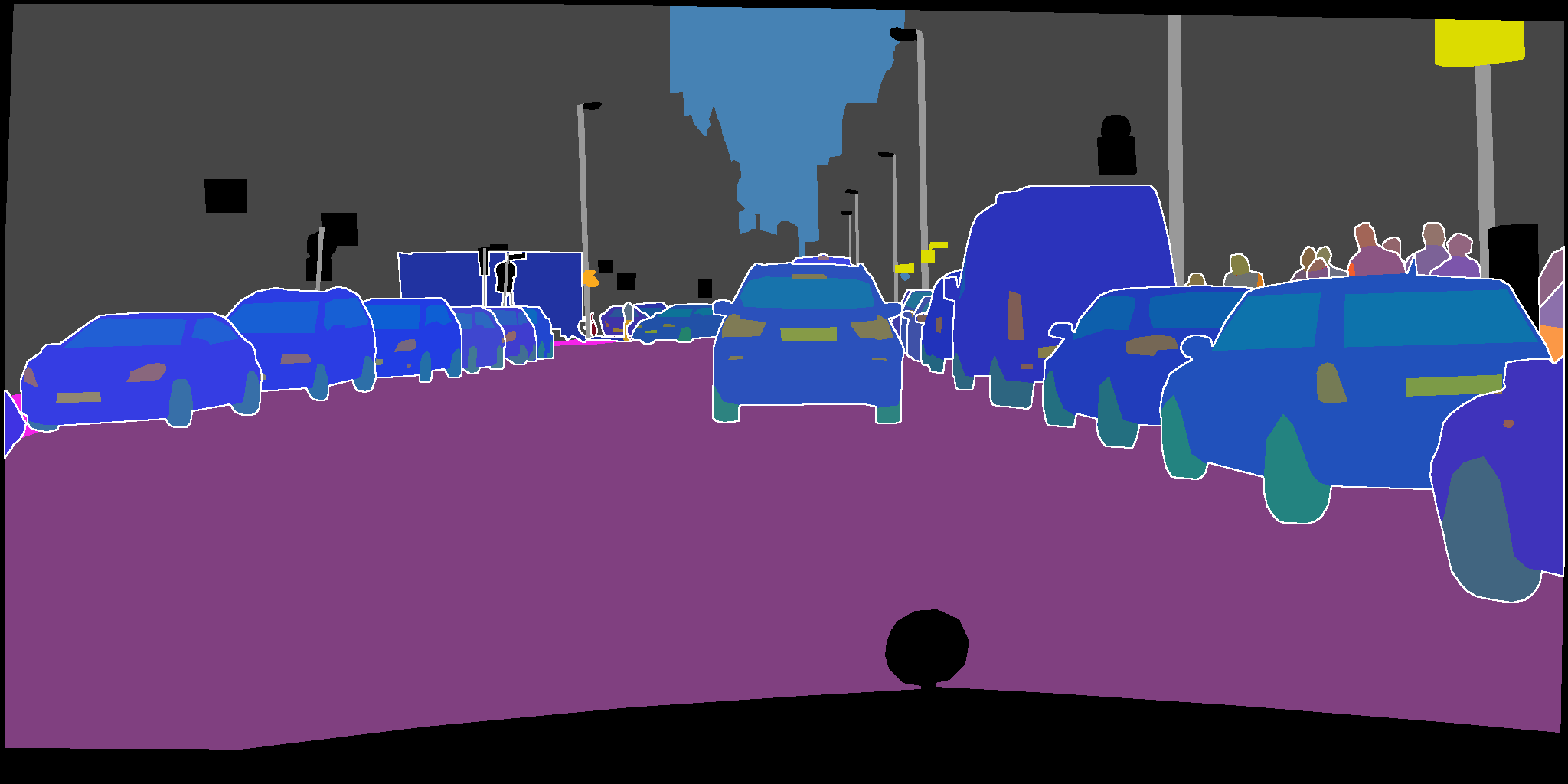}
		\includegraphics[width=0.22\linewidth]{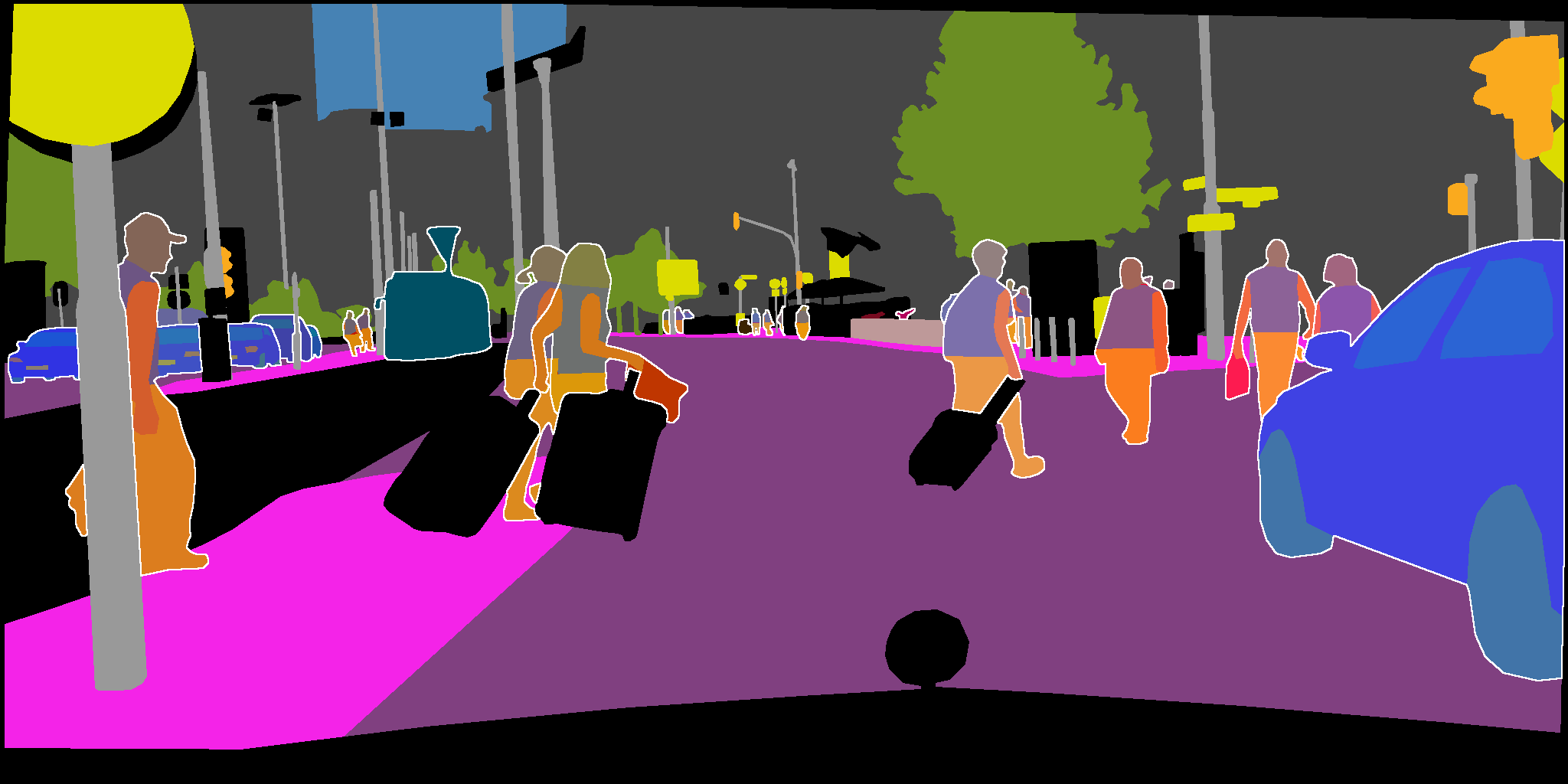} \\
		\includegraphics[width=0.22\linewidth]{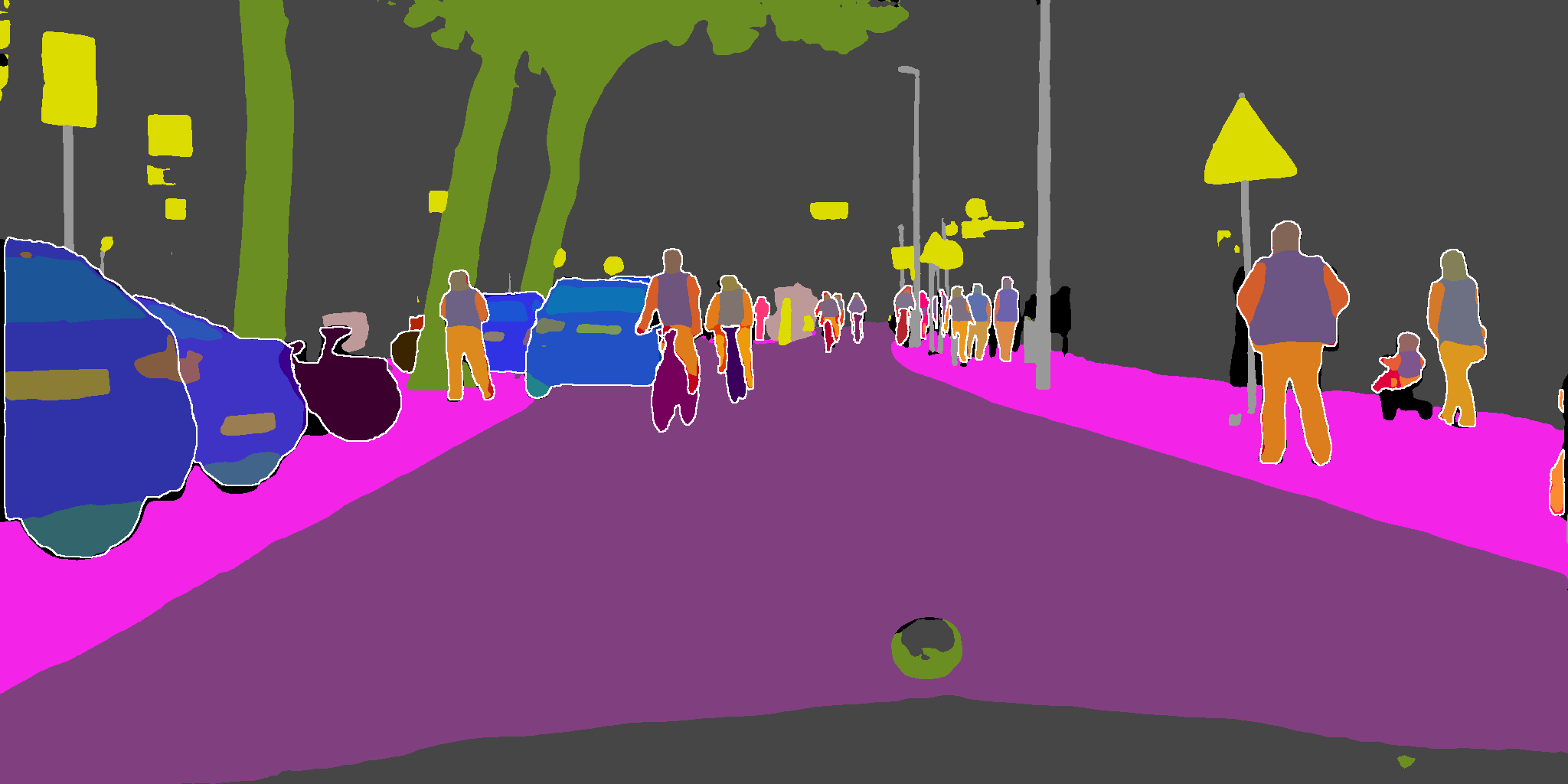}
		\includegraphics[width=0.22\linewidth]{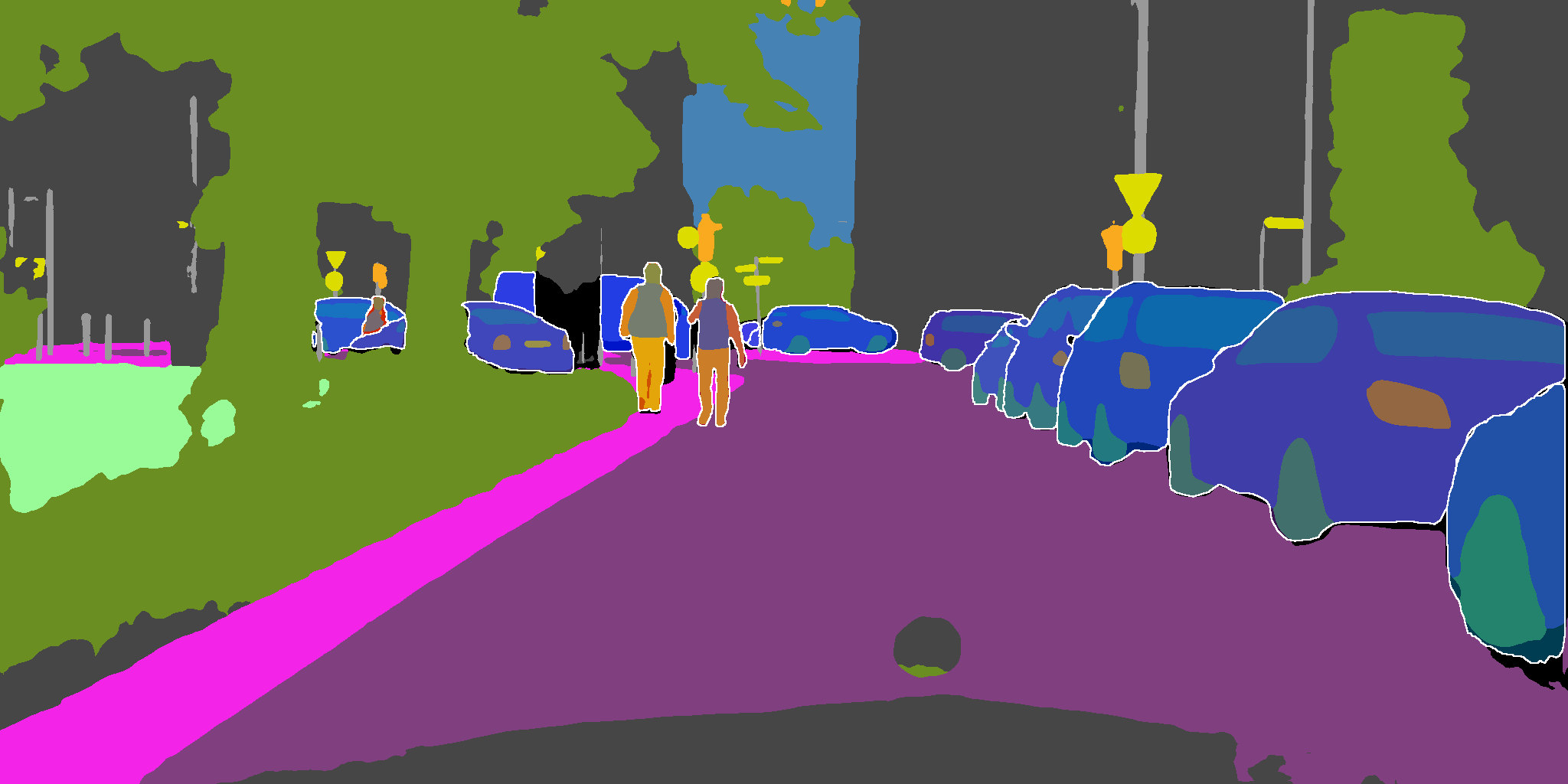}
		\includegraphics[width=0.22\linewidth]{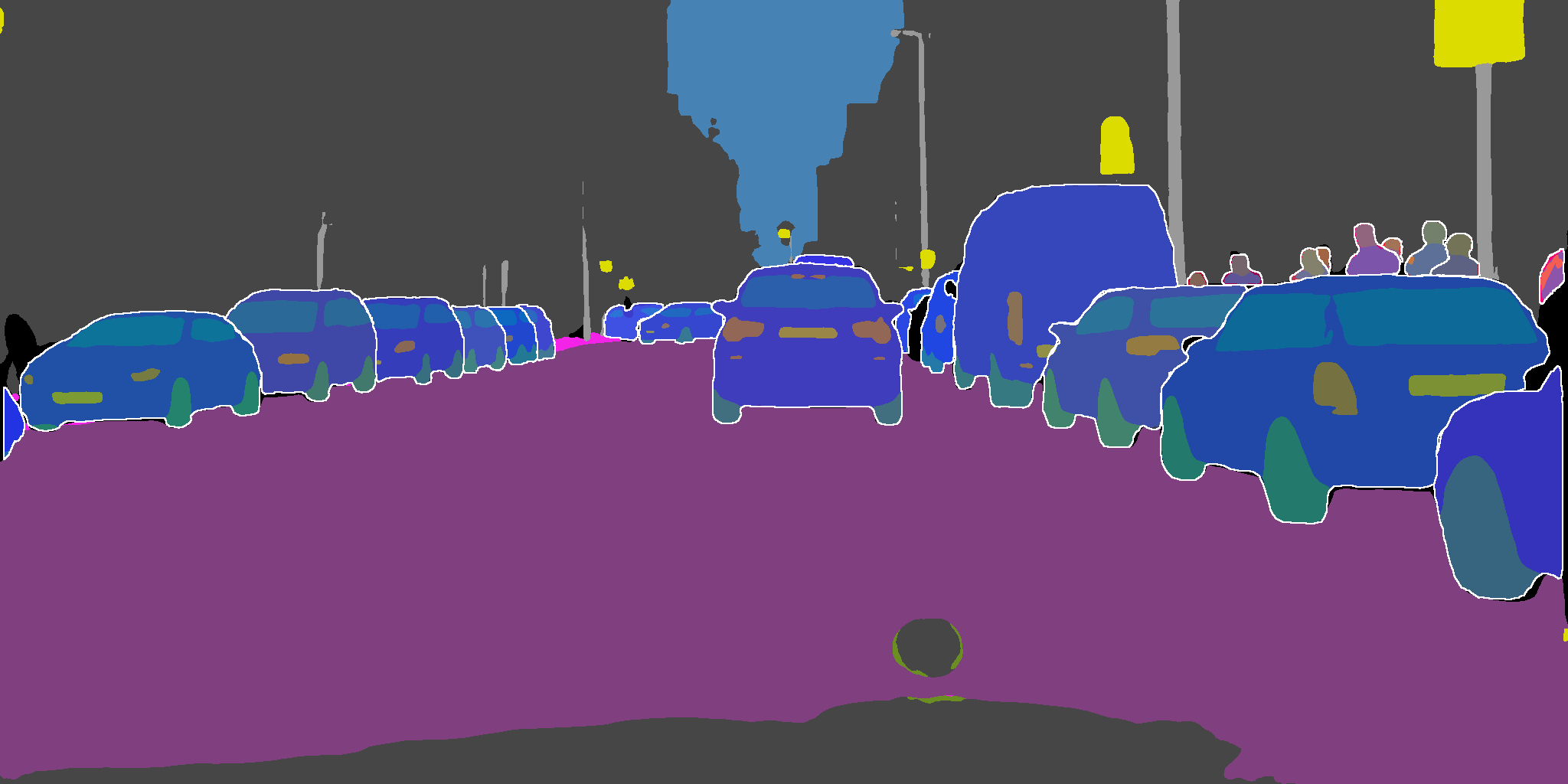}
		\includegraphics[width=0.22\linewidth]{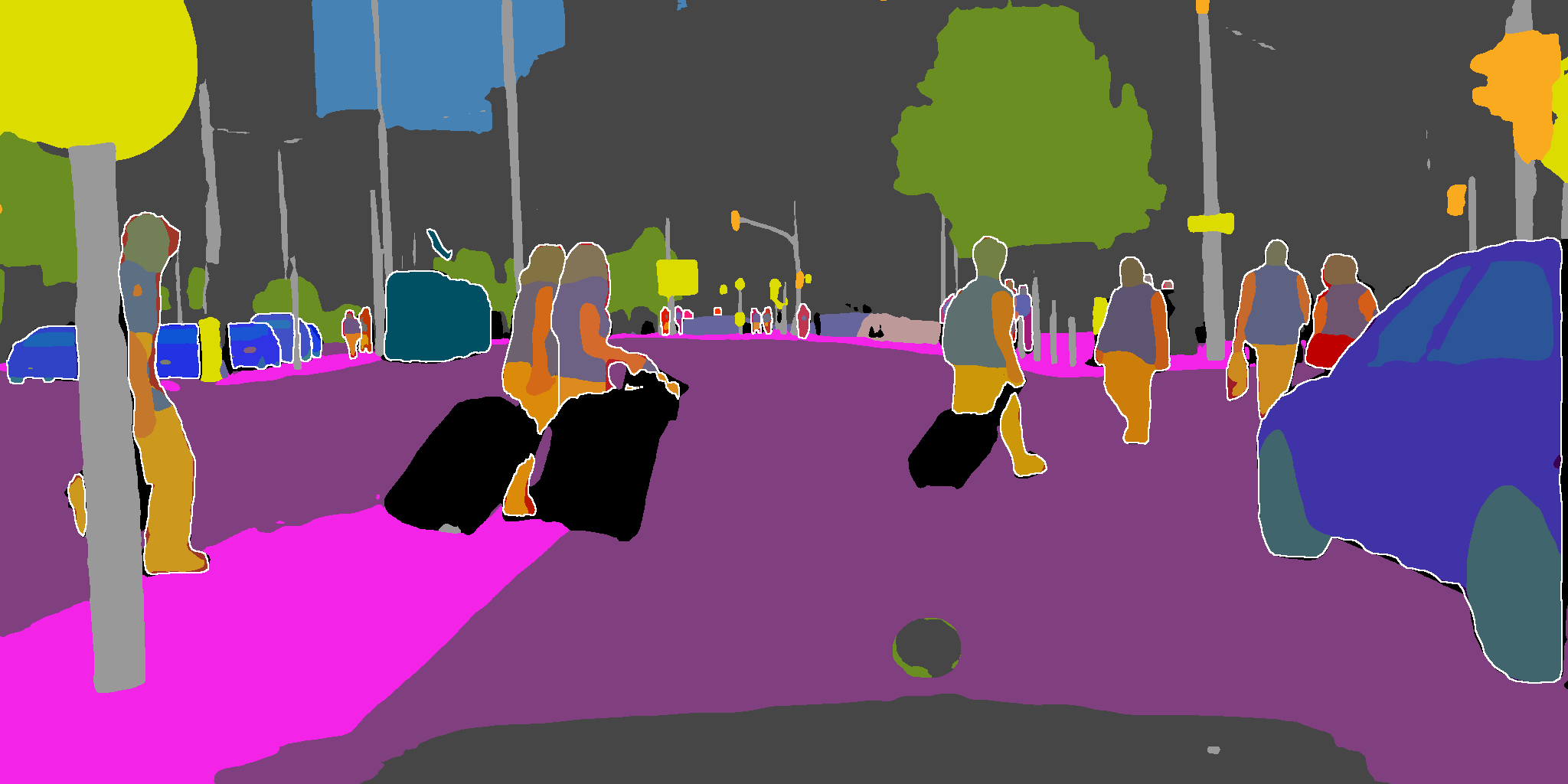} \\
	\caption{\textbf{Examples Cityscapes Panoptic Parts.} Top: input; middle: ground truth; bottom: predictions highest-scoring PPS baseline.}
\vspace{-6pt}
\label{fig:cpp_examples}
\end{figure*}

\vspace{2pt}
\noindent\textbf{Results.} \enskip
With the aforementioned state-of-the-art methods and merging strategy, we set state-of-the-art baselines for part-level panoptic segmentation. The results for Cityscapes Panoptic Parts are reported in Table \ref{tab:experiments:baselines}, and qualitative examples are shown in Figure \ref{fig:cpp_examples}. The results show that the scores on PartPQ are lower than the regular PQ, which is expectable, as we add complexity to the problem with part-level segmentation of segments, and the per-instance IOU of PQ is replaced with the part-level IOU in PartPQ. As expected, the scores for PartPQ\textsubscript{NP} are identical to PQ\textsubscript{NP}, as the results and metric for $\mathcal{L}^\text{no-parts}$ are unchanged. When comparing the PartPQ to other metrics, we see that a difference in scores between methods is comparable to the existing metrics on the subtasks. This indicates that the metric captures the aspects covered by those metrics, while being a single metric for the unified task of part-aware panoptic segmentation. 

\subsubsection{Pascal Panoptic Parts}
\label{sec:experiments:baselines:ppp}
\noindent\textbf{Methods.} \enskip
Due to a lack of existing work on panoptic segmentation for the Pascal VOC dataset \cite{Everingham2010Pascal}, we generate panoptic segmentation results by fusing semantic segmentation \cite{chen2017deeplabv3, zhang2020resnest} and instance segmentation \cite{he2017mask, qiao2020detectors} results, following~\cite{Kirillov2019PS}. Specifically, the semantic segmentation methods are generated using existing models trained on 59 classes of the Pascal-Context dataset \cite{mottaghi14pascalcontext}, and we train the instance segmentation models on the 20 things classes of our Pascal Panoptic Parts dataset. For part segmentation, we generate state-of-the-art results using an existing model \cite{Zhao2019BSANet} trained on a dataset that includes 58 part classes from the Pascal-Parts dataset \cite{chen2014detect}, and we train another commonly used model \cite{Chen2018deeplabv3plus} on that same dataset. Despite the different annotations used for training, all models are trained on the same 4998 images in the Pascal VOC 2010 \texttt{training} split.

\begin{figure*}[t]
	\centering
    	\includegraphics[height=0.083\textheight]{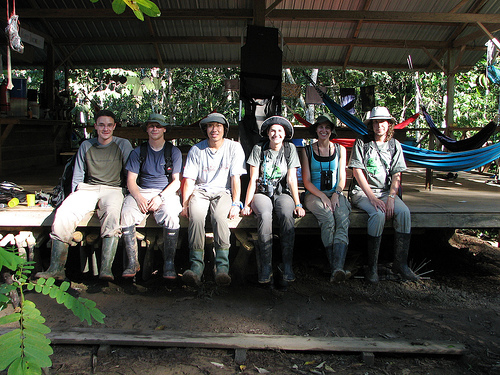}
    	\includegraphics[height=0.083\textheight]{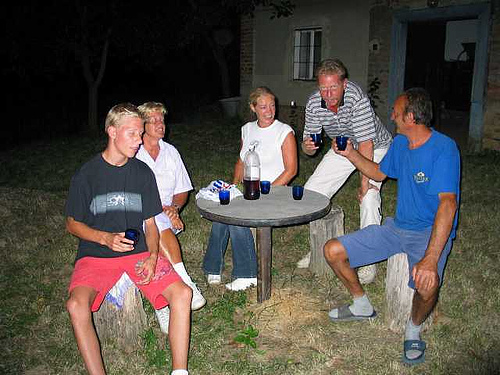}
    	\includegraphics[height=0.083\textheight]{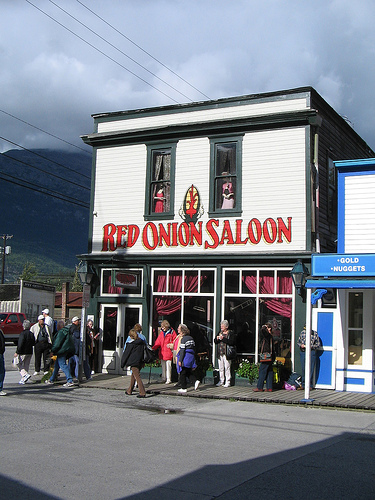}
    	\includegraphics[height=0.083\textheight]{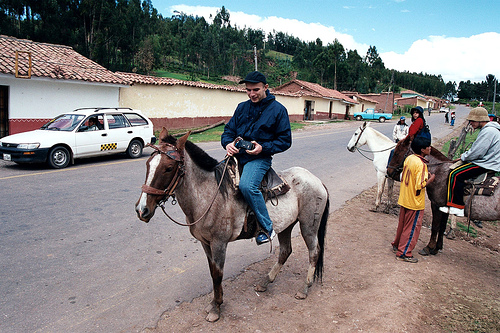}
    	\includegraphics[height=0.083\textheight]{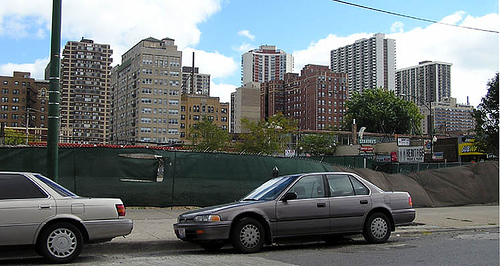}
    	\includegraphics[height=0.083\textheight]{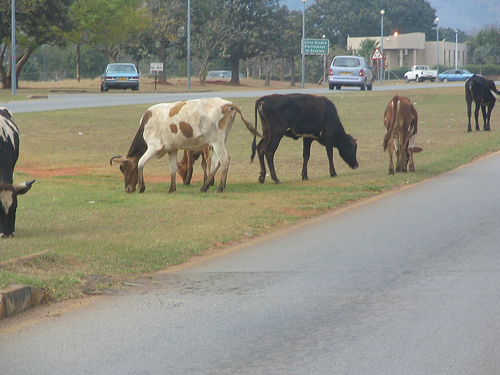}\\

		\includegraphics[height=0.083\textheight]{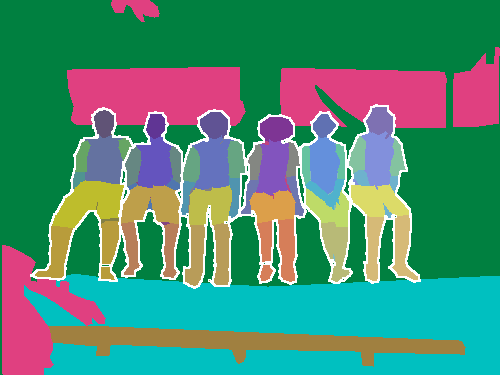}
		\includegraphics[height=0.083\textheight]{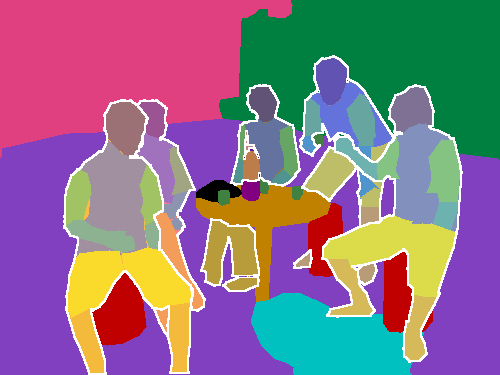}
		\includegraphics[height=0.083\textheight]{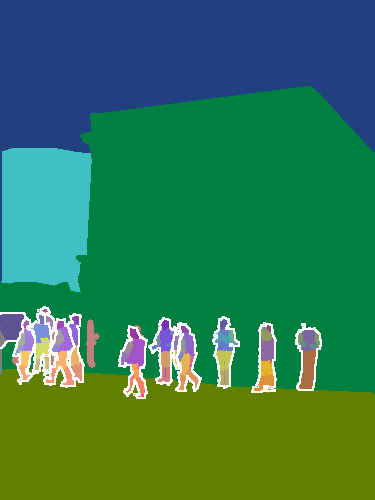}
		\includegraphics[height=0.083\textheight]{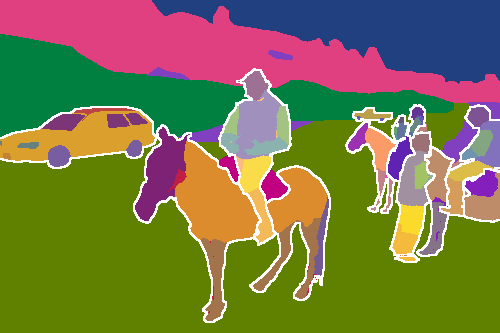}
		\includegraphics[height=0.083\textheight]{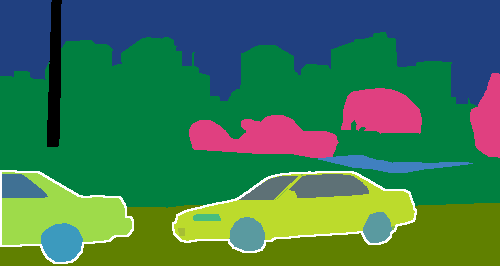}
		\includegraphics[height=0.083\textheight]{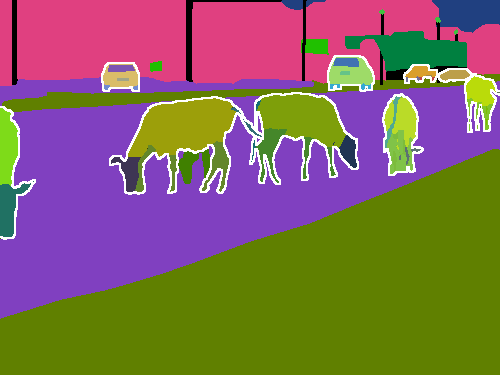}\\

		\includegraphics[height=0.083\textheight]{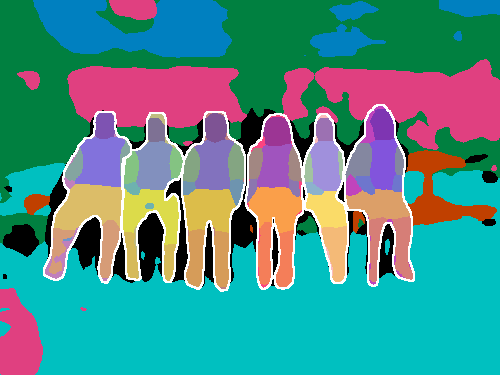}
		\includegraphics[height=0.083\textheight]{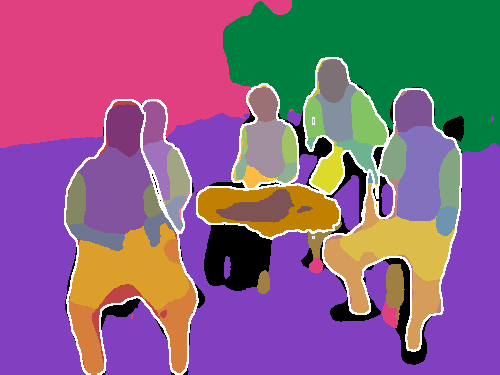}
		\includegraphics[height=0.083\textheight]{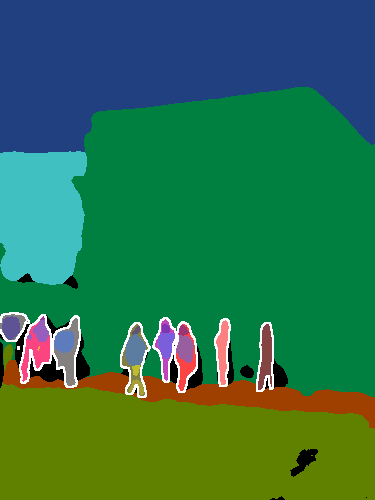}
		\includegraphics[height=0.083\textheight]{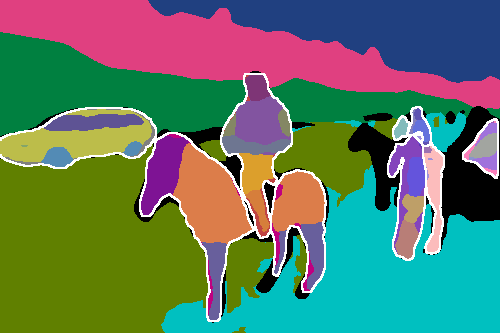}
		\includegraphics[height=0.083\textheight]{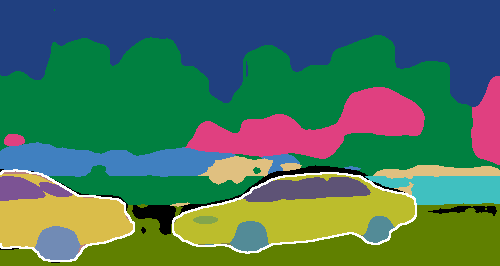}
		\includegraphics[height=0.083\textheight]{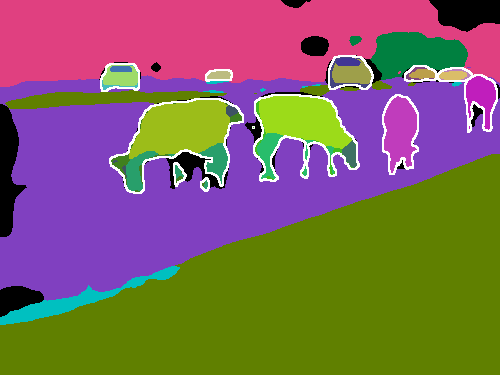} \\
	\caption{\textbf{Examples Pascal Panoptic Parts.} Top: input; middle: ground truth; bottom: predictions highest-scoring PPS baseline.}
\vspace{-10pt}
\label{fig:ppp_examples}
\end{figure*}

\vspace{2pt}
\noindent\textbf{Results.} \enskip
The results for the baselines on the PPP dataset are reported in Table \ref{tab:experiments:baselines}. From the table, it is clear that, again, scores for PartPQ increase proportionally to the existing metrics for the subtasks, and that PartPQ\textsubscript{NP} remains identical to PQ\textsubscript{NP}. Qualitative examples are displayed in Figure \ref{fig:ppp_examples}.

\subsection{Ablation experiments}
\label{sec:experiments:ablations}

\begin{table}
\centering
\begin{adjustbox}{width=0.9\linewidth}
\begin{tabular}{c|cc|cc}
\toprule
& \multicolumn{2}{l|}{\textit{Before merging}} & \multicolumn{2}{l}{\textit{After merging}}  \\
\textbf{Merging str.} & \textbf{PQ} & \textbf{mIOU\textsubscript{PartS}} & \textbf{PartPQ}  & \textbf{PartPQ\textsubscript{P}} \\ \hline\hline
\multicolumn{5}{l}{\textit{State-of-the-art results on Cityscapes Panoptic Parts}} \\
original & 66.2 &  67.2 &  \textbf{60.9} & \textbf{44.0} \\ 
alternative & 66.2 &  67.2 & 60.2 & 41.3 \\\hline
\multicolumn{5}{l}{\textit{State-of-the-art results on Pascal Panoptic Parts}} \\
original & 42.0 & 58.6 & \textbf{38.3} & \textbf{51.6} \\ 
alternative & 42.0 & 58.6 & 37.5 & 50.6  \\ \bottomrule
\end{tabular}
\end{adjustbox}
\vspace{3pt}
\caption{The results of \textbf{different merging procedures}, on the \texttt{val} split of CPP and the \texttt{validation} split of PPP. The \textit{original} merging strategy prioritizes panoptic segmentation; the \textit{alternative} strategy requires both predictions to agree on the  scene-level label.}
\label{tab:experiments:merging}
\vspace{-10pt}
\end{table}

\subsubsection{Merging panoptic and part segmentation}
\label{sec:experiments:merging}
\noindent\textbf{Experiment.} \enskip
For the aforementioned baselines, we use a top-down merging strategy that effectively prioritizes panoptic segmentation over part segmentation, by taking the scene-level semantic label from the panoptic output. It is also possible to take a more conservative approach that also considers bottom-up information, by requiring panoptic and part segmentation predictions to agree on the scene-level semantic label. For this alternative approach, a panoptic segment is compared with the part predictions at the corresponding pixels, and for each pixel, the panoptic prediction is only kept if the part prediction is possible for the scene-level label of that panoptic segment (\eg, \textit{truck-wheel} for a \textit{truck} instance). Otherwise, the pixel is removed from the segment, and both the scene-level and the part-level predictions are set to \textit{void}. This merging approach would lead to better results than the original, if the panoptic segmentation method frequently makes mistakes that the part segmentation method does not make, and if part segmentation predictions are not incorrect where panoptic segmentation is correct.

\begin{table}
\begin{adjustbox}{width=1\linewidth}
\begin{tabular}{c|ccc|cc}
\toprule
\textbf{Grouping} & \textbf{PQ} & \textbf{mIOU} & \textbf{mIOU\textsubscript{grouped}} & \textbf{PartPQ}  & \textbf{PartPQ\textsubscript{P}} \\ \hline\hline
\multicolumn{6}{l}{\textit{Commonly used methods for panoptic seg. and part seg.}} \\
- & 61.0  &  54.3 & 74.5 &  55.8 & 38.8 \\ 
\checkmark & 61.0  &  \textit{n/a} & \textbf{75.6} &  \textbf{56.9} & \textbf{43.0} \\\hline
\multicolumn{6}{l}{\textit{State-of-the-art methods for panoptic seg. and part seg.}} \\
- & 66.2 & 67.2 & 75.3 &  60.9 &  44.0 \\ 
\checkmark & 66.2 & \textit{n/a} & \textbf{76.0} & \textbf{61.4} &  \textbf{45.8} \\ \bottomrule
                          
\end{tabular}
\end{adjustbox}
\vspace{0pt}
\caption{\textbf{Grouping parts}. Trained on the Cityscapes Panoptic Parts set using grouped parts: 1) \textit{car}, \textit{bus} and \textit{truck} parts, and 2) \textit{person} and \textit{rider} parts. Reported mIOU scores are for part-level semantics.}
\label{tab:experiments:grouping_cs}
\vspace{-5pt}
\end{table}

\vspace{2pt}
\noindent\textbf{Results.} \enskip
The results, reported in Table \ref{tab:experiments:merging}, clearly show that the original merging method performs better. For the alternative approach, the PartPQ for classes with parts is consistently lower. This occurs as pixels are incorrectly removed from segments. These results clearly indicate that it is better to prioritize the scene-level label from panoptic segmentation over that from part segmentation.

\begin{table}[t]
\begin{adjustbox}{width=1\linewidth}
\begin{tabular}{cc|cc|cc}
\toprule
\multicolumn{2}{l|}{\textit{Panoptic seg.}} & \multicolumn{2}{l|}{\textit{Part seg.}} & \multicolumn{2}{l}{\textit{Semantic information gain}}  \\
\textbf{mPA} & \textbf{mIOU} & \textbf{mPA} & \textbf{mIOU} & \textbf{mSIG\textsubscript{pan$\rightarrow$part}} & \textbf{mSIG\textsubscript{part$\rightarrow$pan}} \\ 
\midrule\midrule
\vspace{-1pt}
91.6 & 85.9 & 88.6 & 82.5 & 54.1 & 39.4 \\ 
\bottomrule

\end{tabular}
\end{adjustbox}
\vspace{0pt}
\caption{\textbf{Comparing performance on scene-level semantics} between \textit{state-of-the-art} methods for panoptic segmentation and part segmentation, on Cityscapes Panoptic Parts \texttt{val}.}
\label{tab:experiments:potential_improvement_cs}
\vspace{-10pt}
\end{table}

\subsubsection{Grouping semantically similar parts}
\label{sec:experiments:grouping}

\noindent\textbf{Experiment.} \enskip
The results from Section \ref{sec:experiments:merging} suggest that methods trained on panoptic segmentation are better able to predict scene-level semantics than part segmentation methods, favoring a top-down approach to PPS. To further explore the potential benefits of a top-down approach, we conduct experiments where we train a part segmentation method on parts that are grouped by semantic similarity (\eg, \textit{bus-wheel} and \textit{car-wheel} are grouped as \textit{wheel}). This is likely to work because 1) there is more data per part class and 2) there is less ambiguity between the part classes. This favors a top-down approach because it means that, to get a prediction in the PPS format, the scene-level label needs to be extracted from panoptic segmentation, and that part segmentation is used to learn the specific parts only.

\vspace{2pt}
\noindent\textbf{Results.} \enskip
We train part segmentation networks for which the parts for 1) \textit{car}, \textit{bus}, and \textit{truck}, and 2) \textit{person} and \textit{rider}, are grouped, effectively reducing the amount of parts from 23 to 9. The results for this experiment are shown in Table \ref{tab:experiments:grouping_cs}, and they show that the PartPQ for classes with parts, PartPQ\textsubscript{P}, increases with up to 4.2 points when parts are grouped. This supports our hypothesis.

\subsubsection{Comparing levels of abstraction}
\label{sec:experiments:potential_improvement}
In the previous experiments, we have seen results that indicate that it is sensible to approach PPS in a top-down manner, \ie, to first predict the scene-level semantic label, and then look for parts within those regions. To further substantiate this hypothesis, and to assess what the main information source should be for scene-level semantics, we conduct an additional experiment that compares the scene-level performance of methods trained on panoptic and part segmentation.

\vspace{2pt}
\noindent\textbf{Metrics.} \enskip
To assess the extent to which correct scene-level information is available in one method, but not in another, we introduce the \textit{Semantic Information Gain} (SIG) metric, 
which quantifies the extent to which errors made by a given method \textit{B} can be compensated for by the correct predictions of a method \textit{A}. We define the the SIG of method A with respect to method B, $SIG_{{A}\rightarrow{B}}$, as 
\begin{equation}
    SIG_{{A}\rightarrow{B}} = \frac{1}{\left|{X}_{FP_{B}}\right|}\sum_{x\in{X}_{FP_{B}}}{TP_{A,x}} \times 100 \%,
\end{equation}
where ${X}_{FP_{B}}$ is the set of pixels incorrectly predicted by method B, 
and $TP_{A,x} = 1$ if method A is correct at pixel $x$ and $TP_{A,x} = 0$ otherwise. We evaluate the SIG per class in the ground truth, and report the mean SIG (mSIG) over all scene-level classes with parts, $\mathcal{L}^\text{parts}$. We also report on mean Pixel Accuracy (mPA) and mean Intersection Over Union (mIOU).

\vspace{2pt}
\noindent\textbf{Results.} \enskip
When looking at the results in Table \ref{tab:experiments:potential_improvement_cs}, it is clear that the panoptic segmentation method is considerably more accurate on the concerning five scene-level classes than part segmentation. Moreover, panoptic segmentation predictions can resolve, on average, 54.1\% of the errors made by part segmentation. Specifically, these errors seem to occur for classes that have parts that could be confused with each other (\eg, \textit{bus} and \textit{truck}). This supports the aforementioned hypothesis about a top-down approach being a good way to approach part-aware panoptic segmentation.

This does not mean, however, that the bottom-up alternative has no potential at all. Table \ref{tab:experiments:potential_improvement_cs} shows that, to a lesser degree, part segmentation can also solve errors made by the panoptic segmentation method. Therefore, it is likely that a future top-down method for PPS could be improved when enriched with specific bottom-up features.

\section{Conclusion}
In this work, we presented the novel task of part-aware panoptic segmentation (PPS), which takes the next step in holistic scene understanding by unifying scene parsing and part parsing. With the accompanying metric and datasets, we have generated state-of-the-art results for this task, constructed from state-of-the-art results on the underlying subtasks. We hope that this work will spark new innovations in the area of scene understanding, for which our results can serve as baselines.

Specifically, we hope to see innovations in single-network PPS methods that learn the levels of abstraction -- \ie, part-level and scene-level -- jointly, to leverage the interaction between these levels during training. An important design choice is the way in which information from these levels of abstraction is combined. From the experiments conducted in this work, we observe that results suggest that it is best to maintain a top-down approach, where panoptic predictions are extended with part-level predictions.

To provide a foundation for future research, the code and data used to realize this work are shared with the research community.

\vspace{-10pt}
\paragraph{Acknowledgements}
We thank the authors of EfficientPS~\cite{mohan2020efficientps} and PolyTransform~\cite{Liang2020polytransform} for providing us with the predictions by their networks. This work is supported by Eindhoven Engine, NXP Semiconductors and Brainport Eindhoven, and partly funded by the Netherlands Organization for Scientific Research (NWO) in the context of the i-CAVE programme.

{\small
\bibliographystyle{ieee_fullname}
\bibliography{biblio}

\begin{thebibliography}{10}\itemsep=-1pt

\bibitem{alp2018densepose}
R{\i}za Alp~G{\"u}ler, Natalia Neverova, and Iasonas Kokkinos.
\newblock Densepose: Dense human pose estimation in the wild.
\newblock In {\em CVPR}, 2018.

\bibitem{Chen2019HTC}
Kai Chen, Jiangmiao Pang, Jiaqi Wang, Yu Xiong, Xiaoxiao Li, Shuyang Sun,
  Wansen Feng, Ziwei Liu, Jianping Shi, Wanli Ouyang, Chen~Change Loy, and
  Dahua Lin.
\newblock {Hybrid Task Cascade for Instance Segmentation}.
\newblock In {\em CVPR}, 2019.

\bibitem{chen2019mmdetection}
Kai Chen, Jiaqi Wang, Jiangmiao Pang, Yuhang Cao, Yu Xiong, Xiaoxiao Li,
  Shuyang Sun, Wansen Feng, Ziwei Liu, Jiarui Xu, Zheng Zhang, Dazhi Cheng,
  Chenchen Zhu, Tianheng Cheng, Qijie Zhao, Buyu Li, Xin Lu, Rui Zhu, Yue Wu,
  Jifeng Dai, Jingdong Wang, Jianping Shi, Wanli Ouyang, Chen~Change Loy, and
  Dahua Lin.
\newblock {MMDetection}: Open mmlab detection toolbox and benchmark.
\newblock {\em arXiv preprint arXiv:1906.07155}, 2019.

\bibitem{chen2017deeplabv3}
Liang-Chieh Chen, George Papandreou, Florian Schroff, and Hartwig Adam.
\newblock {Rethinking Atrous Convolution for Semantic Image Segmentation}.
\newblock {\em arXiv preprint arXiv:1706.05587}, 2017.

\bibitem{Chen2018deeplabv3plus}
Liang-Chieh Chen, Yukun Zhu, George Papandreou, Florian Schroff, and Hartwig
  Adam.
\newblock {Encoder-Decoder with Atrous Separable Convolution for Semantic Image
  Segmentation}.
\newblock In {\em ECCV}, 2018.

\bibitem{chen2014detect}
Xianjie Chen, Roozbeh Mottaghi, Xiaobai Liu, Sanja Fidler, Raquel Urtasun, and
  Alan Yuille.
\newblock Detect what you can: Detecting and representing objects using
  holistic models and body parts.
\newblock In {\em CVPR}, 2014.

\bibitem{Cheng2020PanopticDeepLab}
Bowen {Cheng}, Maxwell~D. {Collins}, Yukun {Zhu}, Ting {Liu}, Thomas~S.
  {Huang}, Hartwig {Adam}, and Liang-Chieh {Chen}.
\newblock {Panoptic-DeepLab: A Simple, Strong, and Fast Baseline for Bottom-Up
  Panoptic Segmentation}.
\newblock In {\em CVPR}, 2020.

\bibitem{Chollet2017Xception}
Francois Chollet.
\newblock {Xception: Deep Learning With Depthwise Separable Convolutions}.
\newblock In {\em CVPR}, 2017.

\bibitem{Cordts2016Cityscapes}
Marius Cordts, Mohamed Omran, Sebastian Ramos, Timo Rehfeld, Markus Enzweiler,
  Rodrigo Benenson, Uwe Franke, Stefan Roth, and Bernt Schiele.
\newblock {The Cityscapes Dataset for Semantic Urban Scene Understanding}.
\newblock In {\em CVPR}, 2016.

\bibitem{dang2019deep}
Qi Dang, Jianqin Yin, Bin Wang, and Wenqing Zheng.
\newblock {Deep learning based 2D human pose estimation: A survey}.
\newblock {\em Tsinghua Science and Technology}, 24(6):663--676, 2019.

\bibitem{degeus2019single}
Daan de Geus, Panagiotis Meletis, and Gijs Dubbelman.
\newblock {Single Network Panoptic Segmentation for Street Scene
  Understanding}.
\newblock In {\em IEEE Intelligent Vehicles Symposium (IV)}, 2019.

\bibitem{degeus2020fpsnet}
Daan de Geus, Panagiotis Meletis, and Gijs Dubbelman.
\newblock {Fast Panoptic Segmentation Network}.
\newblock {\em IEEE Robotics and Automation Letters}, 5(2):1742--1749, 2020.

\bibitem{Deng2009ImageNet}
Jia {Deng}, Wei {Dong}, Richard {Socher}, Li-Jia {Li}, {Kai Li}, and {Li
  Fei-Fei}.
\newblock {ImageNet: A large-scale hierarchical image database}.
\newblock In {\em CVPR}, 2009.

\bibitem{dong2014towards}
Jian Dong, Qiang Chen, Xiaohui Shen, Jianchao Yang, and Shuicheng Yan.
\newblock Towards unified human parsing and pose estimation.
\newblock In {\em CVPR}, 2014.

\bibitem{dong2013deformable}
Jian Dong, Qiang Chen, Wei Xia, Zhongyang Huang, and Shuicheng Yan.
\newblock A deformable mixture parsing model with parselets.
\newblock In {\em ICCV}, 2013.

\bibitem{Everingham2010Pascal}
Mark Everingham, Luc Van~Gool, Christopher~K.I. Williams, John Winn, and Andrew
  Zisserman.
\newblock {The Pascal Visual Object Classes (VOC) Challenge}.
\newblock {\em International Journal of Computer Vision}, 88(2):303--338, 2010.

\bibitem{gong2019graphonomy}
Ke Gong, Yiming Gao, Xiaodan Liang, Xiaohui Shen, Meng Wang, and Liang Lin.
\newblock Graphonomy: Universal human parsing via graph transfer learning.
\newblock In {\em CVPR}, 2019.

\bibitem{gong2018instance}
Ke Gong, Xiaodan Liang, Yicheng Li, Yimin Chen, Ming Yang, and Liang Lin.
\newblock Instance-level human parsing via part grouping network.
\newblock In {\em ECCV}, 2018.

\bibitem{he2017mask}
Kaiming He, Georgia Gkioxari, Piotr Doll{\'a}r, and Ross Girshick.
\newblock {Mask R-CNN}.
\newblock In {\em ICCV}, 2017.

\bibitem{He2016ResNet}
Kaiming {He}, Xiangyu {Zhang}, Shaoqing {Ren}, and Jian {Sun}.
\newblock {Deep Residual Learning for Image Recognition}.
\newblock In {\em CVPR}, 2016.

\bibitem{Hou2020Real}
Rui {Hou}, Jie {Li}, Arjun {Bhargava}, Allan {Raventos}, Vitor {Guizilini},
  Chao {Fang}, Jerome {Lynch}, and Adrien {Gaidon}.
\newblock {Real-Time Panoptic Segmentation From Dense Detections}.
\newblock In {\em CVPR}, 2020.

\bibitem{jiang2018cnn}
Yalong Jiang and Zheru Chi.
\newblock {A CNN Model for Semantic Person Part Segmentation With Capacity
  Optimization}.
\newblock {\em IEEE Transactions on Image Processing}, 28(5):2465--2478, 2018.

\bibitem{jiang2019cnn}
Yalong Jiang and Zheru Chi.
\newblock {A CNN Model for Human Parsing Based on Capacity Optimization}.
\newblock {\em Applied Sciences}, 9(7):1330, 2019.

\bibitem{Kirillov2019PanopticFPN}
Alexander Kirillov, Ross Girshick, Kaiming He, and Piotr Doll{\'{a}}r.
\newblock {Panoptic Feature Pyramid Networks}.
\newblock In {\em CVPR}, 2019.

\bibitem{Kirillov2019PS}
Alexander Kirillov, Kaiming He, Ross Girshick, Carsten Rother, and Piotr
  Doll{\'a}r.
\newblock {Panoptic Segmentation}.
\newblock In {\em CVPR}, 2019.

\bibitem{ladicky2013human}
Lubor Ladicky, Philip~HS Torr, and Andrew Zisserman.
\newblock Human pose estimation using a joint pixel-wise and part-wise
  formulation.
\newblock In {\em CVPR}, 2013.

\bibitem{Lazarow2020OCFusion}
Justin {Lazarow}, Kwonjoon {Lee}, Kunyu {Shi}, and Zhuowen {Tu}.
\newblock {Learning Instance Occlusion for Panoptic Segmentation}.
\newblock In {\em CVPR}, 2020.

\bibitem{li2018multi}
Jianshu Li, Jian Zhao, Yunpeng Chen, Sujoy Roy, Shuicheng Yan, Jiashi Feng, and
  Terence Sim.
\newblock Multi-human parsing machines.
\newblock In {\em ACM Multimedia Conference on Multimedia Conference}, 2018.

\bibitem{li2019self}
Peike {Li}, Yunqiu {Xu}, Yunchao {Wei}, and Yi {Yang}.
\newblock {Self-Correction for Human Parsing}.
\newblock {\em IEEE Transactions on Pattern Analysis and Machine Intelligence},
  2020.

\bibitem{li2017holistic}
Qizhu Li, Anurag Arnab, and Philip~HS Torr.
\newblock Holistic, instance-level human parsing.
\newblock In {\em BMVC}, 2017.

\bibitem{Li2020Unifying}
Qizhu {Li}, Xiaojuan {Qi}, and Philip H.~S. {Torr}.
\newblock {Unifying Training and Inference for Panoptic Segmentation}.
\newblock In {\em CVPR}, 2020.

\bibitem{li2019attention}
Yanwei Li, Xinze Chen, Zheng Zhu, Lingxi Xie, Guan Huang, Dalong Du, and
  Xingang Wang.
\newblock Attention-guided unified network for panoptic segmentation.
\newblock In {\em CVPR}, 2019.

\bibitem{Liang2020polytransform}
Justin Liang, Namdar Homayounfar, Wei-Chiu Ma, Yuwen Xiong, Rui Hu, and Raquel
  Urtasun.
\newblock {PolyTransform: Deep Polygon Transformer for Instance Segmentation}.
\newblock In {\em CVPR}, 2020.

\bibitem{liang2018look}
Xiaodan Liang, Ke Gong, Xiaohui Shen, and Liang Lin.
\newblock Look into person: Joint body parsing \& pose estimation network and a
  new benchmark.
\newblock {\em IEEE Transactions on Pattern Analysis and Machine Intelligence},
  41(4):871--885, 2018.

\bibitem{liang2016semantic}
Xiaodan Liang, Xiaohui Shen, Jiashi Feng, Liang Lin, and Shuicheng Yan.
\newblock {Semantic Object Parsing with Graph LSTM}.
\newblock In {\em ECCV}, 2016.

\bibitem{liang2015human}
Xiaodan Liang, Chunyan Xu, Xiaohui Shen, Jianchao Yang, Si Liu, Jinhui Tang,
  Liang Lin, and Shuicheng Yan.
\newblock Human parsing with contextualized convolutional neural network.
\newblock In {\em ICCV}, 2015.

\bibitem{lin2019face}
Jinpeng Lin, Hao Yang, Dong Chen, Ming Zeng, Fang Wen, and Lu Yuan.
\newblock {Face Parsing with RoI Tanh-Warping}.
\newblock In {\em CVPR}, 2019.

\bibitem{lin2019cross}
Kevin Lin, Lijuan Wang, Kun Luo, Yinpeng Chen, Zicheng Liu, and Ming-Ting Sun.
\newblock {Cross-Domain Complementary Learning Using Pose for Multi-Person Part
  Segmentation}.
\newblock {\em IEEE Transactions on Circuits and Systems for Video Technology},
  2020.

\bibitem{Lin2014COCO}
Tsung-Yi Lin, Michael Maire, Serge Belongie, James Hays, Pietro Perona, Deva
  Ramanan, Piotr Doll{\'a}r, and C.~Lawrence Zitnick.
\newblock {Microsoft COCO: Common Objects in Context}.
\newblock In {\em ECCV}, 2014.

\bibitem{liu2019end}
Huanyu Liu, Chao Peng, Changqian Yu, Jingbo Wang, Xu Liu, Gang Yu, and Wei
  Jiang.
\newblock An end-to-end network for panoptic segmentation.
\newblock In {\em CVPR}, 2020.

\bibitem{liu2018cross}
Si Liu, Yao Sun, Defa Zhu, Guanghui Ren, Yu Chen, Jiashi Feng, and Jizhong Han.
\newblock {Cross-domain Human Parsing via Adversarial Feature and Label
  Adaptation}.
\newblock In {\em AAAI}, 2018.

\bibitem{liu2015survey}
Zhao Liu, Jianke Zhu, Jiajun Bu, and Chun Chen.
\newblock A survey of human pose estimation: The body parts parsing based
  methods.
\newblock {\em Journal of Visual Communication and Image Representation},
  32:10--19, 2015.

\bibitem{luo2013pedestrian}
Ping Luo, Xiaogang Wang, and Xiaoou Tang.
\newblock Pedestrian parsing via deep decompositional network.
\newblock In {\em ICCV}, 2013.

\bibitem{luo2018trusted}
Xianghui Luo, Zhuo Su, Jiaming Guo, Gengwei Zhang, and Xiangjian He.
\newblock {Trusted Guidance Pyramid Network for Human Parsing}.
\newblock In {\em ACM Multimedia Conference on Multimedia Conference}, 2018.

\bibitem{michieli2020gmnet}
Umberto Michieli, Edoardo Borsato, Luca Rossi, and Pietro Zanuttigh.
\newblock {GMNet: Graph Matching Network for Large Scale Part Semantic
  Segmentation in the Wild}.
\newblock In {\em ECCV}, 2020.

\bibitem{mohan2020efficientps}
Rohit Mohan and Abhinav Valada.
\newblock {EfficientPS: Efficient panoptic segmentation}.
\newblock {\em International Journal of Computer Vision}, 2021.

\bibitem{mottaghi14pascalcontext}
Roozbeh Mottaghi, Xianjie Chen, Xiaobai Liu, Nam-Gyu Cho, Seong-Whan Lee, Sanja
  Fidler, Raquel Urtasun, and Alan Yuille.
\newblock {The Role of Context for Object Detection and Semantic Segmentation
  in the Wild}.
\newblock In {\em CVPR}, 2014.

\bibitem{mmsegmentation}
OpenMMLab.
\newblock {\em MMSegmentation}.
\newblock \url{https://github.com/open-mmlab/mmsegmentation} (accessed October
  5, 2020).

\bibitem{porzi2019seamless}
Lorenzo Porzi, Samuel~Rota Bulo, Aleksander Colovic, and Peter Kontschieder.
\newblock Seamless scene segmentation.
\newblock In {\em CVPR}, 2019.

\bibitem{qiao2020detectors}
Siyuan Qiao, Liang-Chieh Chen, and Alan Yuille.
\newblock {DetectoRS: Detecting Objects with Recursive Feature Pyramid and
  Switchable Atrous Convolution}.
\newblock {\em arXiv preprint arXiv:2006.02334}, 2020.

\bibitem{ruan2019devil}
Tao Ruan, Ting Liu, Zilong Huang, Yunchao Wei, Shikui Wei, and Yao Zhao.
\newblock Devil in the details: Towards accurate single and multiple human
  parsing.
\newblock In {\em AAAI}, 2019.

\bibitem{tu2005image}
Zhuowen Tu, Xiangrong Chen, Alan~L Yuille, and Song-Chun Zhu.
\newblock Image parsing: Unifying segmentation, detection, and recognition.
\newblock {\em International Journal of Computer Vision}, 63(2):113--140, 2005.

\bibitem{Wang2020HRNet}
Jingdong Wang, Ke Sun, Tianheng Cheng, Borui Jiang, Chaorui Deng, Yang Zhao,
  Dong Liu, Yadong Mu, Mingkui Tan, Xinggang Wang, et~al.
\newblock {Deep High-Resolution Representation Learning for Visual
  Recognition}.
\newblock {\em IEEE Transactions on Pattern Analysis and Machine Intelligence},
  2020.

\bibitem{wang2015joint}
Peng Wang, Xiaohui Shen, Zhe Lin, Scott Cohen, Brian Price, and Alan~L Yuille.
\newblock Joint object and part segmentation using deep learned potentials.
\newblock In {\em ICCV}, 2015.

\bibitem{xiong2019upsnet}
Yuwen Xiong, Renjie Liao, Hengshuang Zhao, Rui Hu, Min Bai, Ersin Yumer, and
  Raquel Urtasun.
\newblock {UPSNet: A unified panoptic segmentation network}.
\newblock In {\em CVPR}, 2019.

\bibitem{yang2019parsing}
Lu Yang, Qing Song, Zhihui Wang, and Ming Jiang.
\newblock {Parsing R-CNN for Instance-Level Human Analysis}.
\newblock In {\em CVPR}, 2019.

\bibitem{yang2019deeperlab}
Tien-Ju Yang, Maxwell~D Collins, Yukun Zhu, Jyh-Jing Hwang, Ting Liu, Xiao
  Zhang, Vivienne Sze, George Papandreou, and Liang-Chieh Chen.
\newblock {DeeperLab: Single-shot image parser}.
\newblock {\em arXiv preprint arXiv:1902.05093}, 2019.

\bibitem{Yang2020SOGNet}
Yibo Yang, Hongyang Li, Xia Li, Qijie Zhao, Jianlong Wu, and Zhouchen Lin.
\newblock {{SOGNet}: Scene Overlap Graph Network for Panoptic Segmentation}.
\newblock In {\em AAAI}, 2020.

\bibitem{yao2012describing}
Jian Yao, Sanja Fidler, and Raquel Urtasun.
\newblock Describing the scene as a whole: joint object detection.
\newblock In {\em CVPR}, 2012.

\bibitem{Yuan2020ocr}
Yuhui Yuan, Xilin Chen, and Jingdong Wang.
\newblock {Object-Contextual Representations for Semantic Segmentation}.
\newblock In {\em ECCV}, 2020.

\bibitem{Zhang2020PyTorchEncoding}
Hang Zhang.
\newblock {\em PyTorch-Encoding}.
\newblock \url{https://github.com/zhanghang1989/PyTorch-Encoding} (accessed
  October 14, 2020).

\bibitem{zhang2020resnest}
Hang Zhang, Chongruo Wu, Zhongyue Zhang, Yi Zhu, Zhi Zhang, Haibin Lin, Yue
  Sun, Tong He, Jonas Muller, R. Manmatha, Mu Li, and Alexander Smola.
\newblock {ResNeSt: Split-Attention Networks}.
\newblock {\em arXiv preprint arXiv:2004.08955}, 2020.

\bibitem{zhao2018understanding}
Jian Zhao, Jianshu Li, Yu Cheng, Terence Sim, Shuicheng Yan, and Jiashi Feng.
\newblock Understanding humans in crowded scenes: Deep nested adversarial
  learning and a new benchmark for multi-human parsing.
\newblock In {\em ACM Multimedia Conference on Multimedia Conference}, 2018.

\bibitem{Zhao2019BSANet}
Yifan {Zhao}, Jia {Li}, Yu {Zhang}, and Yonghong {Tian}.
\newblock {Multi-Class Part Parsing With Joint Boundary-Semantic Awareness}.
\newblock In {\em ICCV}, 2019.

\end{thebibliography}
}

\appendix 

\section*{Appendix}
We provide the following information as supplementary material:
\begin{enumerate}[noitemsep]
    \item Elaborate implementation details for the methods on the subtasks used for generating the baselines on part-aware panoptic segmentation, in Appendix \ref{sec:implementation}.
    \item Detailed per-class results of the state-of-the-art baselines, in Appendix \ref{sec:results}.
    \item Details about the annotation procedure for Cityscapes Panoptic Parts, in Appendix \ref{sec:annotations}.
    \item Additional qualitative examples of annotations of the presented datasets, and predictions of the baselines, in Appendix \ref{sec:images}.
\end{enumerate}

\noindent Code and data: \url{https://github.com/tue-mps/panoptic_parts}.

\section{Implementation details baselines}
\label{sec:implementation}
For reproducibility, we provide more information about the implementation details of the methods used to generate baselines for part-level panoptic segmentation (PPS). 

To guarantee the state-of-the-art performance of the used methods, we use existing trained models when possible. If there is no trained model available, we train a network ourselves.

\subsection{Citycapes Panoptic Parts}
\label{subsec:CPP}
The baseline results for Cityscapes Panoptic Parts (CPP), as presented in Section~\ref{sec:experiments:baselines:cpp} and Table~\ref{tab:experiments:baselines}, are generated by merging results from existing methods. Below, we describe, for each method in Table~\ref{tab:experiments:baselines}, how we acquire the results of these methods.

For CPP, all models are trained only on the images in the Cityscapes \texttt{train} split~\cite{Cordts2016Cityscapes}, unless otherwise indicated.

\subsubsection{Panoptic segmentation}
For panoptic segmentation results on Cityscapes Pascal Parts, we use both single-network panoptic segmentation methods, and results generated by merging semantic and instance segmentation methods following the heuristics presented in~\cite{Kirillov2019PS}.

\paragraph{EfficientPS.} For state-of-the-art panoptic segmentation method EfficientPS~\cite{mohan2020efficientps}, the predictions for the Cityscapes \texttt{val} were generously provided to us by the authors of the work. During inference, multi-scale testing is applied.

\paragraph{UPSNet.} The results for UPSNet~\cite{xiong2019upsnet} were generated using the official code repository. Specifically, we run inference using the trained model with a ResNet-50 backbone~\cite{He2016ResNet}.

\paragraph{HRNet-OCR \& PolyTransform.} To get further state-of-the-art panoptic segmentation results, we fuse the state-of-the-art semantic segmentation and instance segmentation results from HRNet-OCR~\cite{Yuan2020ocr} and PolyTransform~\cite{Liang2020polytransform}, respectively. 

The HRNet-OCR~\cite{Yuan2020ocr} results are generated using a trained model from the official code repository. Specifically, we pick the model with a HRNetv2-W48~\cite{Wang2020HRNet} backbone, without using test-time augmentations.

For PolyTransform~\cite{Liang2020polytransform}, the results on the Cityscapes \texttt{val} split were generously provided to us by the authors of the paper. We note that this instance segmentation model is pre-trained on the COCO dataset~\cite{Lin2014COCO}.

\paragraph{DeepLabv3+ \& Mask R-CNN.} The results for both DeepLabv3+~\cite{Chen2018deeplabv3plus} and Mask R-CNN~\cite{He2016ResNet} are generated using existing trained models from official code repositories.

For DeepLabv3+~\cite{Chen2018deeplabv3plus}, we select a model with an Xception-65 backbone~\cite{Chollet2017Xception}, and we do not use test-time augmentations.

For Mask R-CNN~\cite{he2017mask}, we generate the results using a trained model with a ResNet-50 backbone~\cite{He2016ResNet}, pre-trained on the COCO dataset~\cite{Lin2014COCO}. 

\subsubsection{Part segmentation}

To generate the part-level segmentation predictions, which are required to generate the PPS predictions, we use two networks to perform the part segmentation task: state-of-the-art BSANet \cite{Zhao2019BSANet} and commonly used DeepLabv3+ \cite{Chen2018deeplabv3plus}. Since the part-level annotations of Cityscapes dataset \cite{Cordts2016Cityscapes} are newly proposed by us, there are no trained models available, so we train two networks with settings similar to those proposed in~\cite{Zhao2019BSANet, Chen2018deeplabv3plus}.

\paragraph{BSANet.}

We use the official repository provided by the authors of BSANet \cite{Zhao2019BSANet} and keep most of the training settings same as in the official version. We change the crop size for training to 512$\times$1024 and the number of output classes according to our part label definition. During the training, an SGD optimizer is applied with polynomial learning rate decay. We set base learning rate to 0.01, decay power to 0.9, and weight decay to 4e-5. We train the network for 100 epochs with batch size of 3. 

\paragraph{DeepLabv3+.}
We use the popular \textit{mmsegmentation}~\cite{mmsegmentation} respository to train a DeepLabv3+~ \cite{Chen2018deeplabv3plus} model. We use one of default configurations provided by the repository, with a ResNet-50 backbone~\cite{He2016ResNet}. The crop size during training is set to 769$\times$769. Again, we use an SGD optimizer with polynomial learning rate decay. The base learning rate, decay power, and weight decay are set to 0.01, 0.9, and 5e-4, respectively. We train the network for 40k iterations with batch size of 4. 

\subsection{Pascal Panoptic Parts}
For the baselines results for Pascal Panoptic Parts (PPP), as presented in Section~\ref{sec:experiments:baselines:ppp} and Table~\ref{tab:experiments:baselines}, we also provide further implementation details.

Again, to guarantee state-of-the-art performance performance on the subtasks, and to facilitate reproducibility, we use publicly available trained models when possible. Methods for semantic segmentation are trained on 59 classes of Pascal-Context~\cite{mottaghi14pascalcontext}; part segmentation is trained on 58 part-level classes from Pascal-Parts~\cite{chen2014detect}, as defined in~\cite{Zhao2019BSANet}. We train the instance segmentation models on the 20 \textit{things} classes from our PPP dataset. We evaluate all methods on our annotations for the PPP \texttt{validation} set, using the same class definition as for training.

As explained in the main manuscript, there are discrepancies between the various different annotation sets for Pascal VOC 2010~\cite{Everingham2010Pascal}. In our Pascal Panoptic Part dataset, we resolve such conflicts, and generate consistent annotations for multiple levels of abstraction. As a result, the absolute scores on the other annotation sets for Pascal VOC 2010 are not directly comparable with results on our PPP.

\subsubsection{Panoptic segmentation}
Panoptic segmentation results for PPP are generated by fusing semantic segmentation and instance segmentation results using the heuristics described in~\cite{Kirillov2019PS}. 

\paragraph{DeepLabv3-ResNeSt269 \& DetectoRS.}
State-of-the-art results for semantic segmentation are generated using a DeepLabv3 model~\cite{chen2017deeplabv3} with a ResNeSt-269 backbone~\cite{zhang2020resnest}. For this, we use a trained model from the \textit{PyTorch-Encoding} repository~\cite{Zhang2020PyTorchEncoding}, which is the official semantic segmentation repository for the paper introducing ResNeSt~\cite{zhang2020resnest}.

For instance segmentation, there is no trained model available, so we train a model using the commonly used \textit{mmdetection} repository~\cite{chen2019mmdetection}. Specifically, to generate state-of-the-art results, we train a DetectoRS~\cite{qiao2020detectors} model with a HTC-ResNet-50 backbone~\cite{Chen2019HTC, He2016ResNet}. We do not use auxiliary semantic labels for training. For training, we use a batch size of 4, a learning rate of 0.0025 and weight decay of 1e-4. We train for 24 epochs and decrease the learning rate by a factor of 10 after 18 epochs.

\paragraph{DeepLabv3 \& Mask R-CNN.}
To set another reference for semantic segmentation, we also train a DeepLabv3~\cite{chen2017deeplabv3} model with a ResNet-50 backbone~\cite{He2016ResNet}. Again, we use the \textit{PyTorch-Encoding} repository~\cite{Zhang2020PyTorchEncoding}. This model is trained for 80 epochs, with a batch size of 16, a weight decay of 1e-4, and a polynomial learning rate schedule with an initial learning rate of 0.001 and a decay of 0.9.

For instance segmentation results on Mask R-CNN~\cite{he2017mask}, we train another model using the \textit{mmdetection} repository~\cite{chen2019mmdetection}. Specifically, we train Mask R-CNN with a ResNet-50 backbone~\cite{He2016ResNet}. Again, we use a batch size of 4, a learning rate of 0.0025 and weight decay of 1e-4. We train for 24 epochs and reduce the learning rate by a factor of 10 after 18 epochs.

\subsubsection{Part segmentation}

To generate part segmentation results for PPP, we use the same two networks as for CPP (See Section \ref{subsec:CPP}).

\paragraph{BSANet.}

The \texttt{validation} set prediction samples of \cite{Zhao2019BSANet} are made publicly available by the authors. Therefore, we directly use their predictions in our experiments.

\paragraph{DeepLabv3+.}
To generate DeepLabv3+~\cite{Chen2018deeplabv3plus} predictions for PPP, we also train a network using the \textit{mmsegmentation} repository~\cite{mmsegmentation}. As in Section \ref{subsec:CPP}, we use one of the default configurations provided by \textit{mmsegmentation}, with a ResNet-50 backbone~\cite{He2016ResNet}. Again, we use an SGD optimizer with polynomial learning rate decay. The base learning rate, decay power, and weight decay are set to 0.004, 0.9, and 1e-4, respectively. We train the network for 40k iterations with batch size of 8.

\section{Detailed results}
\label{sec:results}
In this section, we provide detailed per-class results for the highest scoring baselines on the PPS task, for both the Cityscapes Panoptic Part and Pascal Panoptic Part datasets. Similarly to the original Panoptic Quality~\cite{Kirillov2019PS}, the Part-aware Panoptic Quality (PartPQ) can be split into two parts, the Segmentation Quality (PartSQ) and the Recognition Quality (PartRQ). Spefically

\begin{equation}
\textrm{PartPQ} = \textrm{PartSQ} \times \textrm{PartRQ},
\label{equ:pq-parts-supp}
\end{equation}
where 
\begin{equation}
\textrm{PartSQ} = \frac{\sum_{(p,g) \in \textit{TP}}\textrm{IOU\textsubscript{p}}(p,g)}{|\textit{TP}|};
\label{equ:sq-parts-supp}
\end{equation}
\begin{equation}
\textrm{PartRQ} = \frac{|\textit{TP}|}{|\textit{TP}| + \frac{1}{2}|\textit{FP}|+ \frac{1}{2}|\textit{FN}|}.
\label{equ:rq-parts-supp}
\end{equation}

In Table \ref{tab:cs_detailed}, we show the per-class results of the highest scoring baseline on Cityscapes Panoptic Parts, for both panoptic segmentation and part-level panoptic segmentation. The Pascal Panoptic Parts results are provided in Table \ref{tab:pascal_detailed}. A few things can be noted. 1) First of all, the performance for scene-level classes without parts is identical for PQ and PartPQ. This is as expected, as the task definition is the same for those classes, \ie, if no part classes are defined, we just predict the scene-level segment. 2) Secondly, for the classes with parts, the scores for PartPQ are consistently lower than for PQ. This is also explainable, as these segments have to be segmented further into parts, and are evaluated on multi-class IOU instead of instance-level IOU. 3) Thirdly, note that, although we expect a change in PartSQ for scene-level classes with parts, we do not expect the PartRQ to change, as the \textit{TP}, \textit{FP} and \textit{FN} are evaluated on scene-level segments, as in the original PQ. However, in both datasets, there are ground-truth segments that are labeled on instance-level, but not on part-level, due to ambiguity or because the instances are so small that parts could not be distinguished. In these rare cases, these segments are ignored during evaluation, causing the PartRQ to differ slightly from the RQ (\eg for \textit{person}, in Table \ref{tab:cs_detailed}).

\begin{table}
\centering
\begin{adjustbox}{width=1\linewidth}
\begin{tabular}{l|ccc|ccc}
Class          & PQ   & SQ   & RQ   & PartPQ & PartSQ & PartRQ \\ \hline
road          & 98.3 & 98.4 & 99.9 & 98.3   & 98.4   & 99.9   \\
sidewalk      & 80.4 & 86.6 & 92.7 & 80.4   & 86.6   & 92.7   \\
building      & 90.3 & 91.0 & 99.3 & 90.3   & 91.0   & 99.3   \\
wall          & 37.7 & 76.8 & 49.2 & 37.7   & 76.8   & 49.2   \\
fence         & 44.0 & 76.0 & 57.9 & 44.0   & 76.0   & 57.9   \\
pole          & 63.4 & 71.0 & 89.3 & 63.4   & 71.0   & 89.3   \\
traffic light & 58.5 & 73.3 & 79.8 & 58.5   & 73.3   & 79.8   \\
traffic sign  & 74.5 & 80.9 & 92.1 & 74.5   & 80.9   & 92.1   \\
vegetation    & 90.9 & 91.6 & 99.2 & 90.9   & 91.6   & 99.2   \\
terrain       & 41.1 & 77.0 & 53.4 & 41.1   & 77.0   & 53.4   \\
sky           & 88.8 & 92.6 & 95.9 & 88.8   & 92.6   & 95.9   \\
person*$^\dagger$ & 60.8 & 79.3 & 76.6 & 44.1   & 57.8   & 76.2   \\
rider*$^\dagger$  & 58.0 & 75.6 & 76.8 & 45.3   & 59.6   & 76.0   \\
car*$^\dagger$    & 71.8 & 85.5 & 84.0 & 53.3   & 63.5   & 84.0   \\
truck*$^\dagger$  & 57.1 & 89.0 & 64.2 & 36.4   & 56.7   & 64.2   \\
bus*$^\dagger$    & 73.5 & 91.1 & 80.7 & 49.7   & 61.6   & 80.7   \\
train*        & 67.9 & 85.9 & 79.1 & 67.9   & 85.9   & 79.1   \\
motorcycle*   & 50.2 & 77.0 & 65.2 & 50.2   & 77.0   & 65.2   \\
bicycle*      & 51.6 & 74.4 & 69.3 & 51.6   & 74.4   & 69.3   \\ \hline
Things        & 61.3 & 82.2 & 74.5 & 49.8   & 67.1   & 74.3   \\
Stuff         & 69.8 & 83.2 & 82.6 & 69.8   & 83.2   & 82.6   \\ \hline
Parts         & 64.2 & 84.1 & 76.4 & 45.8   & 59.9   & 76.2   \\
No Parts      & 67.0 & 82.3 & 80.2 & 67.0   & 82.3   & 80.2   \\ \hline
All           & 66.2 & 82.8 & 79.2 & 61.4   & 76.4   & 79.1   
\end{tabular}
\end{adjustbox}
\caption{Detailed results for the highest scoring baseline on the Cityscapes Panoptic Parts dataset (HRNet-OCR \& PolyTransform \& BSANet~\cite{Yuan2020ocr,Liang2020polytransform,Zhao2019BSANet}). We report both scores for PQ and PartPQ, for the panoptic segmentation and part-level panoptic segmentation task, respectively. * Indicates things classes; $^\dagger$ indicates classes with parts.}
\label{tab:cs_detailed}
\end{table}

\begin{table}
\centering
\begin{adjustbox}{width=0.865\linewidth}
\begin{tabular}{l|ccc|ccc}
Class                   & PQ   & SQ   & RQ   & PartPQ & PartSQ & PartRQ \\ \hline
aeroplane*$^\dagger$   & 69.4 & 82.5 & 84.1 & 45.4   & 53.5   & 85.0   \\
bag                    & 24.4 & 73.6 & 33.1 & 24.4   & 73.6   & 33.1   \\
bed                    & 4.1  & 62.1 & 6.5  & 4.1    & 62.1   & 6.5    \\
bedclothes             & 21.6 & 81.6 & 26.5 & 21.6   & 81.6   & 26.5   \\
bench                  & 7.5  & 61.6 & 12.2 & 7.5    & 61.6   & 12.2   \\
bicycle*$^\dagger$     & 59.1 & 75.9 & 77.9 & 54.8   & 70.4   & 77.9   \\
bird*$^\dagger$        & 68.2 & 84.4 & 80.8 & 44.6   & 54.6   & 81.7   \\
boat*                  & 50.8 & 77.7 & 65.3 & 50.8   & 77.7   & 65.3   \\
book                   & 26.9 & 70.7 & 38.1 & 26.9   & 70.7   & 38.1   \\
bottle*$^\dagger$      & 56.2 & 85.6 & 65.7 & 42.7   & 64.8   & 65.8   \\
building               & 47.9 & 78.6 & 60.9 & 47.9   & 78.6   & 60.9   \\
bus*$^\dagger$         & 79.5 & 92.5 & 85.9 & 64.2   & 73.9   & 86.9   \\
cabinet                & 26.7 & 76.5 & 34.9 & 26.7   & 76.5   & 34.9   \\
car*$^\dagger$         & 69.0 & 86.7 & 79.5 & 45.5   & 56.9   & 80.1   \\
cat*$^\dagger$         & 79.9 & 89.4 & 89.4 & 60.6   & 67.8   & 89.4   \\
ceiling                & 48.3 & 80.1 & 60.4 & 48.3   & 80.1   & 60.4   \\
chair*                 & 38.3 & 77.5 & 49.4 & 38.3   & 77.5   & 49.4   \\
cloth                  & 9.3  & 71.9 & 12.9 & 9.3    & 71.9   & 12.9   \\
computer               & 25.1 & 69.9 & 35.8 & 25.1   & 69.9   & 35.8   \\
cow*$^\dagger$         & 59.0 & 84.4 & 69.9 & 45.6   & 65.2   & 70.0   \\
cup                    & 21.6 & 73.2 & 29.5 & 21.6   & 73.2   & 29.5   \\
curtain                & 38.9 & 78.2 & 49.8 & 38.9   & 78.2   & 49.8   \\
dog*$^\dagger$         & 75.1 & 87.2 & 86.1 & 56.3   & 65.3   & 86.3   \\
door                   & 15.4 & 72.4 & 21.2 & 15.4   & 72.4   & 21.2   \\
fence                  & 25.1 & 70.1 & 35.8 & 25.1   & 70.1   & 35.8   \\
floor                  & 51.7 & 82.9 & 62.4 & 51.7   & 82.9   & 62.4   \\
flower                 & 11.4 & 69.0 & 16.5 & 11.4   & 69.0   & 16.5   \\
food                   & 20.8 & 73.0 & 28.4 & 20.8   & 73.0   & 28.4   \\
grass                  & 61.2 & 84.0 & 72.8 & 61.2   & 84.0   & 72.8   \\
ground                 & 40.5 & 80.0 & 50.7 & 40.5   & 80.0   & 50.7   \\
horse*$^\dagger$       & 62.9 & 80.1 & 78.5 & 51.3   & 65.0   & 78.9   \\
keyboard               & 34.1 & 68.8 & 49.5 & 34.1   & 68.8   & 49.5   \\
light                  & 26.4 & 70.7 & 37.4 & 26.4   & 70.7   & 37.4   \\
motorbike*$^\dagger$   & 67.5 & 81.2 & 83.2 & 63.7   & 76.6   & 83.2   \\
mountain               & 39.3 & 76.8 & 51.2 & 39.3   & 76.8   & 51.2   \\
mouse                  & 16.3 & 74.5 & 21.9 & 16.3   & 74.5   & 21.9   \\
person*$^\dagger$      & 65.5 & 80.8 & 81.0 & 46.6   & 57.5   & 81.1   \\
plate                  & 7.5  & 73.0 & 10.3 & 7.5    & 73.0   & 10.3   \\
platform               & 35.1 & 82.7 & 42.4 & 35.1   & 82.7   & 42.4   \\
pottedplant*$^\dagger$ & 45.5 & 77.8 & 58.5 & 38.9   & 66.5   & 58.5   \\
road                   & 44.9 & 85.9 & 52.3 & 44.9   & 85.9   & 52.3   \\
rock                   & 29.0 & 76.2 & 38.1 & 29.0   & 76.2   & 38.1   \\
sheep*$^\dagger$       & 68.1 & 85.2 & 79.9 & 56.8   & 69.7   & 81.4   \\
shelves                & 11.1 & 66.8 & 16.6 & 11.1   & 66.8   & 16.6   \\
sidewalk               & 15.9 & 70.4 & 22.5 & 15.9   & 70.4   & 22.5   \\
sign                   & 24.9 & 79.6 & 31.3 & 24.9   & 79.6   & 31.3   \\
sky                    & 82.9 & 92.6 & 89.5 & 82.9   & 92.6   & 89.5   \\
snow                   & 53.8 & 81.1 & 66.3 & 53.8   & 81.1   & 66.3   \\
sofa*                  & 47.0 & 82.3 & 57.1 & 47.0   & 82.3   & 57.1   \\
table*                 & 35.3 & 73.0 & 48.4 & 35.3   & 73.0   & 48.4   \\
track                  & 50.2 & 71.4 & 70.3 & 50.2   & 71.4   & 70.3   \\
train*                 & 76.3 & 88.4 & 86.3 & 76.3   & 88.4   & 86.3   \\
tree                   & 63.1 & 82.2 & 76.8 & 63.1   & 82.2   & 76.8   \\
truck                  & 9.9  & 77.2 & 12.8 & 9.9    & 77.2   & 12.8   \\
tvmonitor*$^\dagger$   & 65.3 & 86.2 & 75.8 & 57.6   & 76.0   & 75.8   \\
wall                   & 55.4 & 80.6 & 68.7 & 55.4   & 80.6   & 68.7   \\
water                  & 72.2 & 89.9 & 80.3 & 72.2   & 89.9   & 80.3   \\
window                 & 28.1 & 73.8 & 38.1 & 28.1   & 73.8   & 38.1   \\
wood                   & 9.8  & 74.5 & 13.2 & 9.8    & 74.5   & 13.2   \\\hline
Things                 & 61.9 & 82.9 & 74.1 & 51.1   & 69.1   & 74.4   \\
Stuff                  & 31.7 & 75.8 & 40.5 & 31.7   & 75.8   & 40.5   \\\hline
Parts                  & 66.0 & 84.0 & 78.4 & 51.6   & 65.6   & 78.8   \\
No Parts               & 33.8 & 76.3 & 42.8 & 33.8   & 76.3   & 42.8   \\\hline
All                    & 42.0 & 78.3 & 51.9 & 38.3   & 73.6   & 52.0  
\end{tabular}
\end{adjustbox}
\caption{Detailed results for the highest scoring baseline on the Pascal Panoptic Parts dataset (DeepLabv3-ResNeSt269 \& DetectoRS \& BSANet~\cite{chen2017deeplabv3, zhang2020resnest,qiao2020detectors,Zhao2019BSANet}). We report both scores for PQ and PartPQ, for the panoptic segmentation and part-level panoptic segmentation task, respectively. * Indicates things classes; $^\dagger$ indicates classes with parts.}
\label{tab:pascal_detailed}
\end{table}

\begin{table*}[t]
	\centering
	\small
	\begin{tabular}{ll}
		\toprule
		Part class & Definition\\
		\midrule
		Window & Windows, wind shields and other glass surfaces on vehicles. \\
		Wheel & All wheels and tires under vehicles (excluding spare tires on the back of vehicles). \\
		Light & Light source present on vehicles, including taxi sign. \\
		License plate & License plate on front/back of vehicles. \\
		Chassis & Part of vehicle body not belonging to above classes. \\
		Unlabeled & Ambiguous or not clearly visible regions. \\
		\bottomrule
	\end{tabular}
	\caption{\textit{Vehicle} part classes for Cityscapes Panoptic Parts.}
	\label{tab:vehicle-parts}
\end{table*}

\begin{table*}[t]
	\centering
	\small
	\begin{tabular}{ll}
		\toprule
		Part class & Definition\\
		\midrule
		Torso & Core of human body, excluding limbs and head.\\
		Head & Human head. \\
		Arm & Arms, from shoulders to hands. \\
		Leg & Legs, from hips to feet. \\
		Unlabeled & Ambiguous or not clearly visible regions. \\
		\bottomrule
	\end{tabular}
	\caption{\textit{Human} part classes Cityscapes Panoptic Parts.}
	\label{tab:human-parts}
\end{table*}

\section{Cityscapes Panoptic Parts annotations}
\label{sec:annotations}

\subsection{Part definitions}
The Cityscapes dataset~\cite{Cordts2016Cityscapes} focuses on urban scene understanding and automated driving. Adhering to that direction, we choose to annotate three important \textit{vehicle} classes,~\ie \textit{car}, \textit{truck}, \textit{bus}, and all \textit{human} classes, ~\ie \textit{person}, \textit{rider}. \textit{Vehicle} and \textit{human} categories describe semantic classes with similar parts, thus we define the same semantic parts for each of the classes in these categories. The part classes are defined in Tables~\ref{tab:vehicle-parts} and~\ref{tab:human-parts}.

\subsection{Manual labeling protocol}
The annotators of Cityscapes Panoptic Parts were asked to start annotation from background to foreground objects, and to annotate each object with part labels in the order they appear in Tables~\ref{tab:vehicle-parts} and~\ref{tab:human-parts}. The regions to be annotated were extracted using the instance masks from the original Cityscapes dataset. This way, we maintain consistency with the original dataset. Very small instances, ambiguous regions, or indistinguishable parts are not annotated with part labels, so they stay annotated on scene-level only. 
Moreover, it is not necessary for object instances to contain all part classes.
These aspects are taken into consideration by the PartPQ metric.

More details on the annotation procedure and the protocol can be found on \url{https://github.com/tue-mps/panoptic_parts}.

\section{Qualitative examples}
\label{sec:images}
\begin{figure*}[t]
	\centering
    	\includegraphics[width=0.24\linewidth]{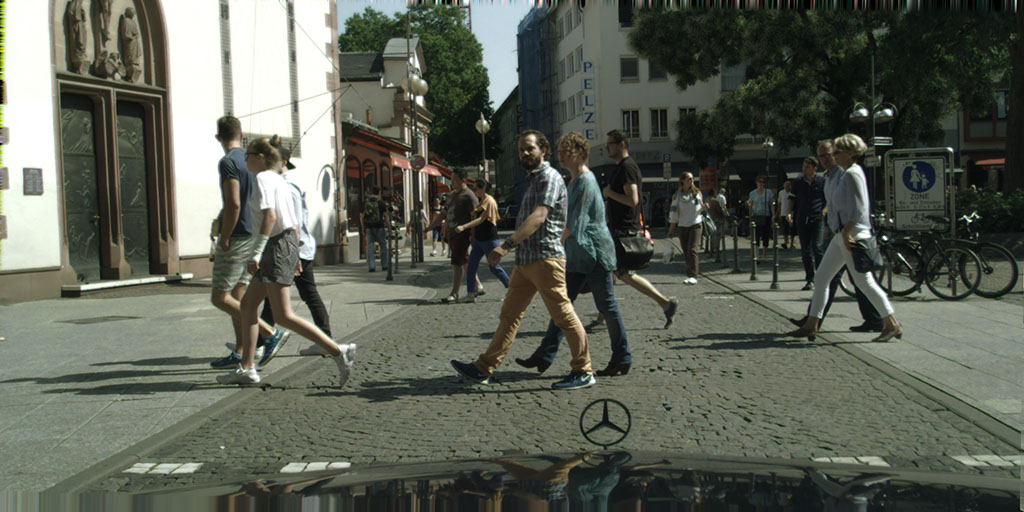}
    	\includegraphics[width=0.24\linewidth]{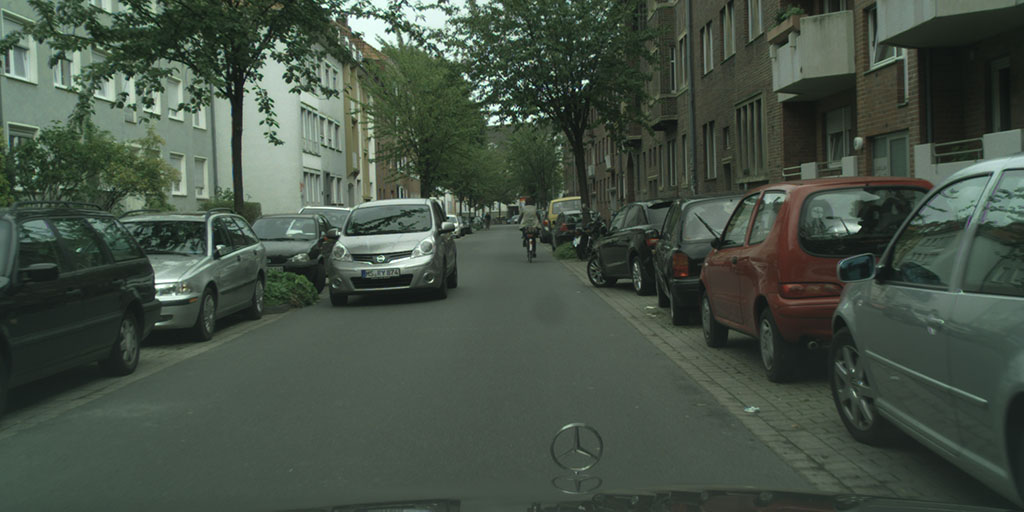}
    	\includegraphics[width=0.24\linewidth]{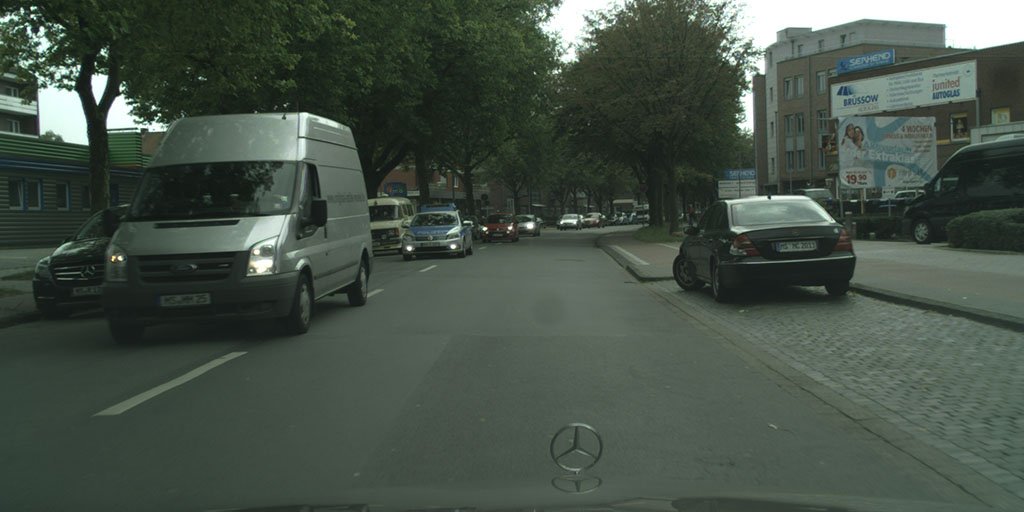}
    	\includegraphics[width=0.24\linewidth]{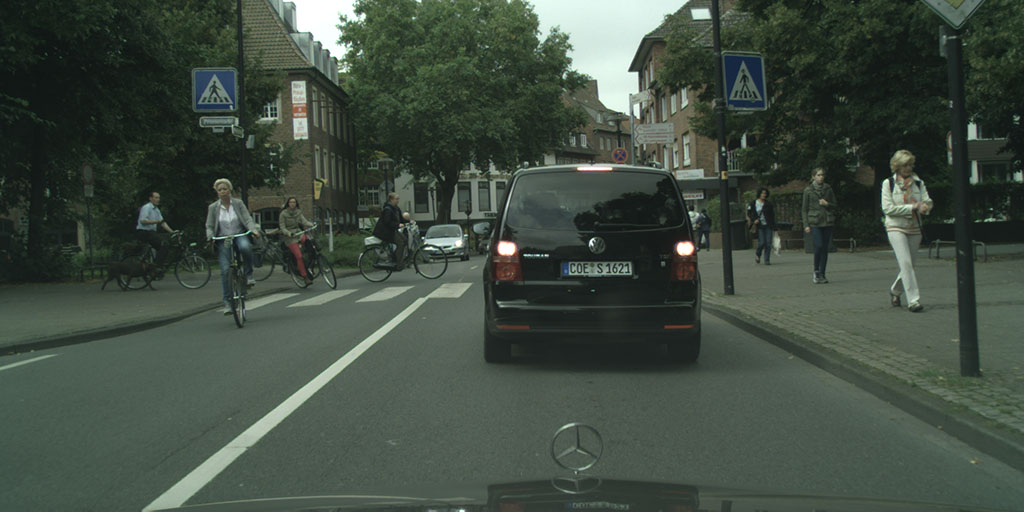}\\
		\includegraphics[width=0.24\linewidth]{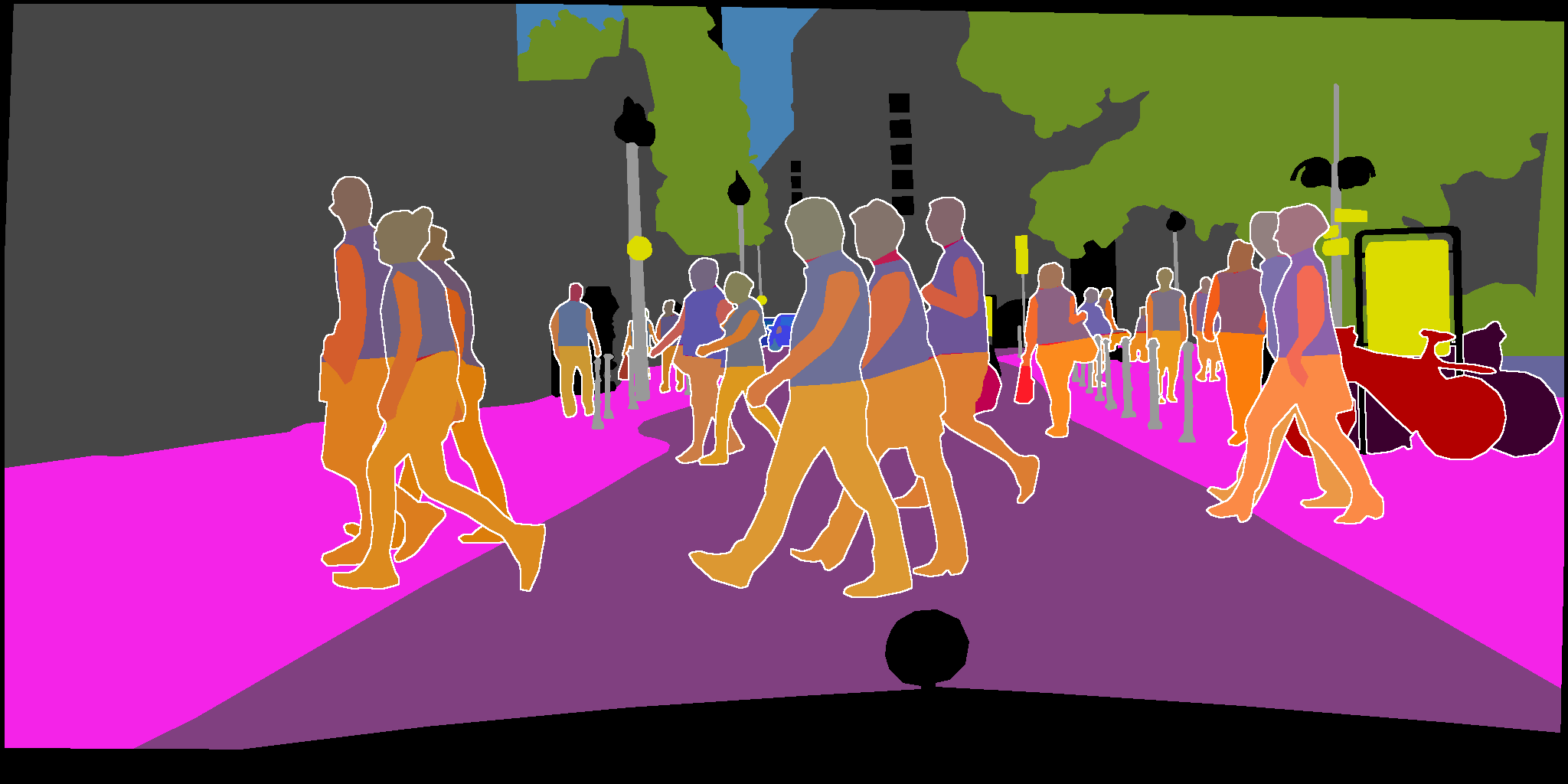}
		\includegraphics[width=0.24\linewidth]{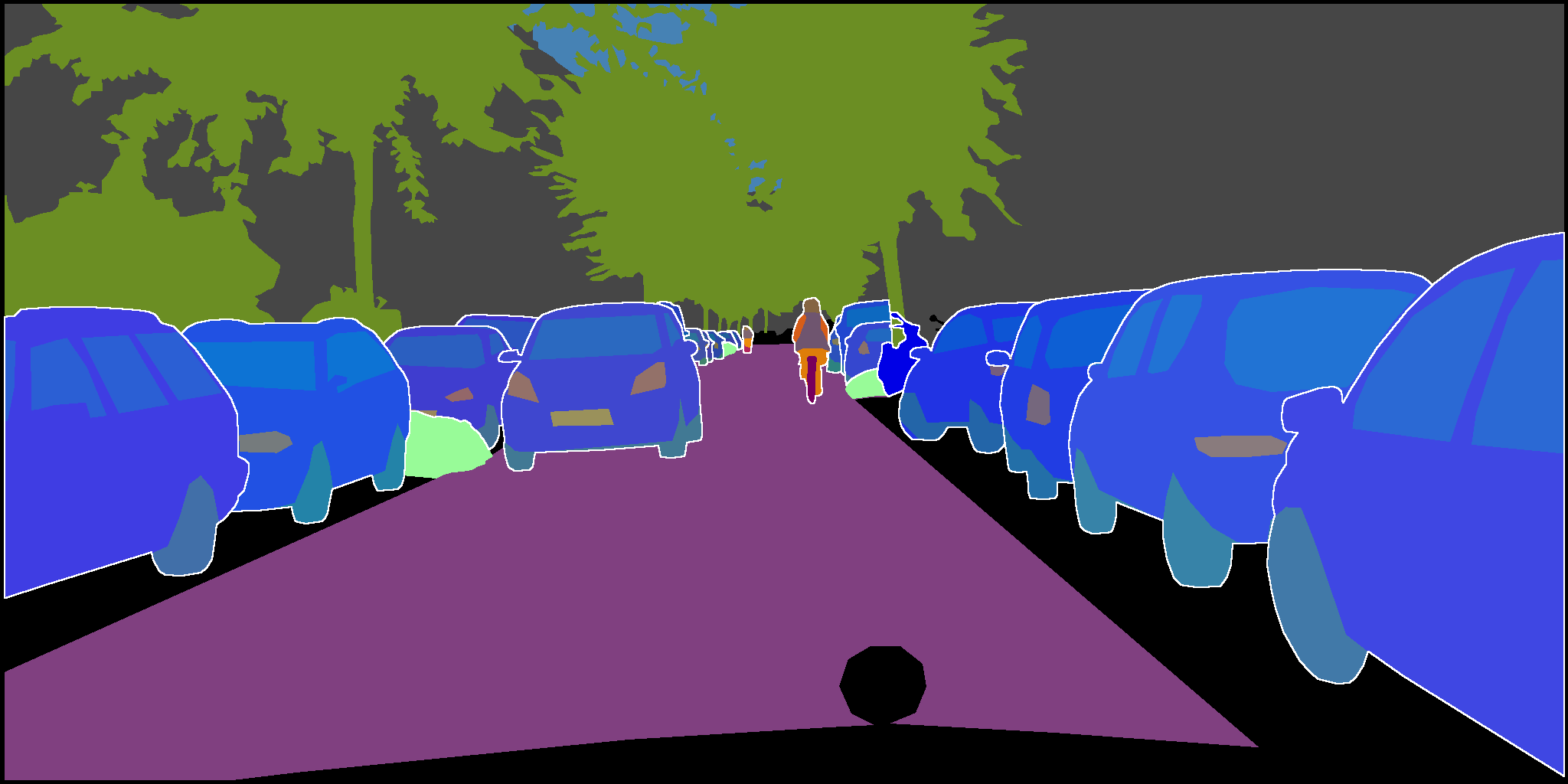}
		\includegraphics[width=0.24\linewidth]{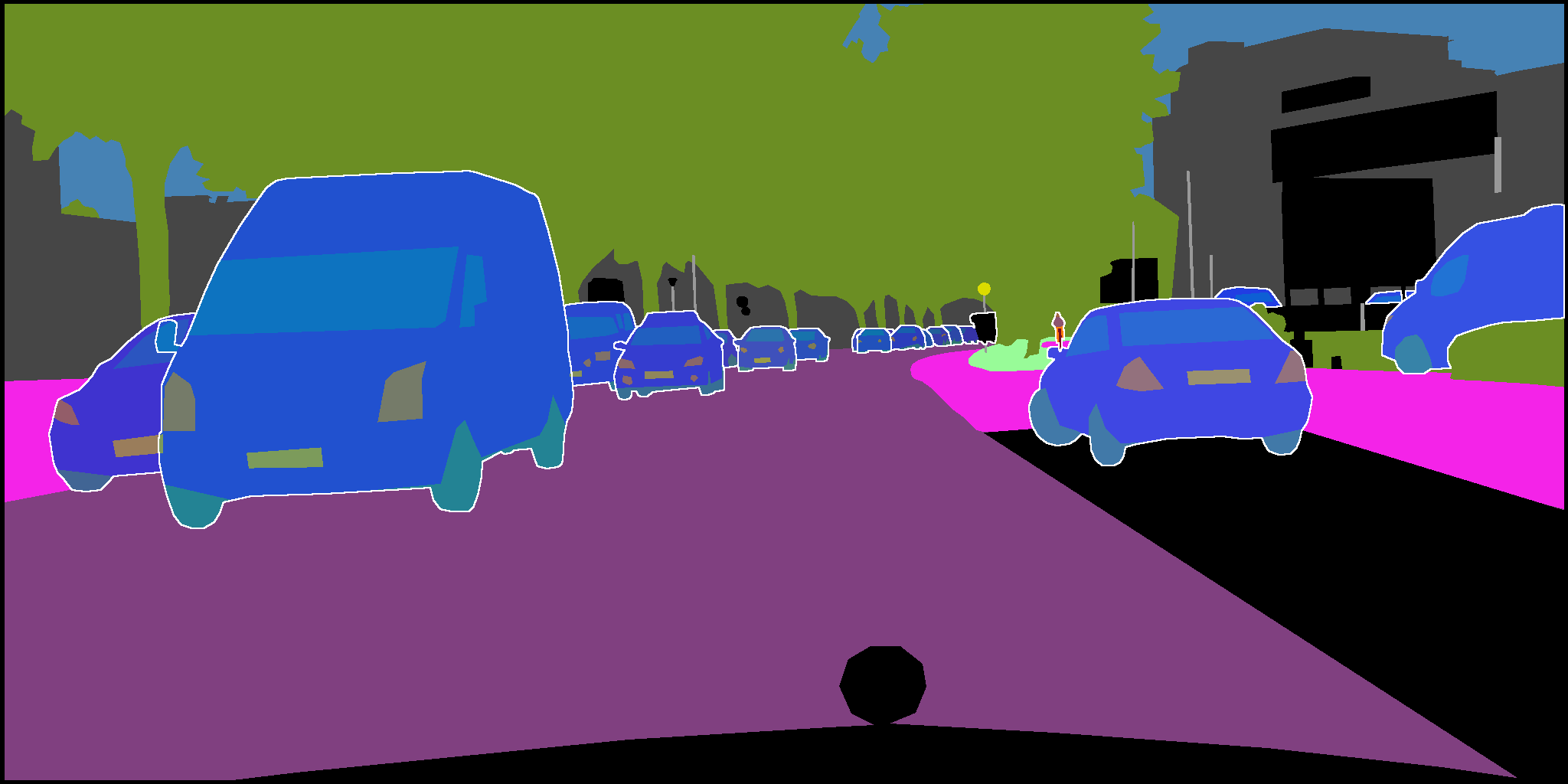}
		\includegraphics[width=0.24\linewidth]{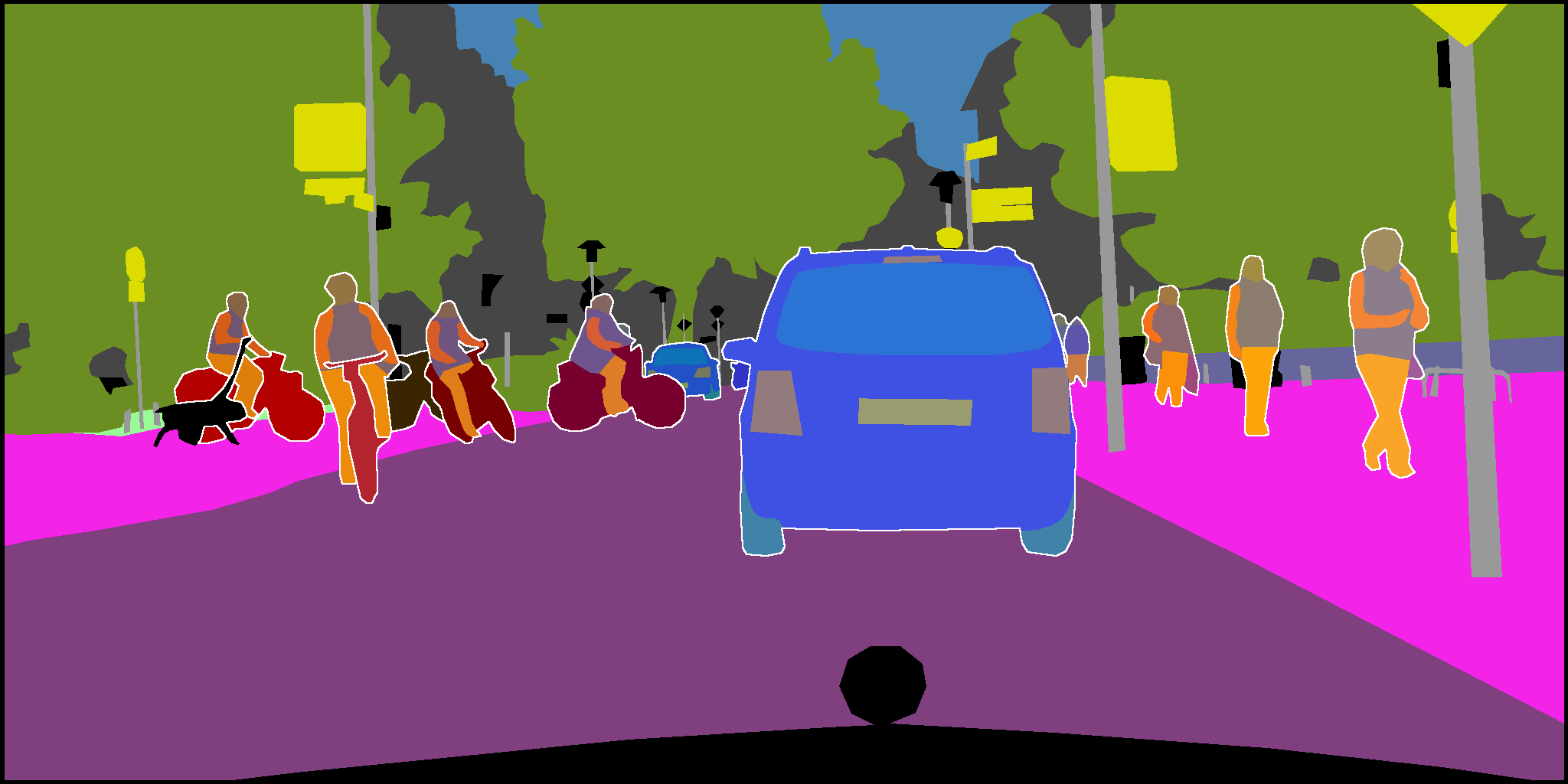} \\
		\includegraphics[width=0.24\linewidth]{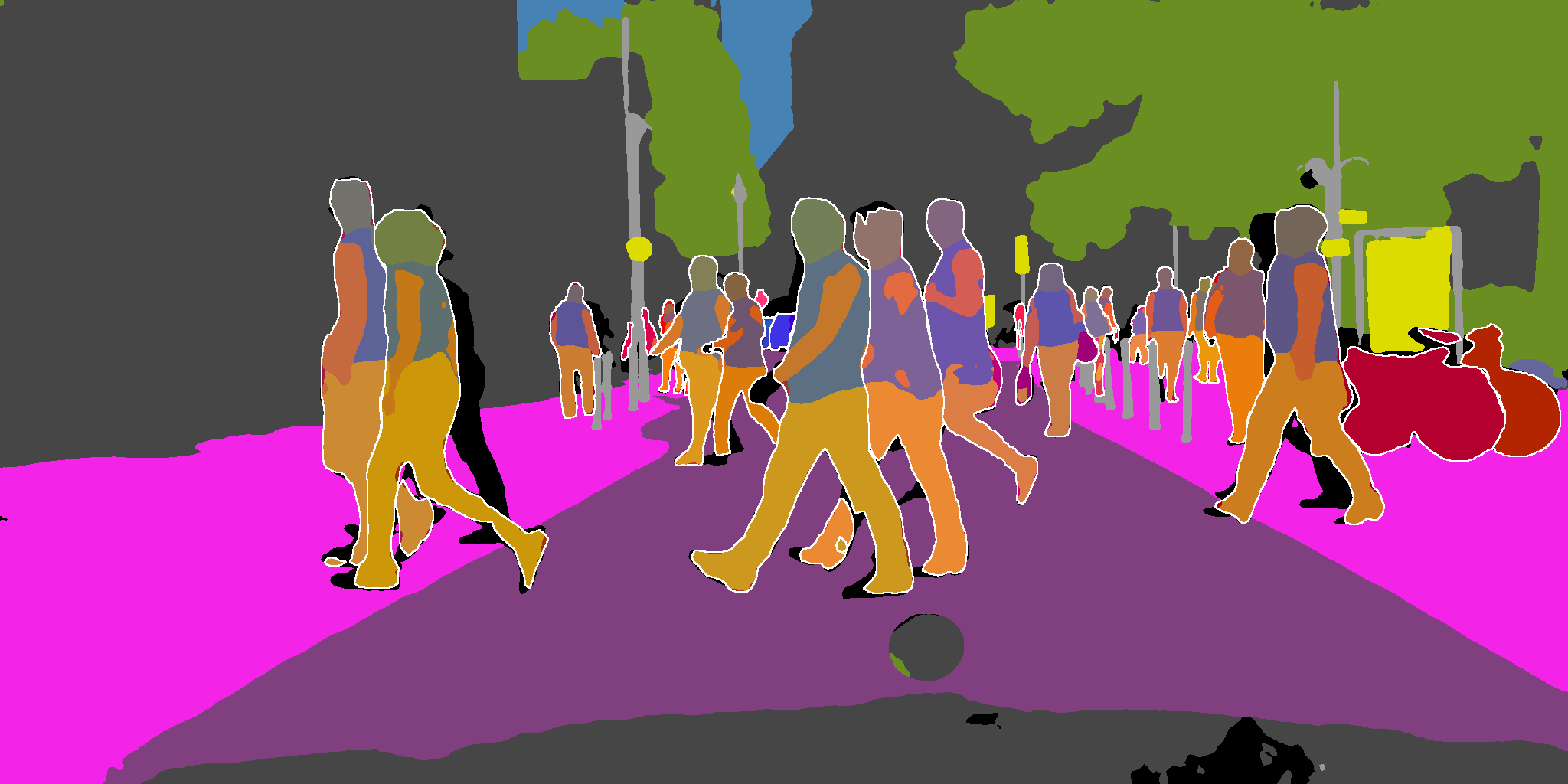}
		\includegraphics[width=0.24\linewidth]{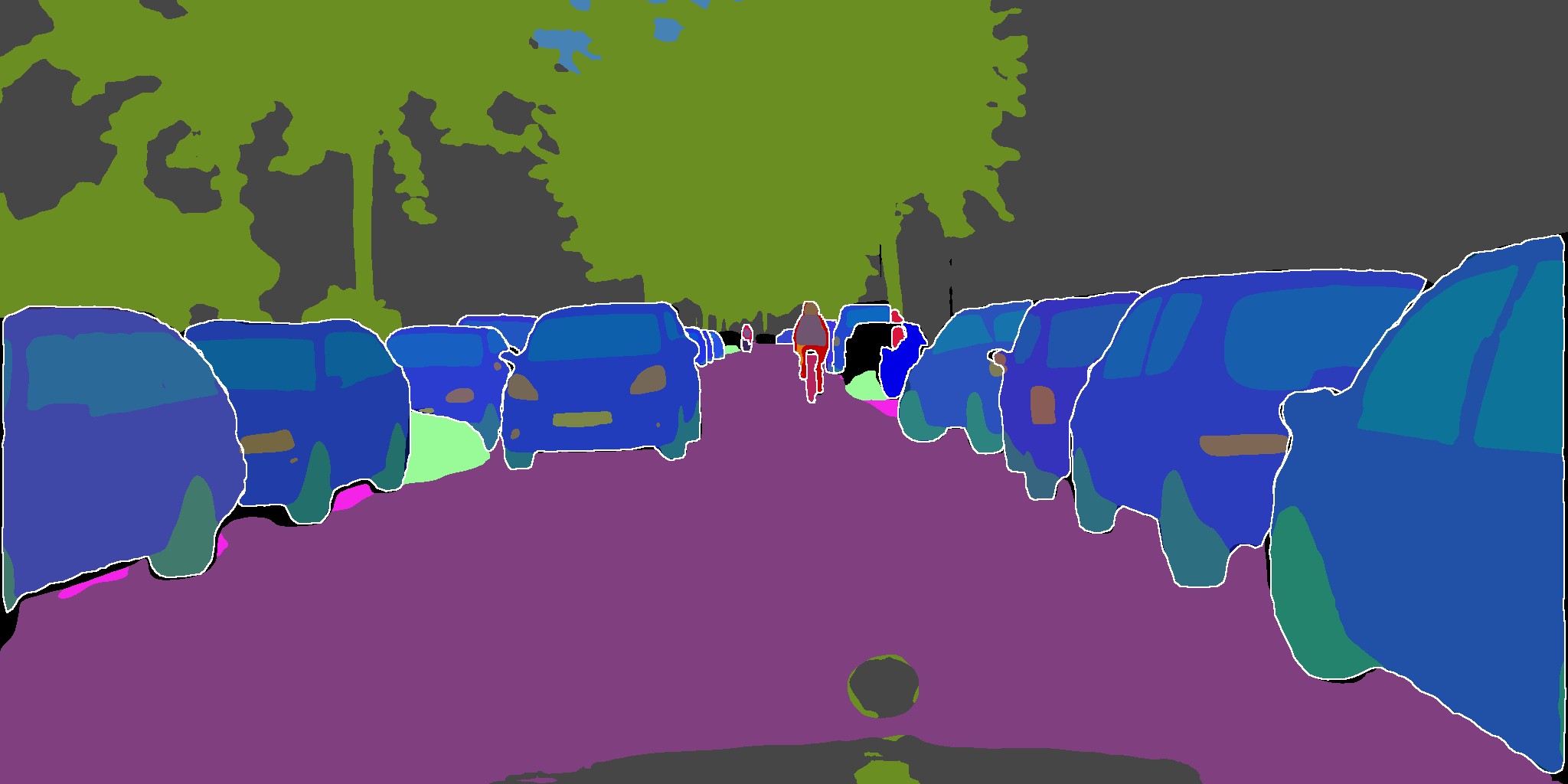}
		\includegraphics[width=0.24\linewidth]{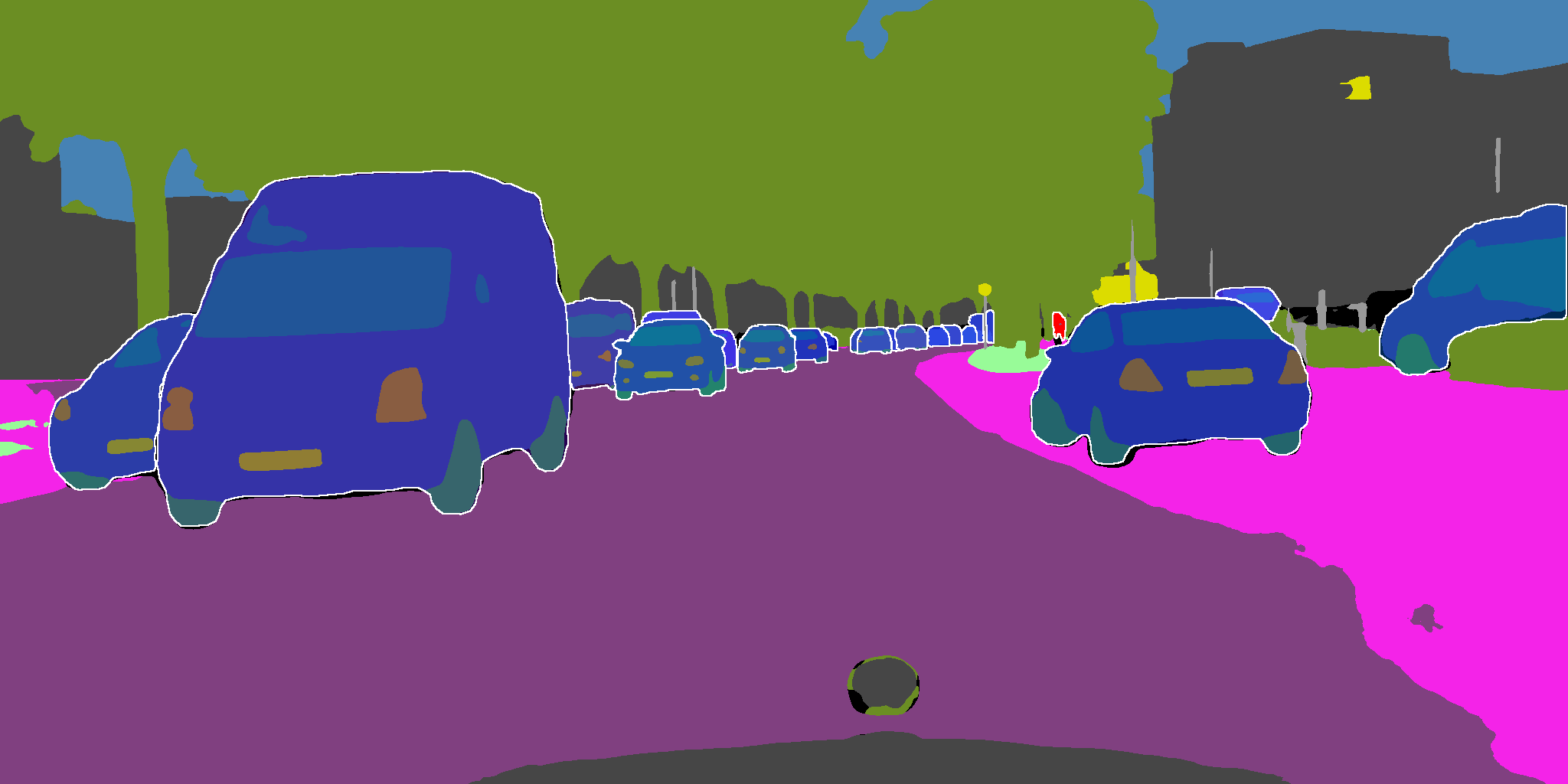}
		\includegraphics[width=0.24\linewidth]{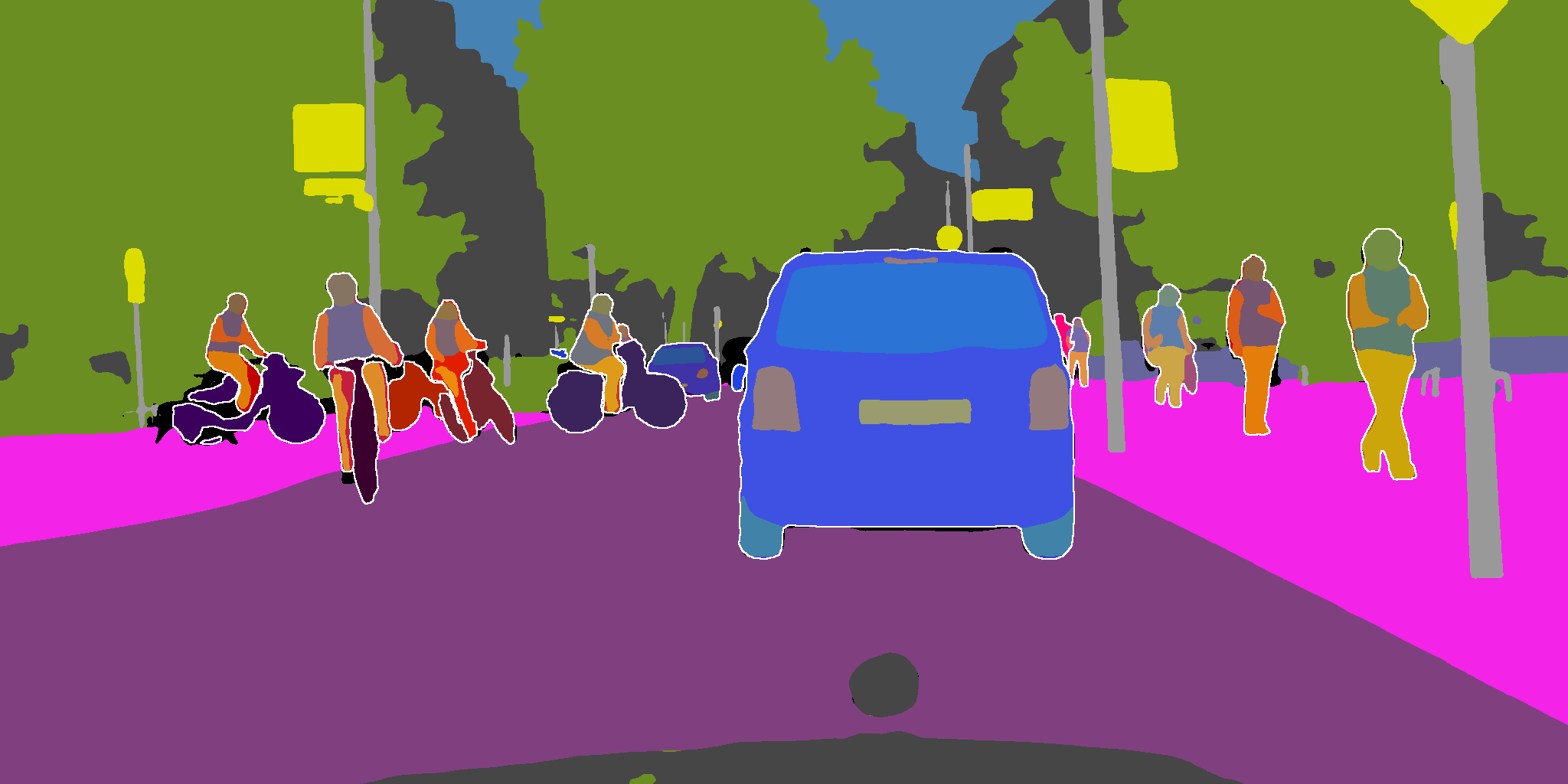} \\
		\vspace{20pt}
		\includegraphics[width=0.24\linewidth]{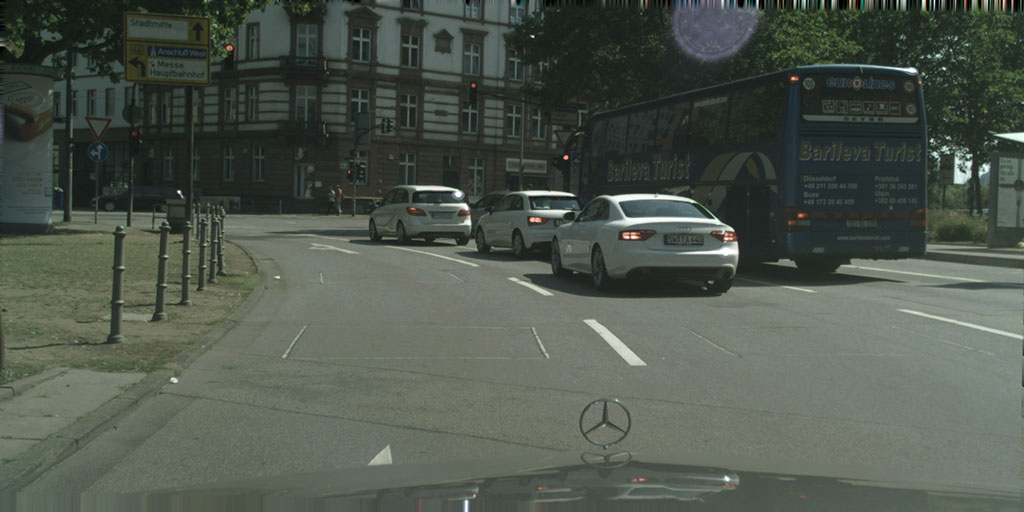}
    	\includegraphics[width=0.24\linewidth]{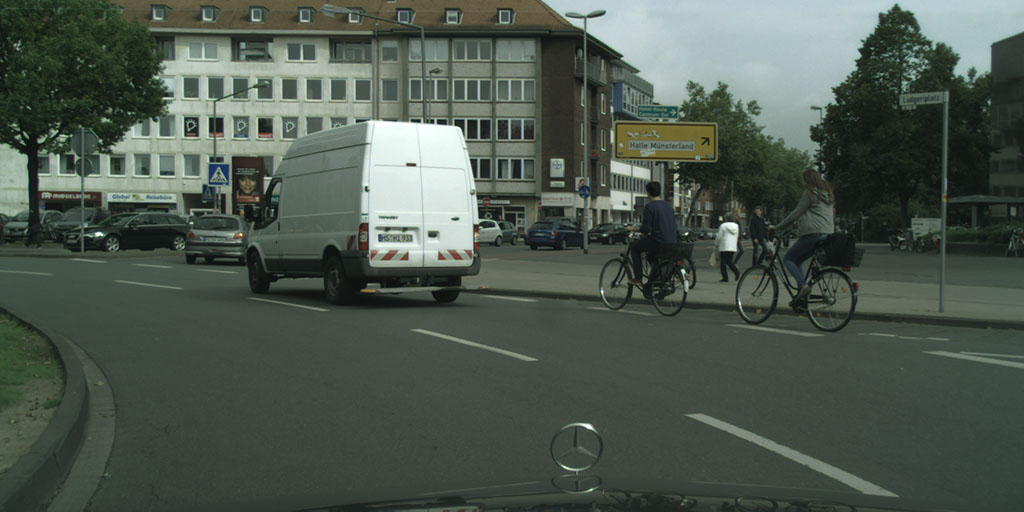}
    	\includegraphics[width=0.24\linewidth]{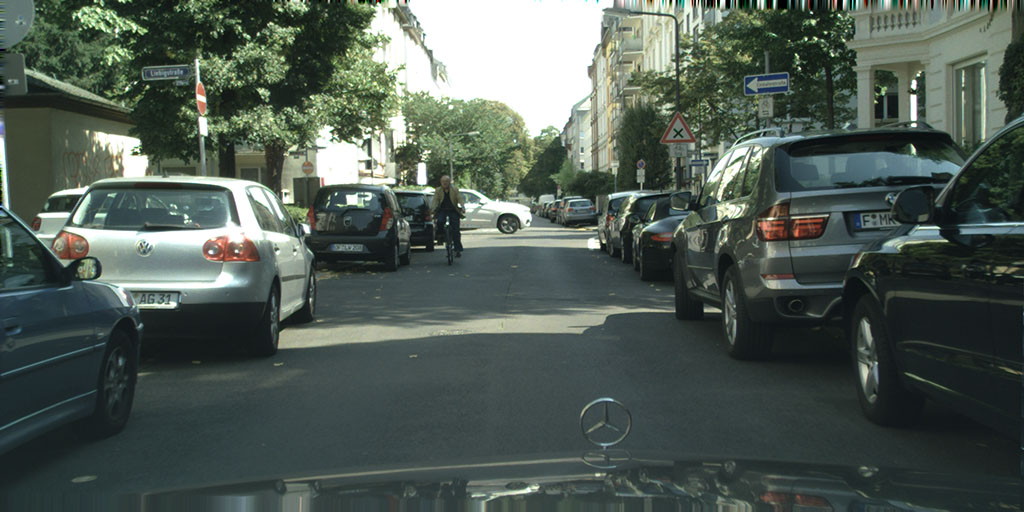}
    	\includegraphics[width=0.24\linewidth]{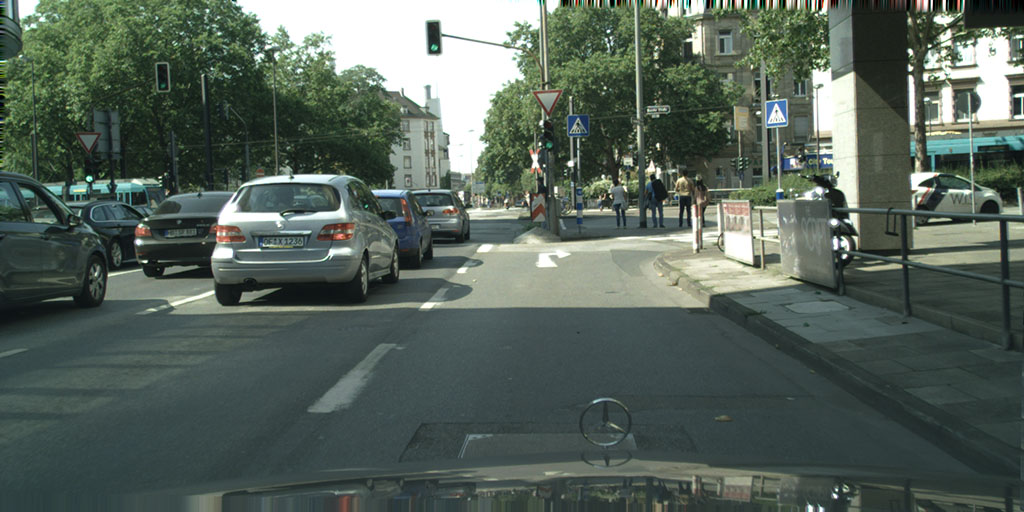}\\
		\includegraphics[width=0.24\linewidth]{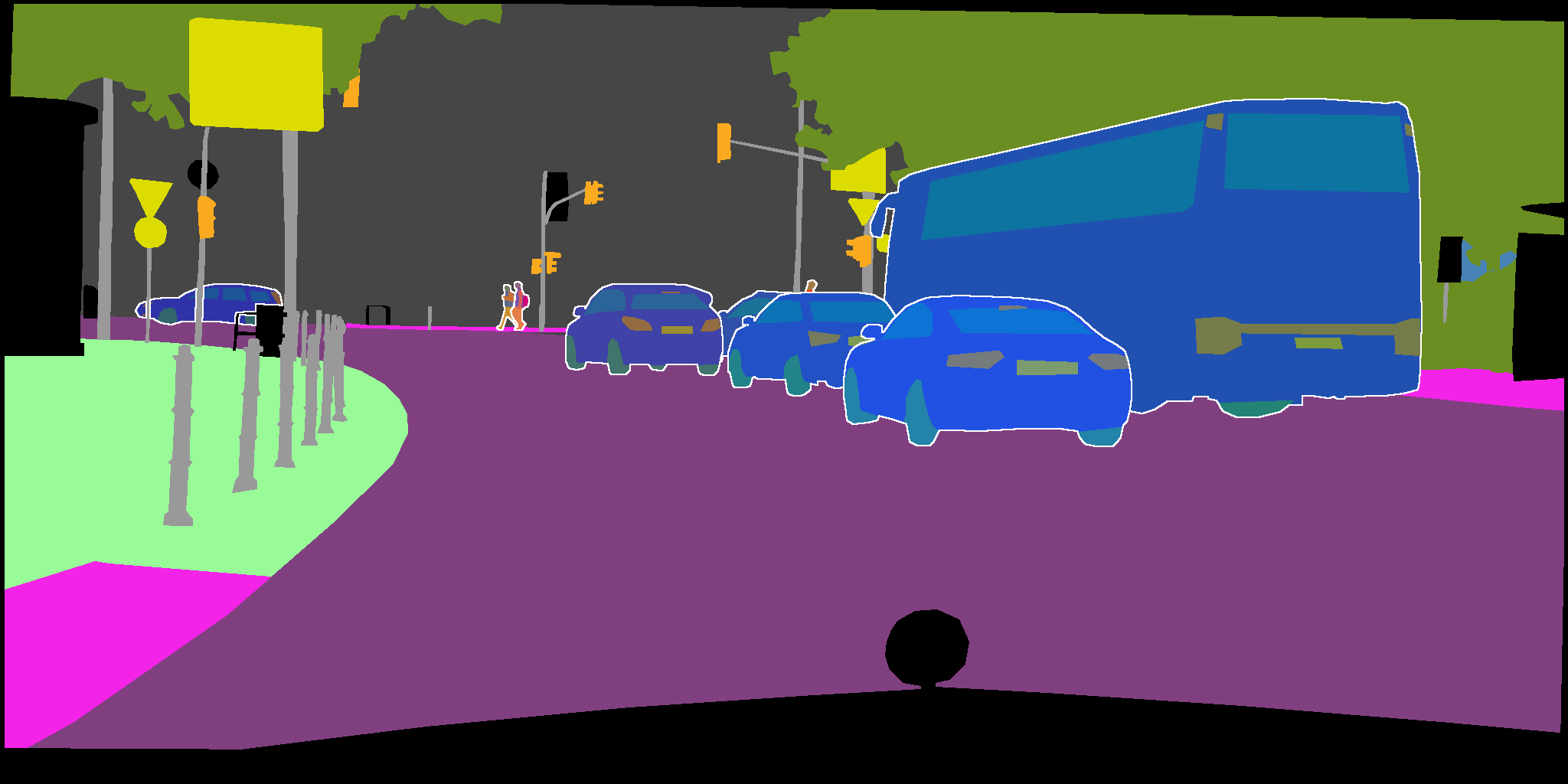}
		\includegraphics[width=0.24\linewidth]{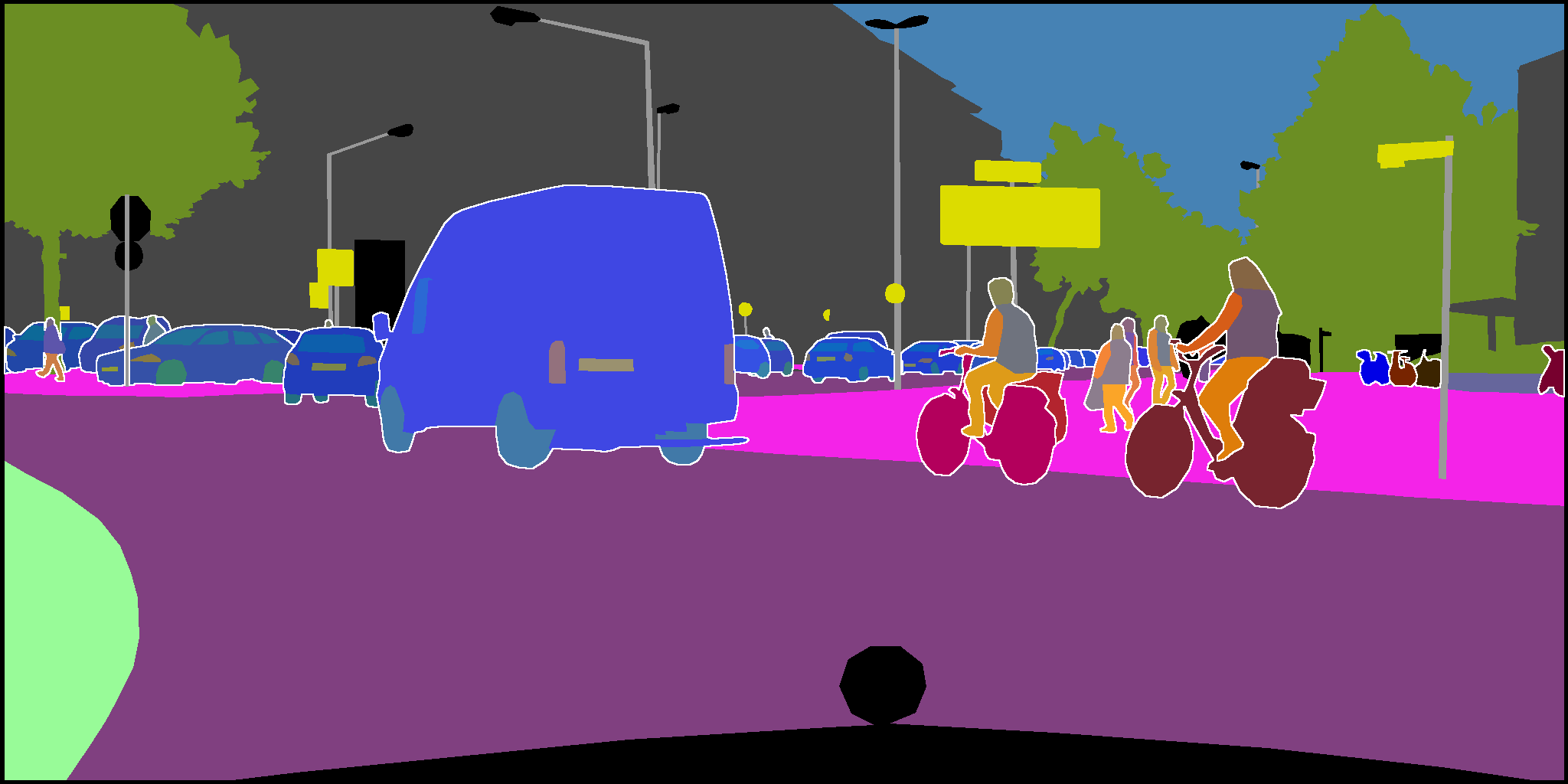}
		\includegraphics[width=0.24\linewidth]{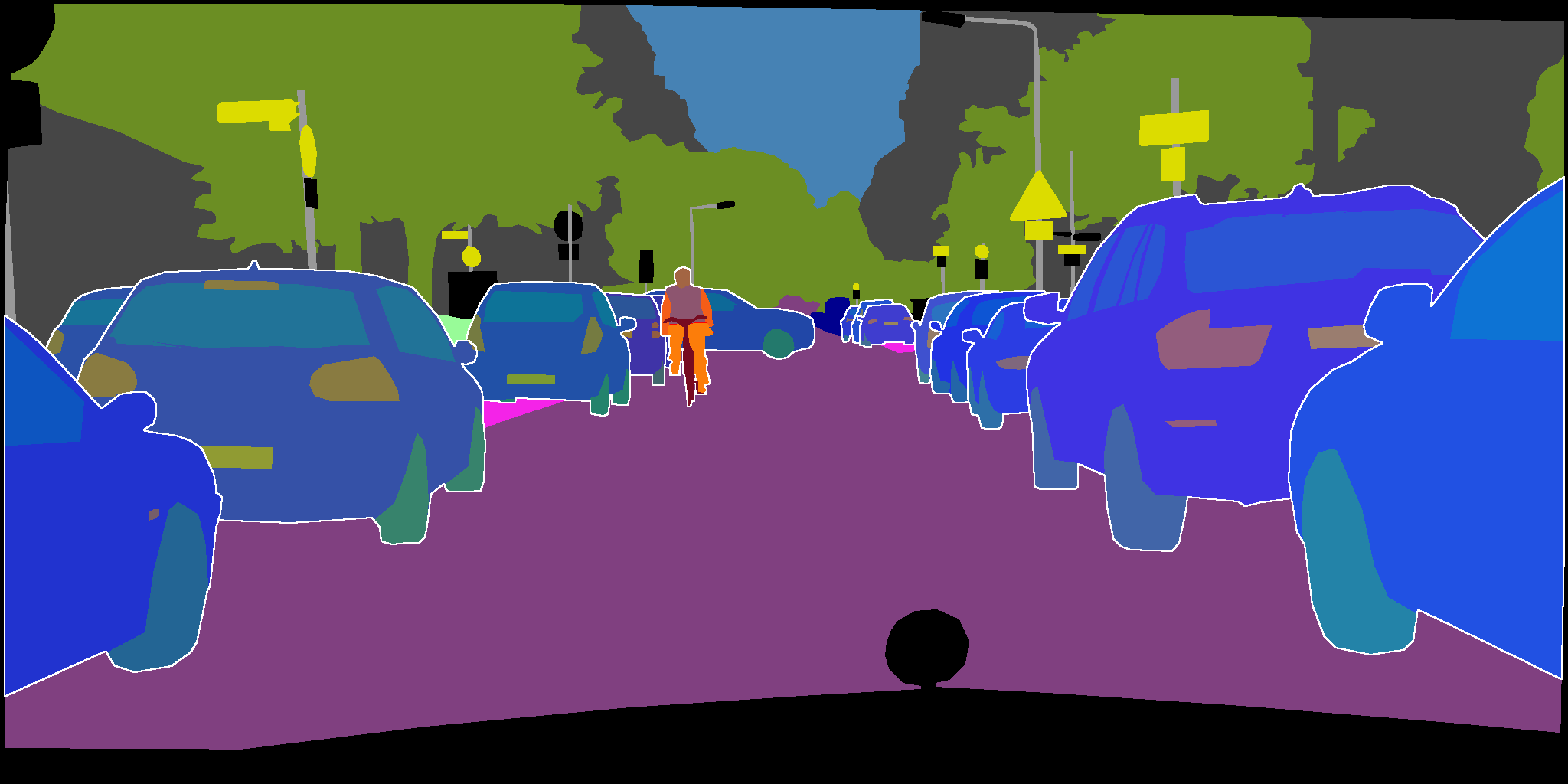}
		\includegraphics[width=0.24\linewidth]{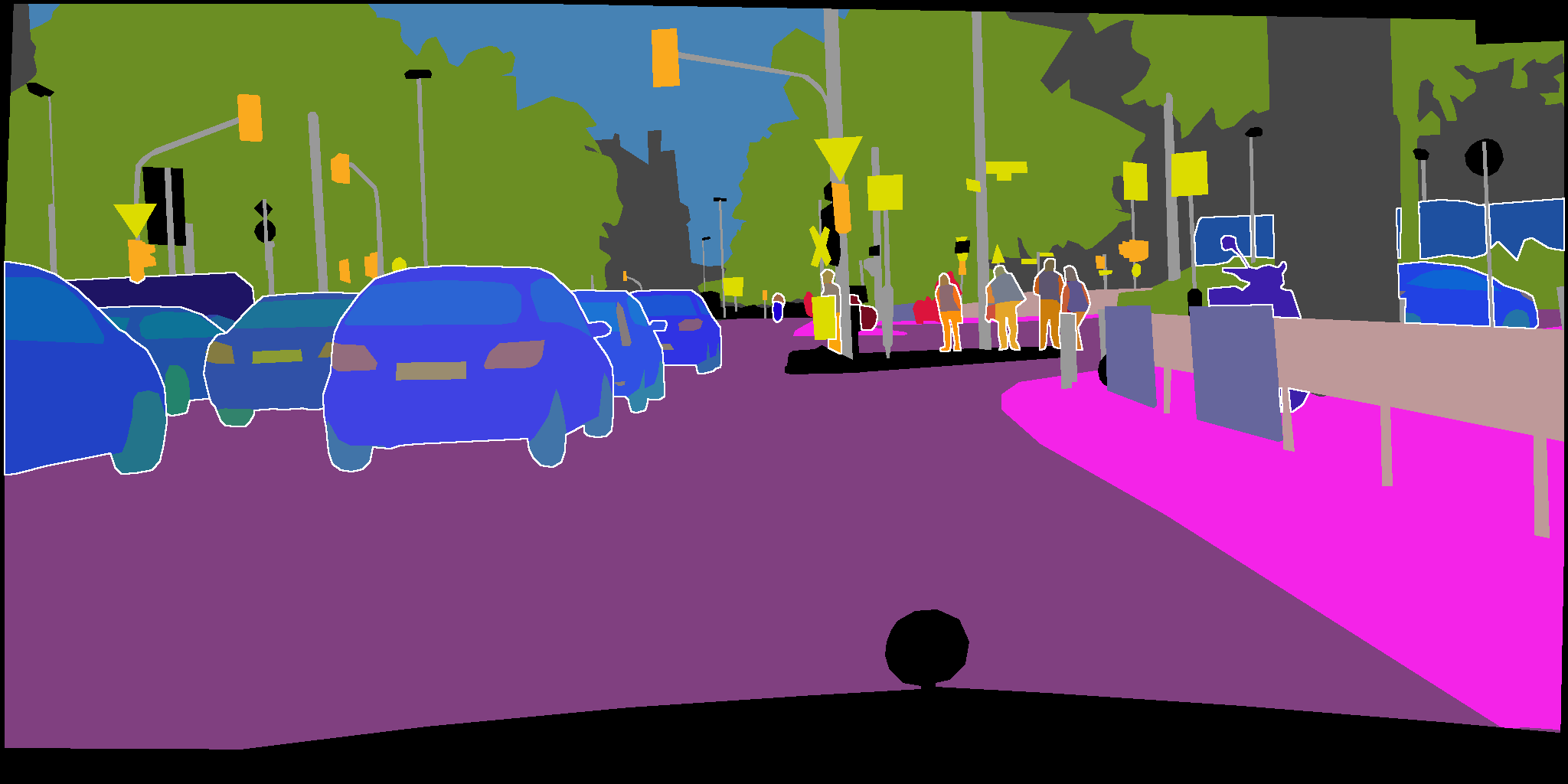} \\
		\includegraphics[width=0.24\linewidth]{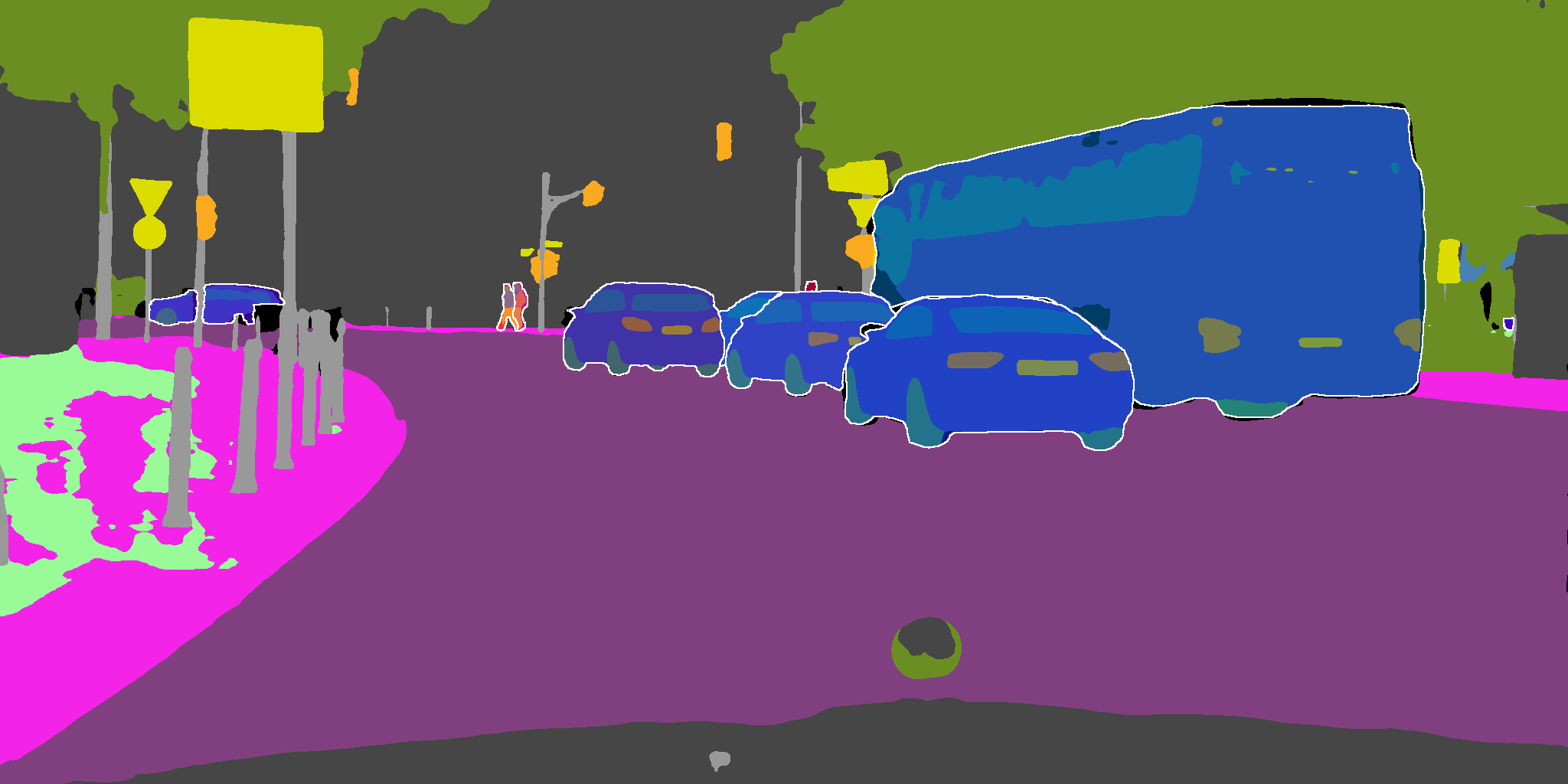}
		\includegraphics[width=0.24\linewidth]{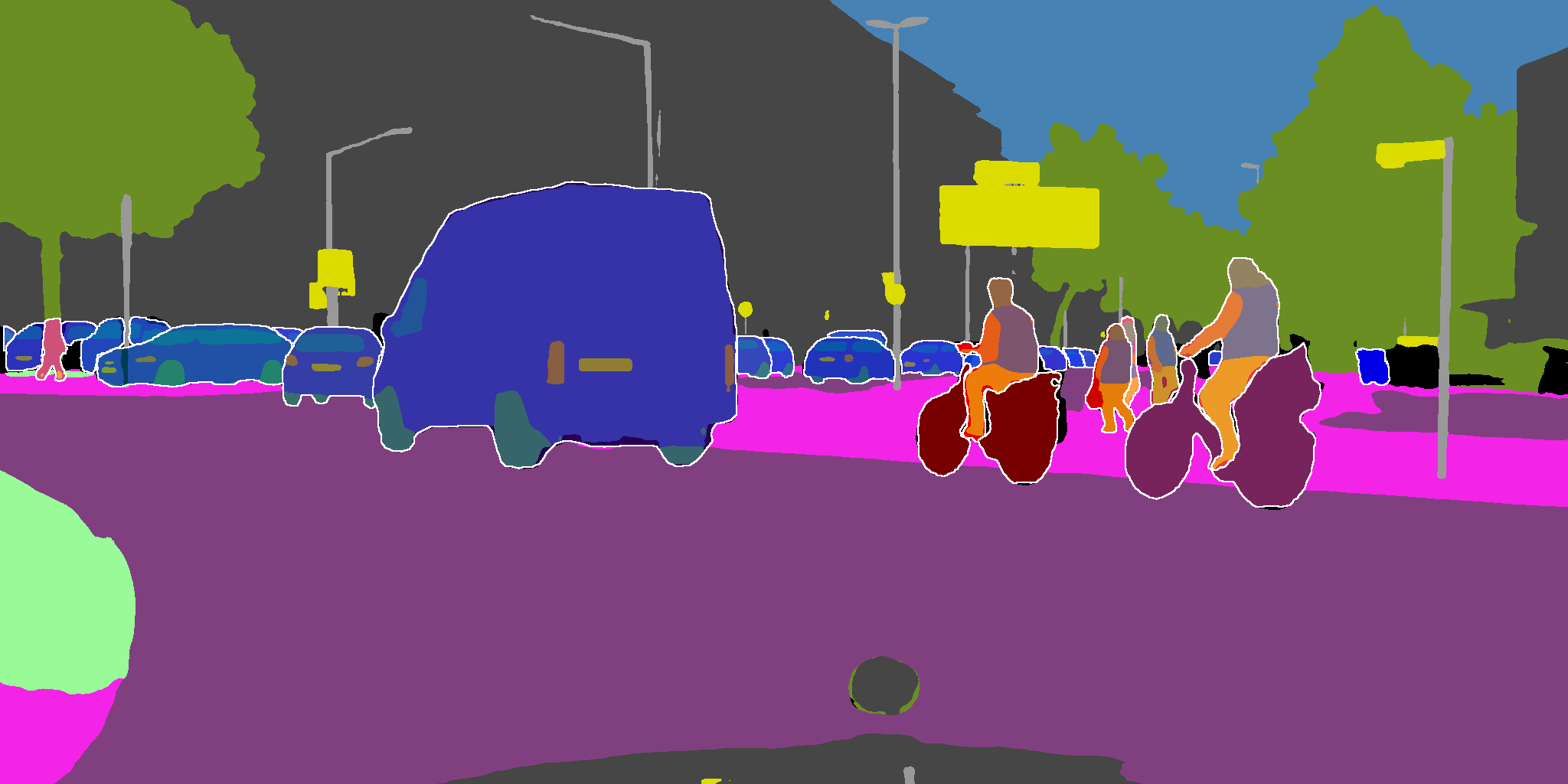}
		\includegraphics[width=0.24\linewidth]{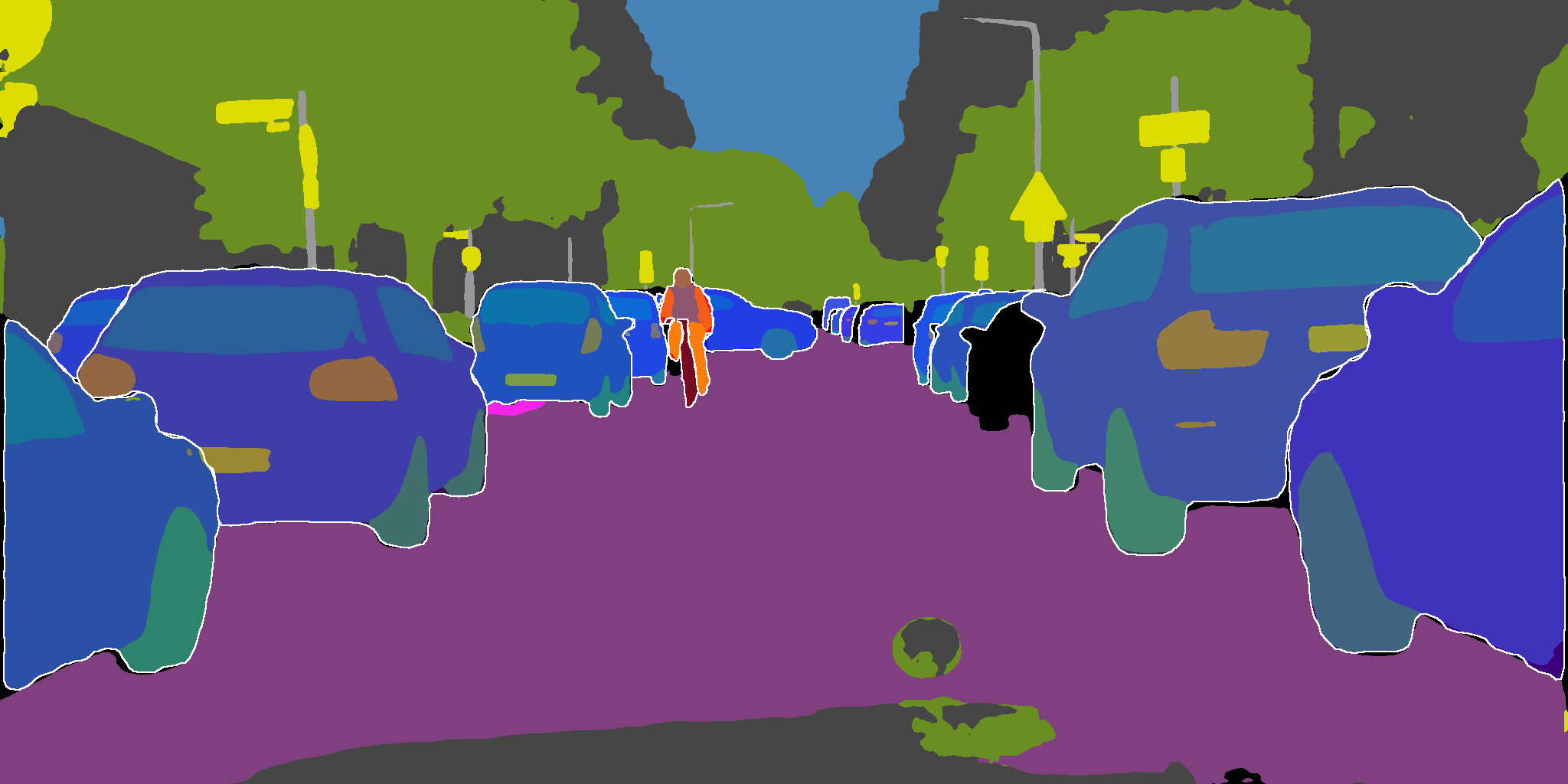}
		\includegraphics[width=0.24\linewidth]{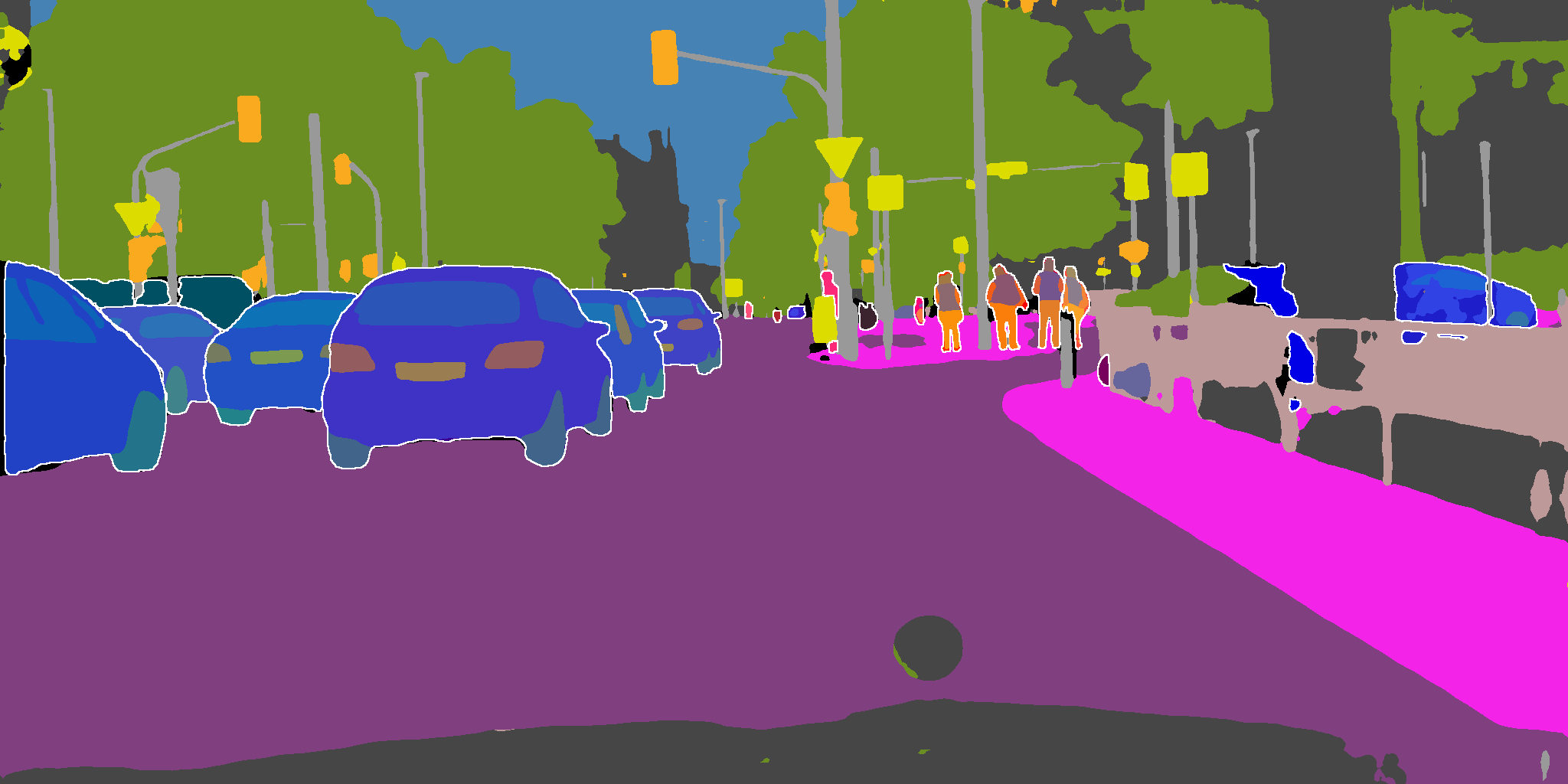} \\
		\vspace{20pt}
		\includegraphics[width=0.24\linewidth]{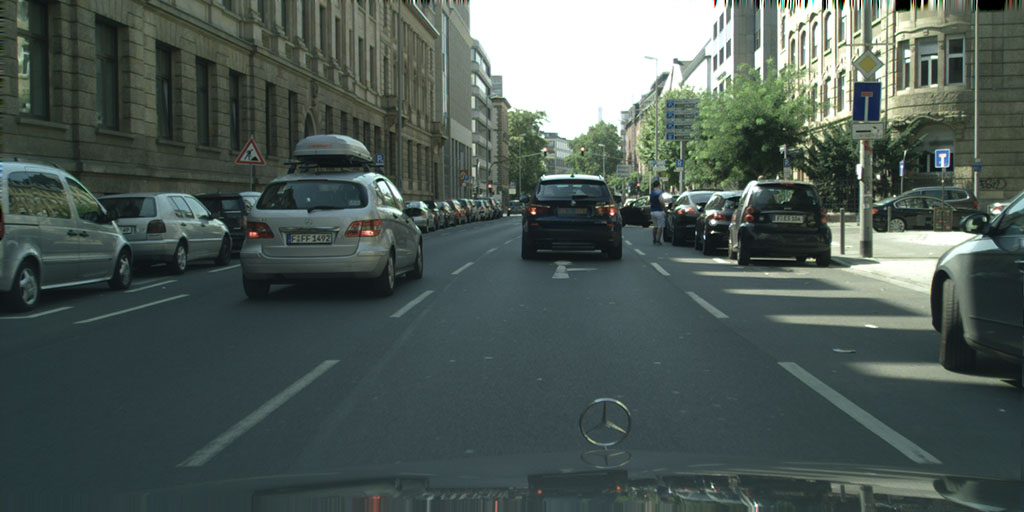}
    	\includegraphics[width=0.24\linewidth]{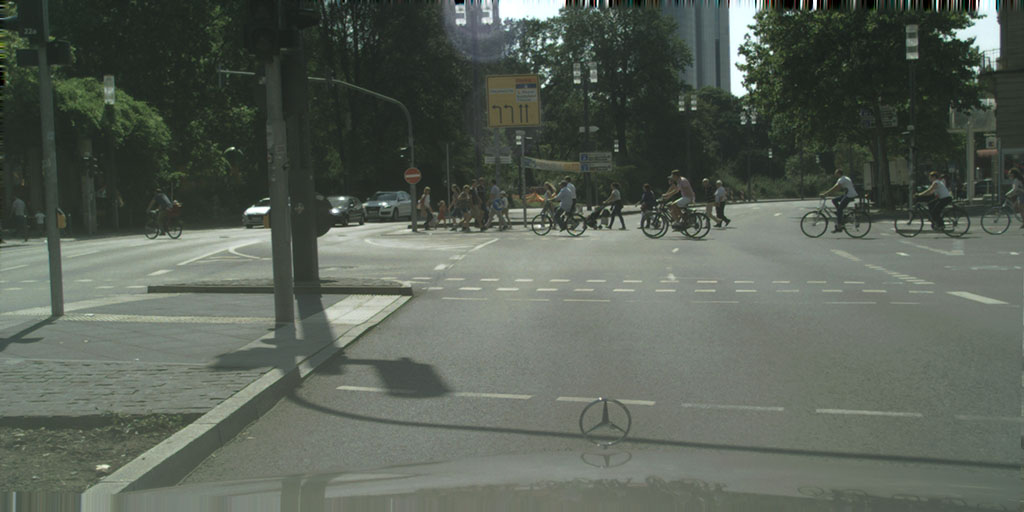}
    	\includegraphics[width=0.24\linewidth]{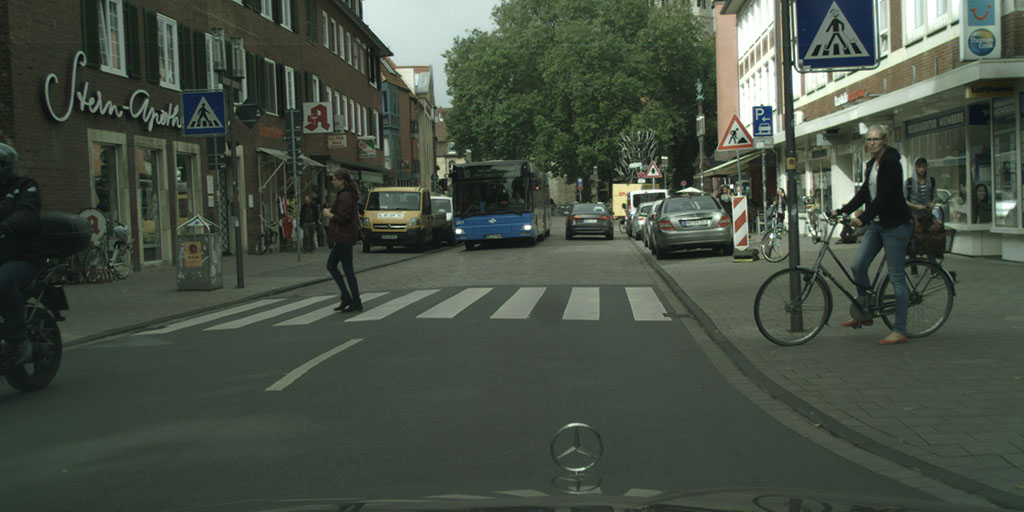}
    	\includegraphics[width=0.24\linewidth]{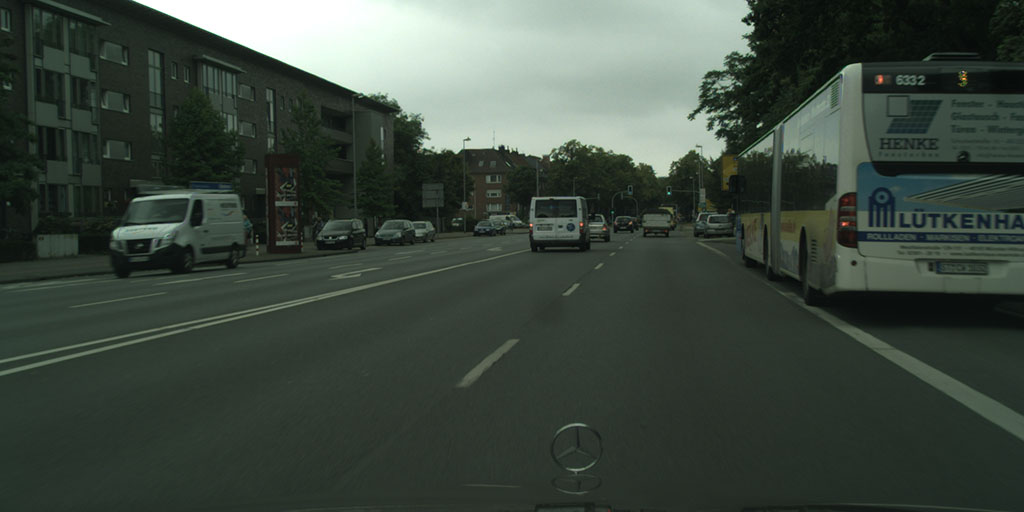}\\
		\includegraphics[width=0.24\linewidth]{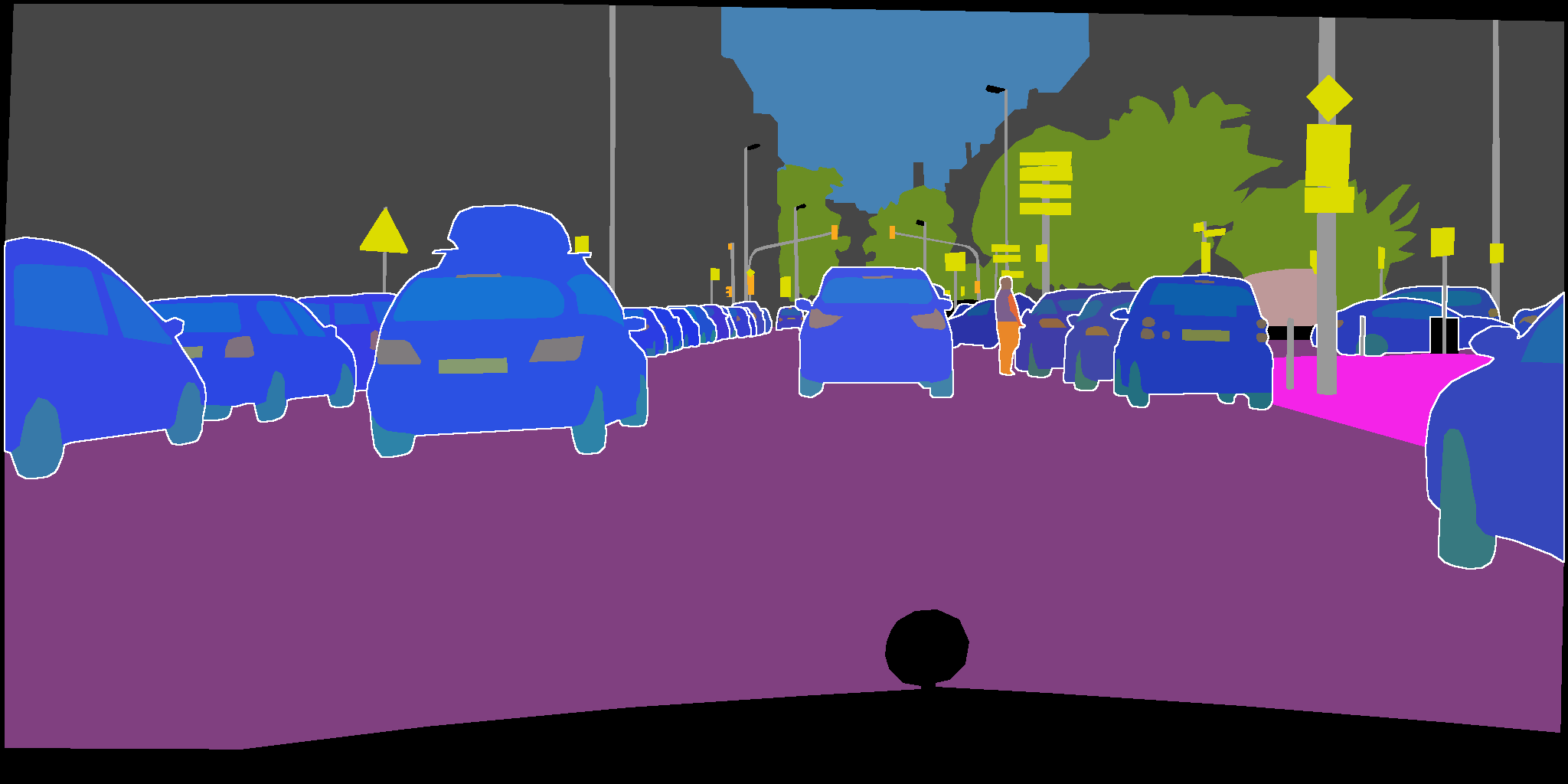}
		\includegraphics[width=0.24\linewidth]{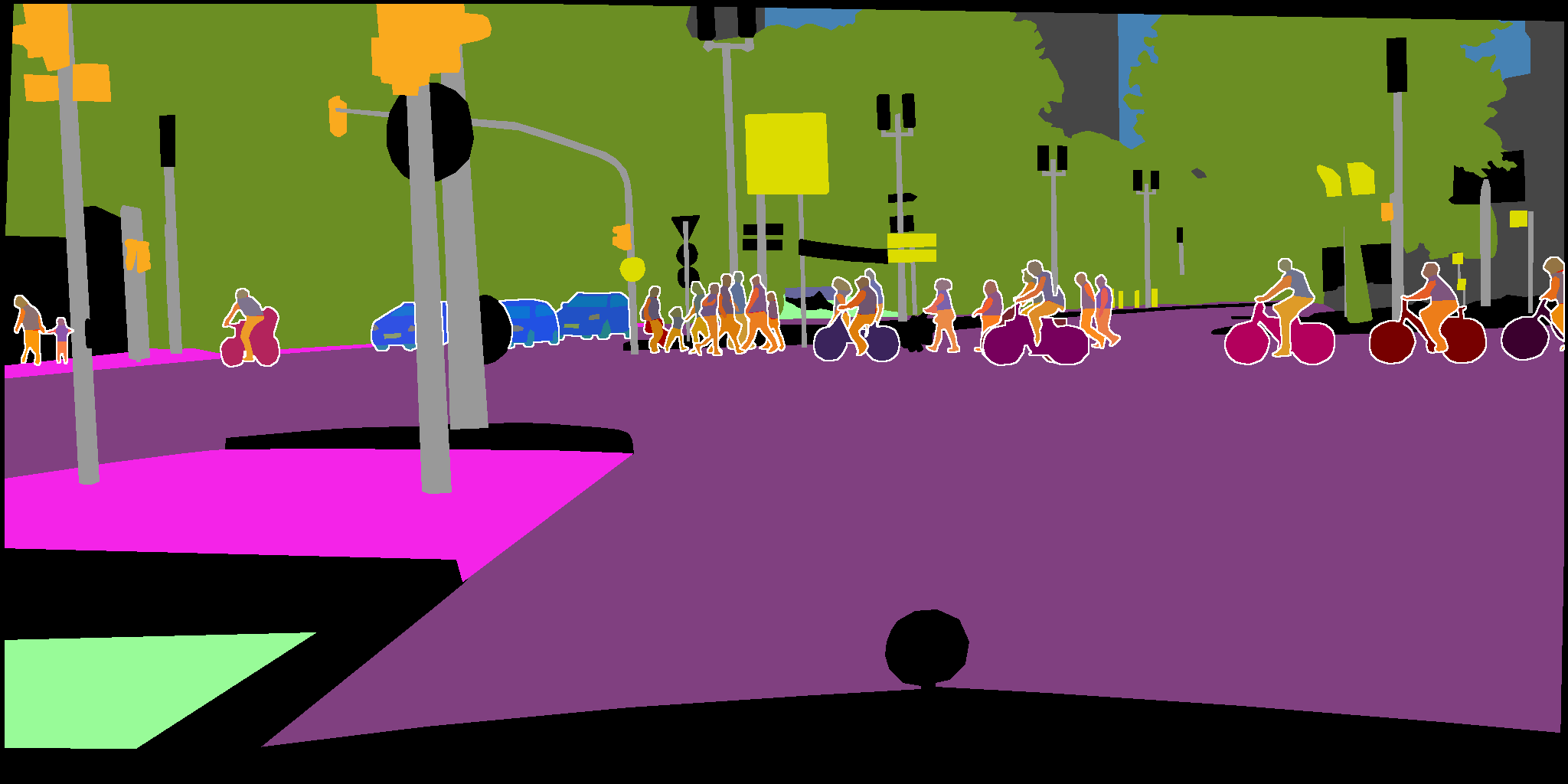}
		\includegraphics[width=0.24\linewidth]{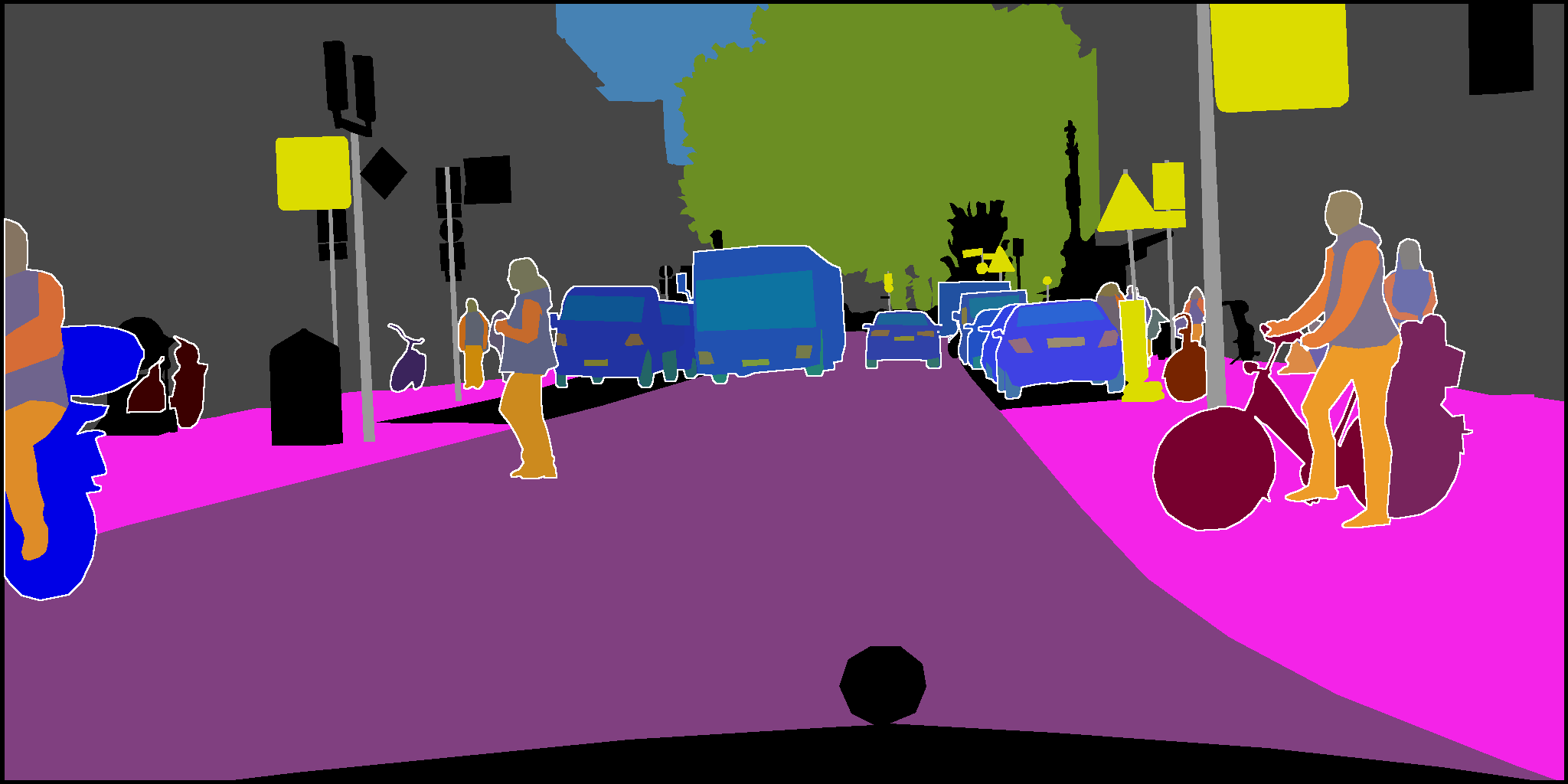}
		\includegraphics[width=0.24\linewidth]{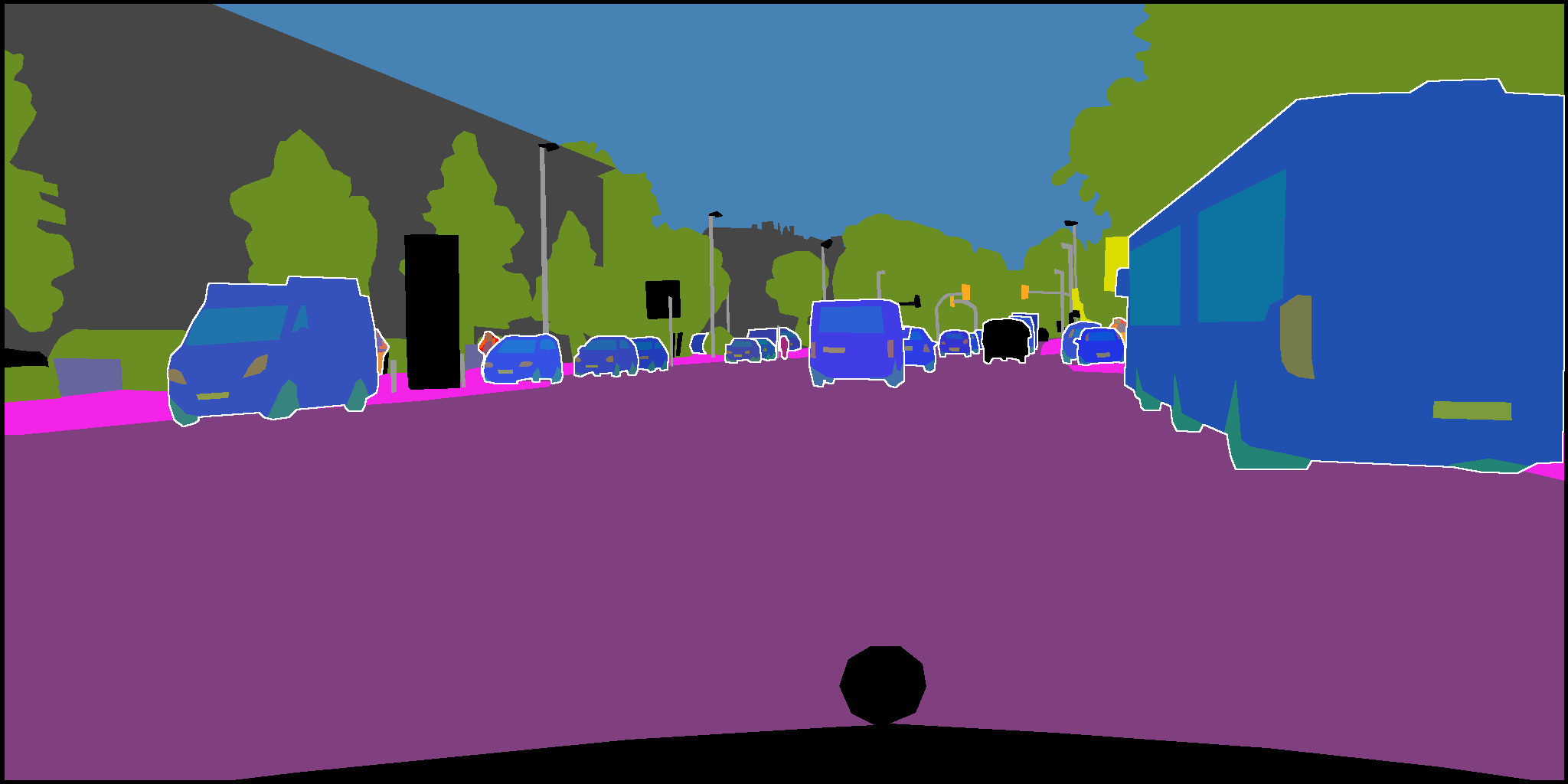} \\
		\includegraphics[width=0.24\linewidth]{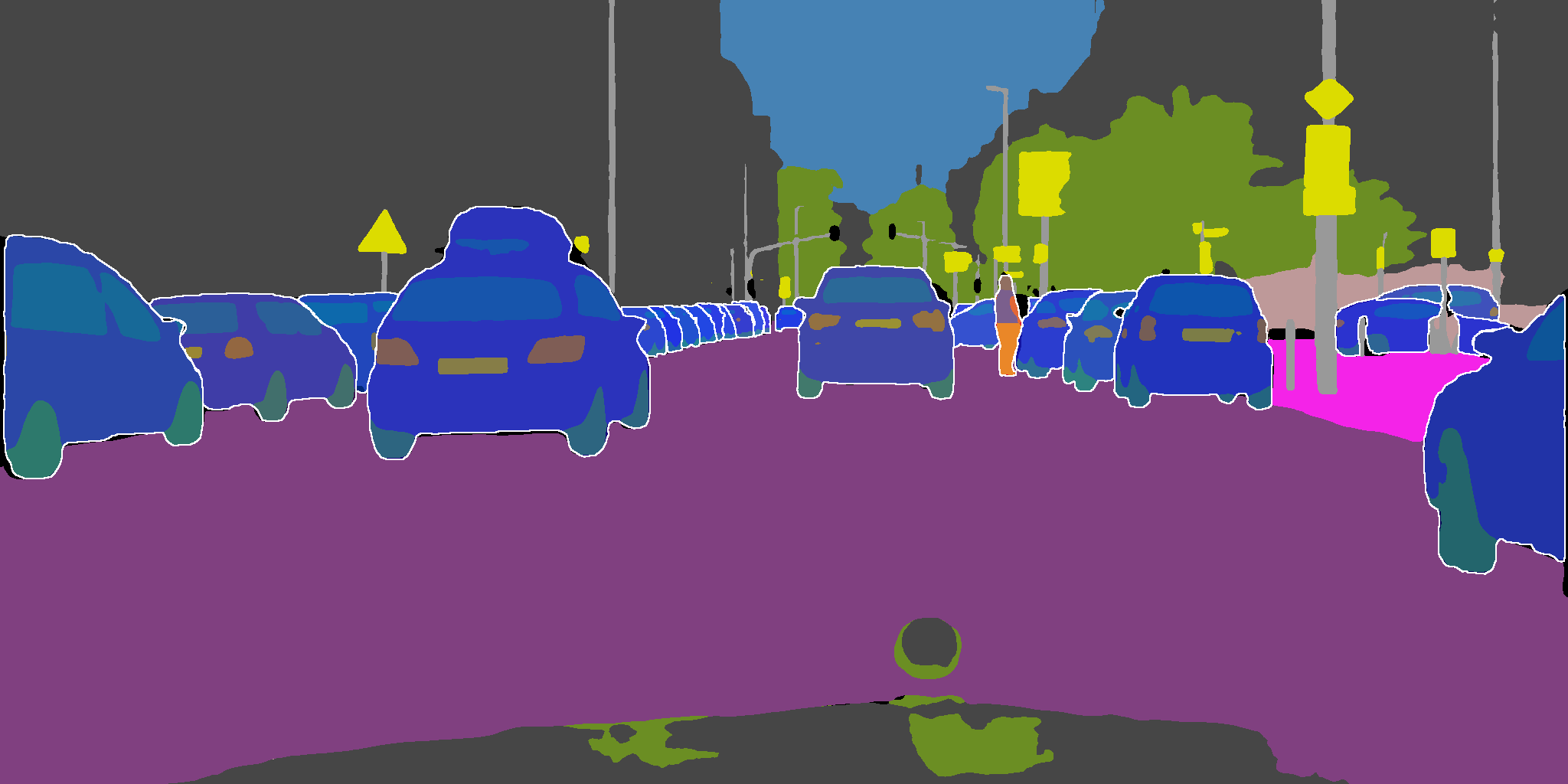}
		\includegraphics[width=0.24\linewidth]{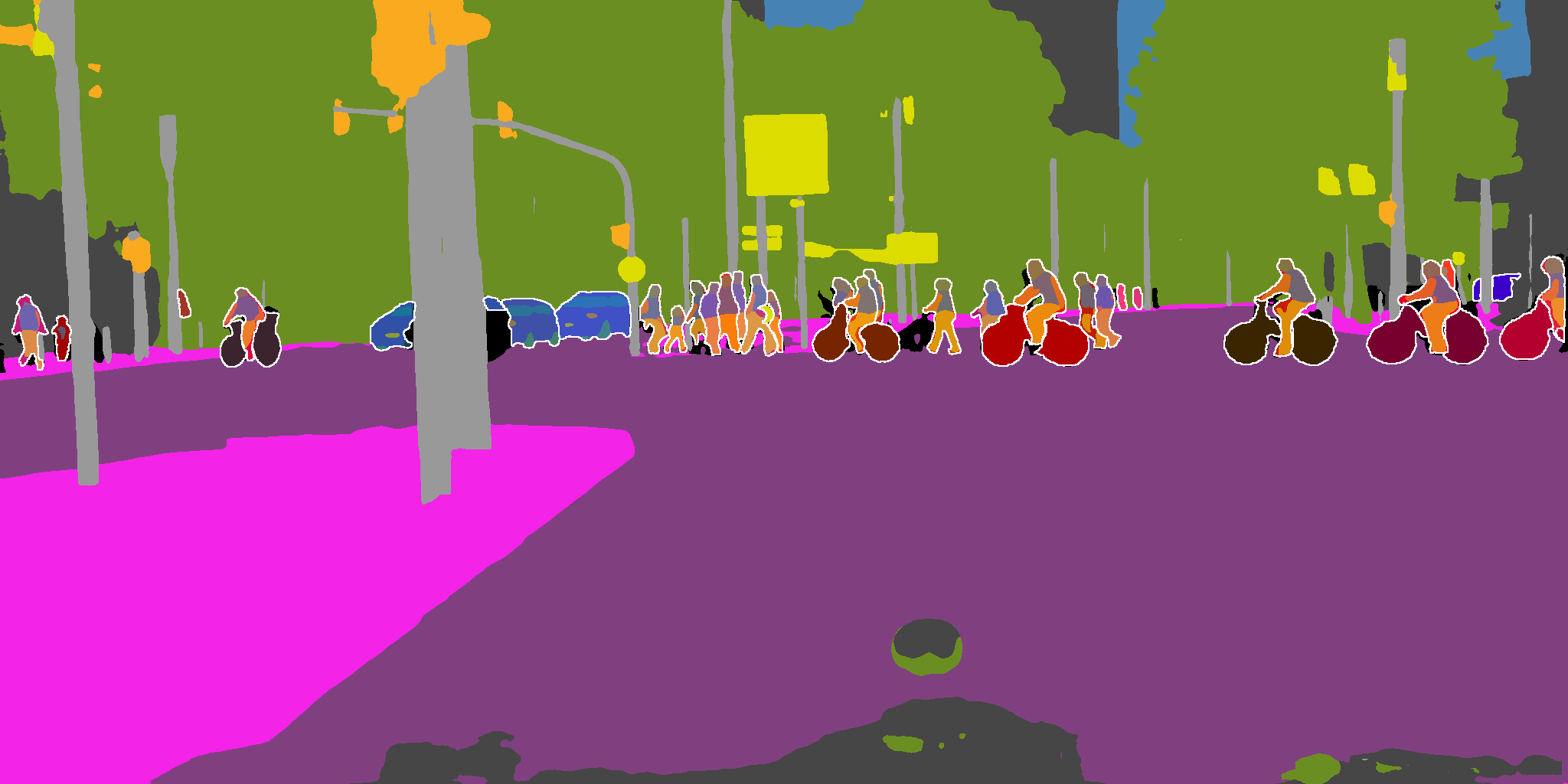}
		\includegraphics[width=0.24\linewidth]{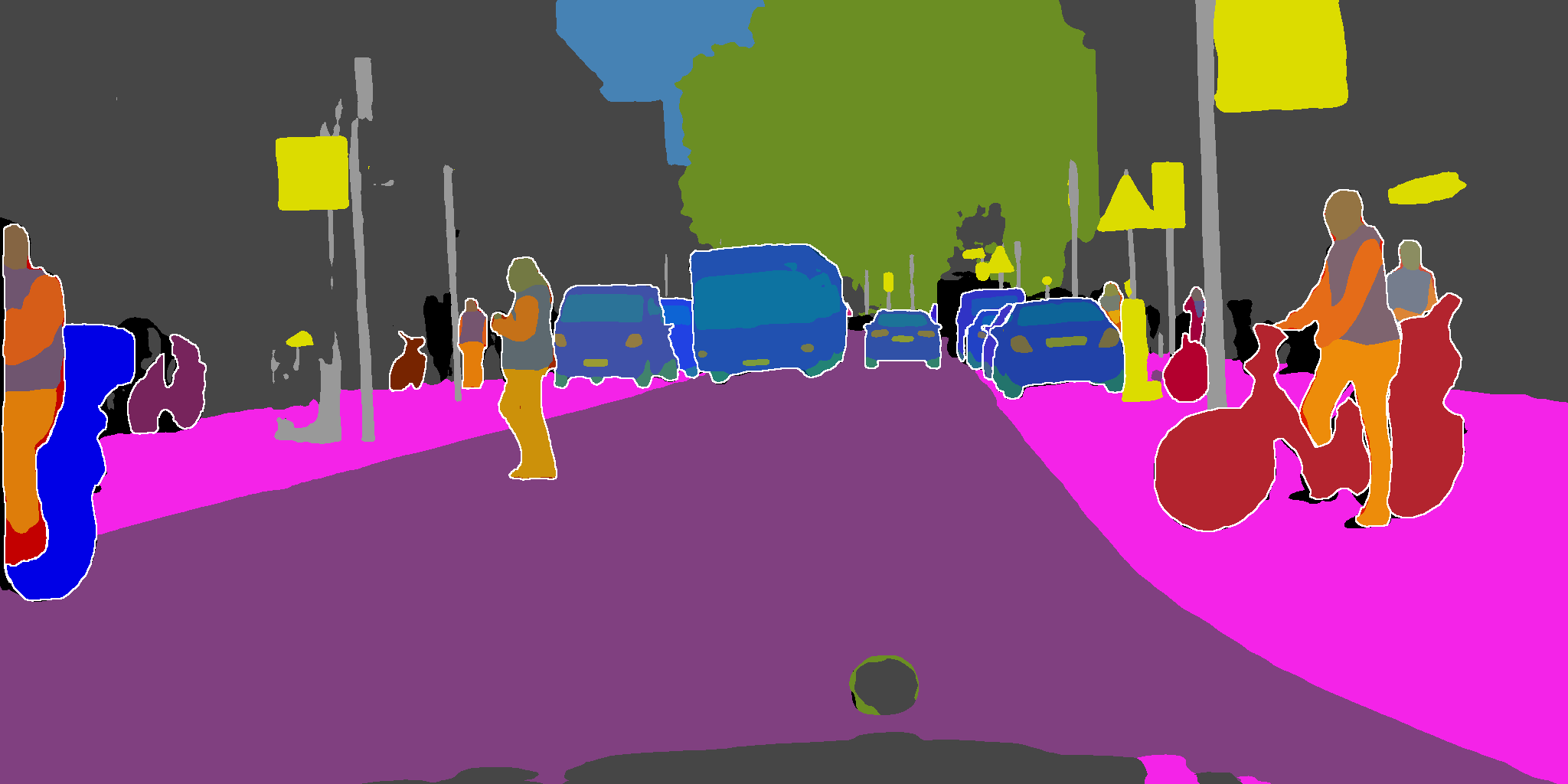}
		\includegraphics[width=0.24\linewidth]{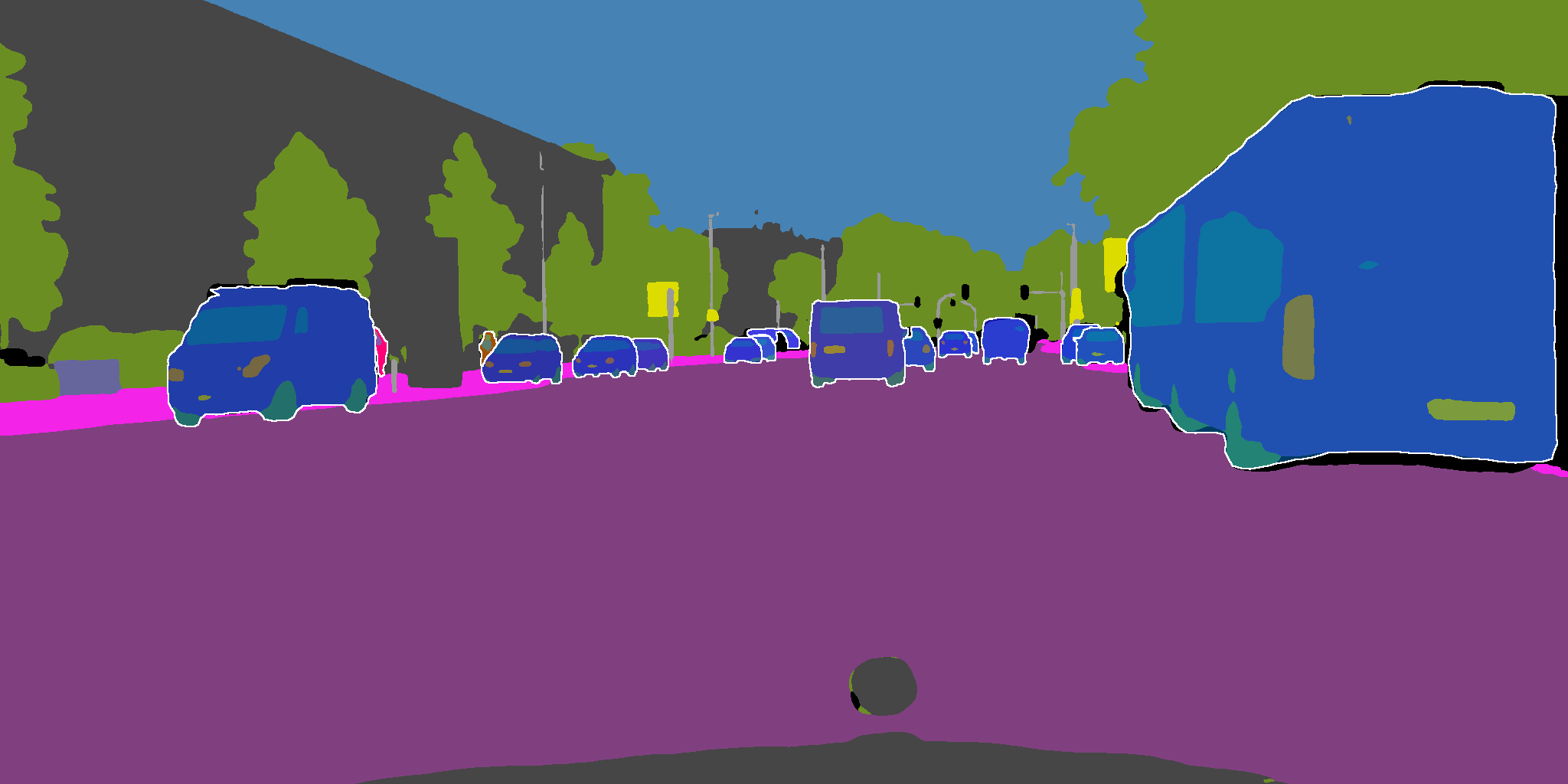} \\
	\caption{Qualitative examples for the highest-scoring part-aware panoptic segmentation baseline on Cityscapes Panoptic Parts (HRNet-OCR \& PolyTransform \& BSANet~\cite{Yuan2020ocr,Liang2020polytransform,Zhao2019BSANet}). On each of the three rows, we show the input images (top), ground truth (middle) and predictions (bottom).}
\label{fig:cpp_examples_supp}
\end{figure*}

\begin{figure*}[t]
	\centering
    	\includegraphics[height=0.079\textheight]{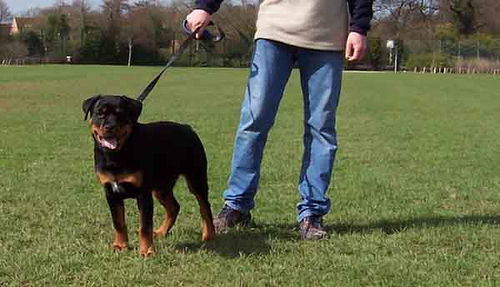}
    	\includegraphics[height=0.079\textheight]{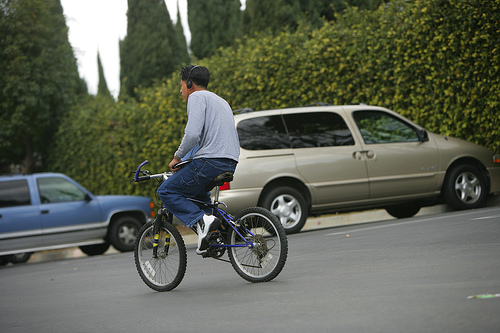}
    	\includegraphics[height=0.079\textheight]{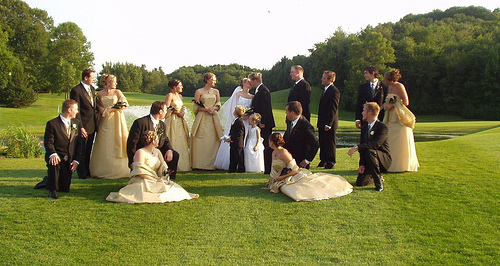}
    	\includegraphics[height=0.079\textheight]{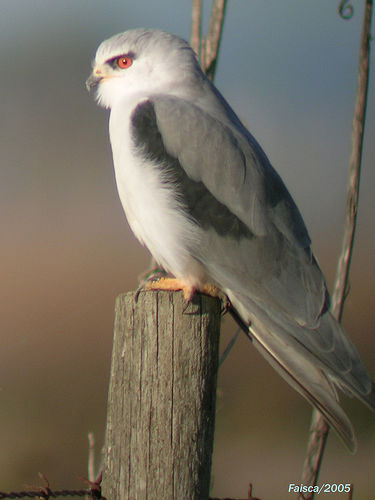}
    	\includegraphics[height=0.079\textheight]{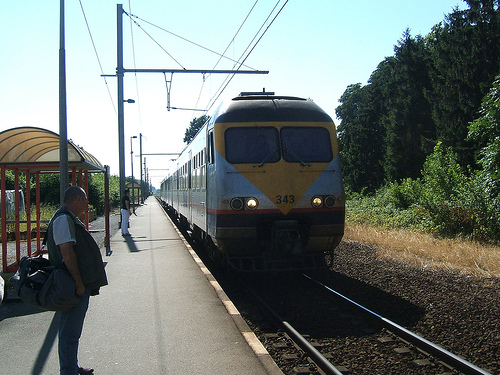}
    	\includegraphics[height=0.079\textheight]{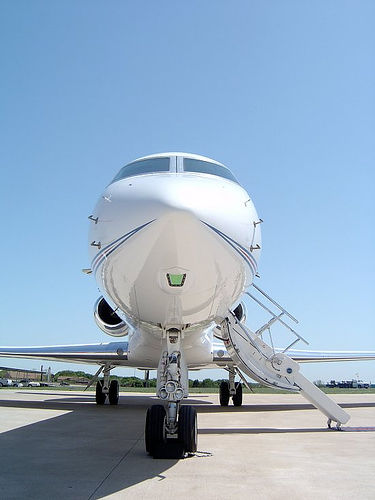}
    	\includegraphics[height=0.079\textheight]{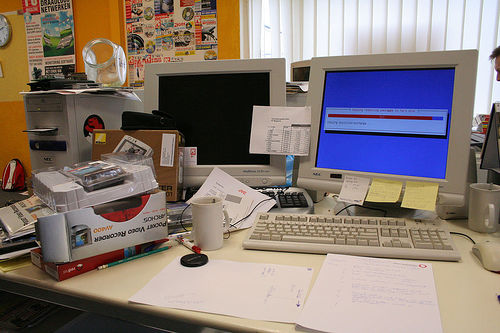}\\

		\includegraphics[height=0.079\textheight]{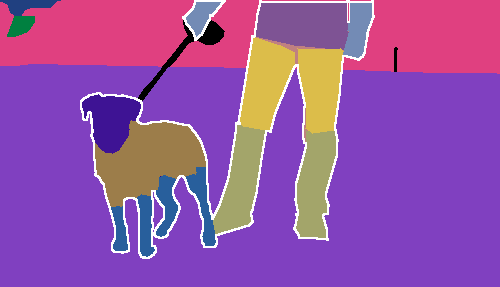}
		\includegraphics[height=0.079\textheight]{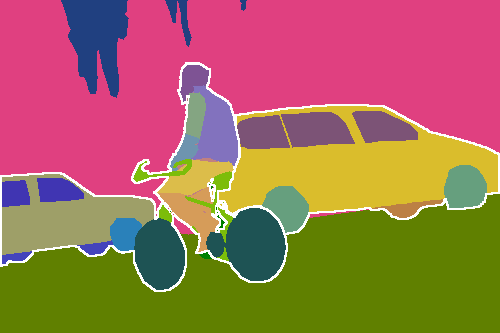}
		\includegraphics[height=0.079\textheight]{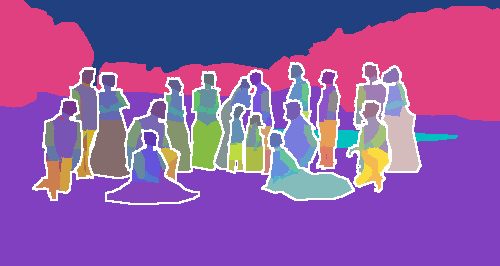}
		\includegraphics[height=0.079\textheight]{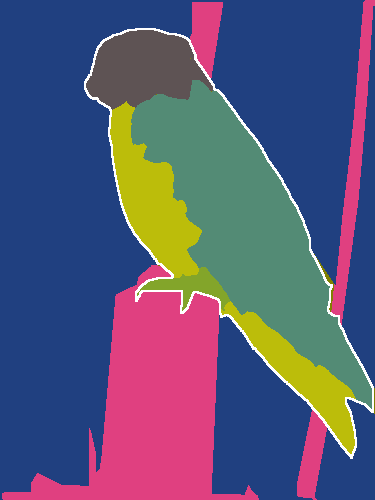}
		\includegraphics[height=0.079\textheight]{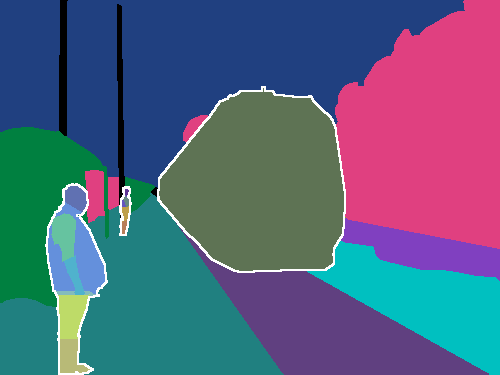}
		\includegraphics[height=0.079\textheight]{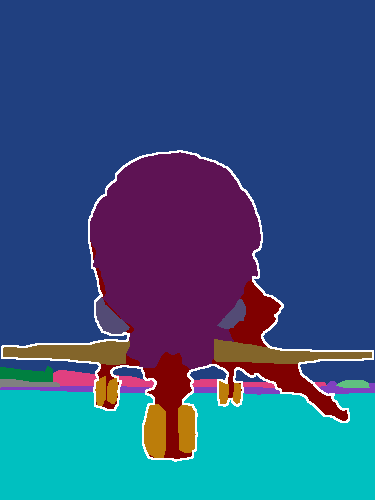}
		\includegraphics[height=0.079\textheight]{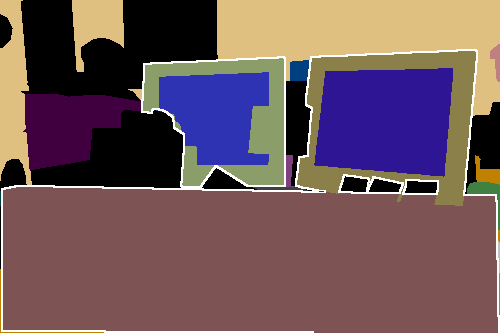}\\

		\includegraphics[height=0.079\textheight]{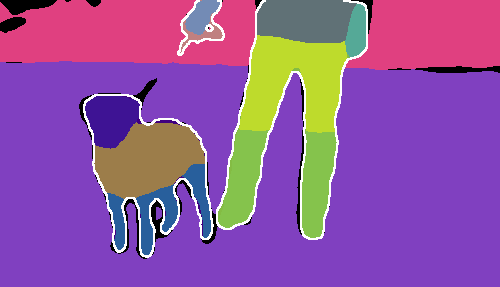}
		\includegraphics[height=0.079\textheight]{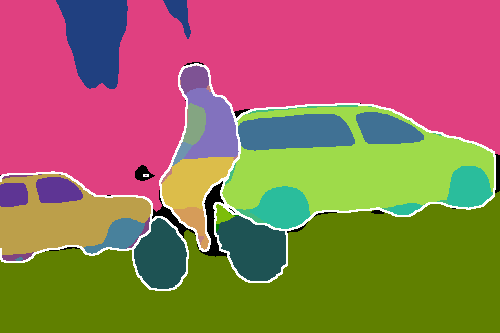}
		\includegraphics[height=0.079\textheight]{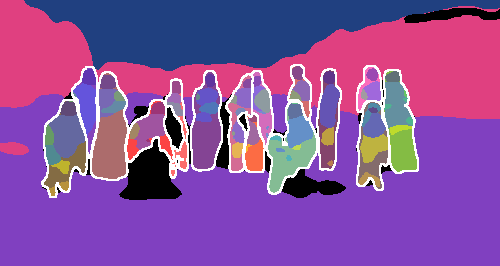}
		\includegraphics[height=0.079\textheight]{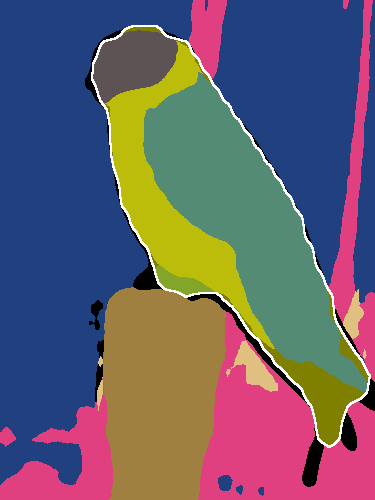}
		\includegraphics[height=0.079\textheight]{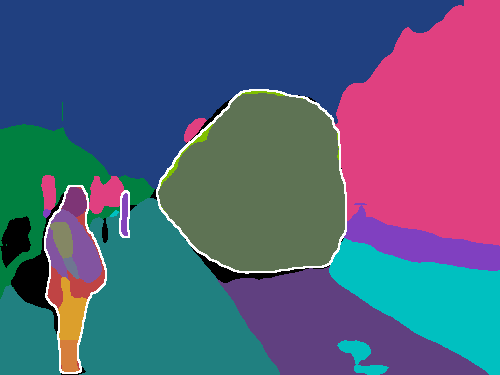}
		\includegraphics[height=0.079\textheight]{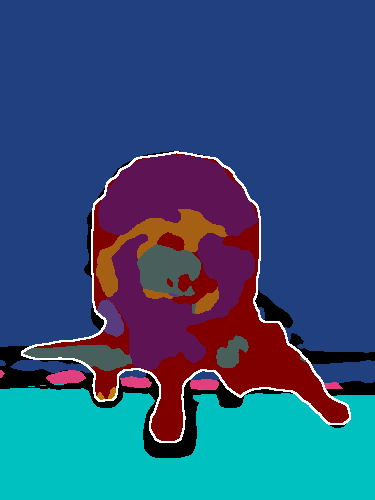}
		\includegraphics[height=0.079\textheight]{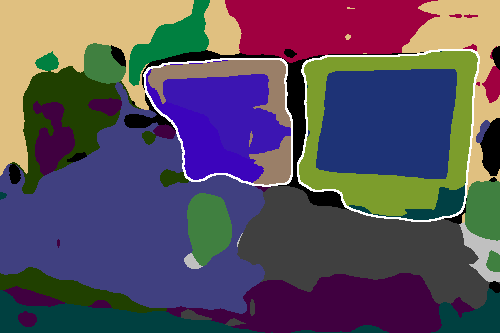}\\
		
		\vspace{10pt}
		
    	\includegraphics[height=0.084\textheight]{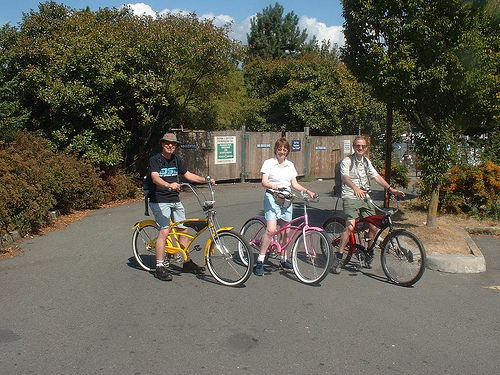}
    	\includegraphics[height=0.084\textheight]{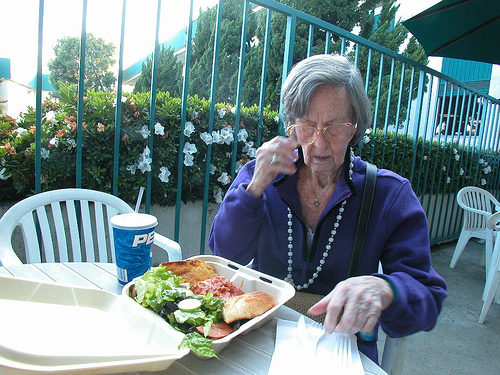}
    	\includegraphics[height=0.084\textheight]{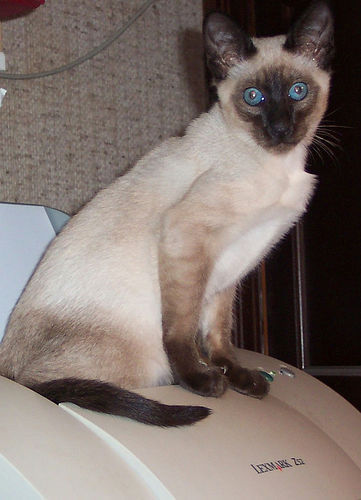}
    	\includegraphics[height=0.084\textheight]{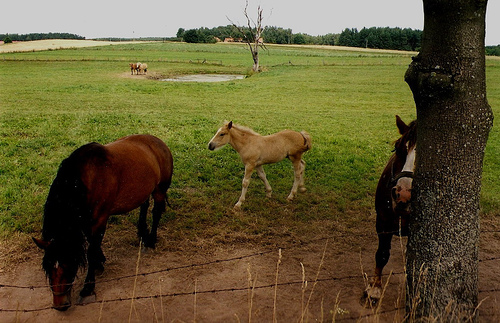}
    	\includegraphics[height=0.084\textheight]{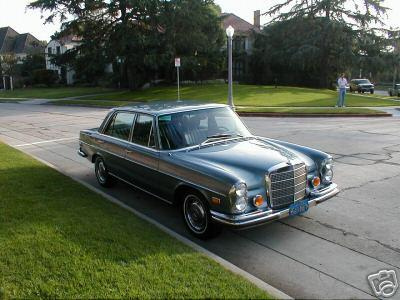}
    	\includegraphics[height=0.084\textheight]{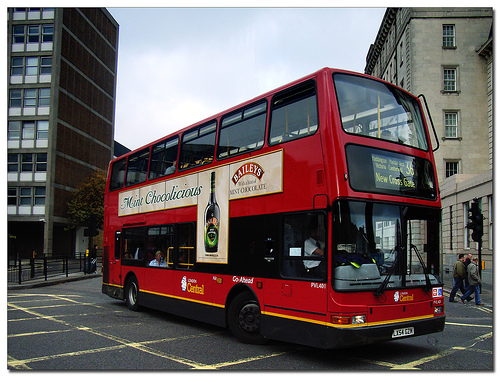}
    	\includegraphics[height=0.084\textheight]{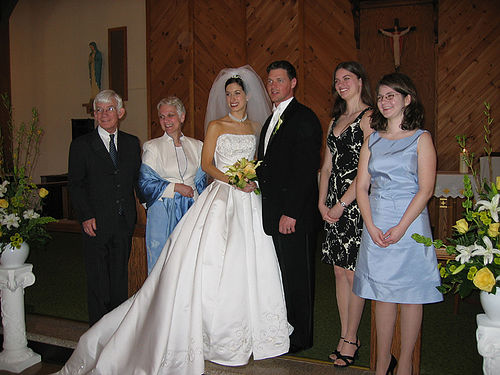}\\

		\includegraphics[height=0.084\textheight]{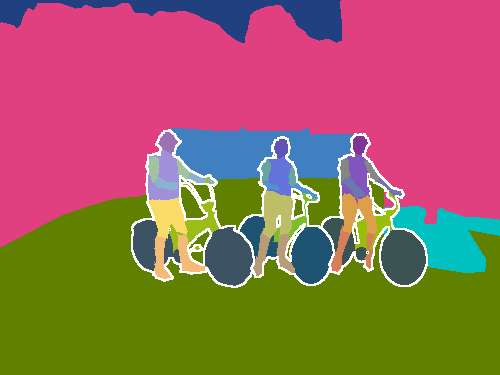}
		\includegraphics[height=0.084\textheight]{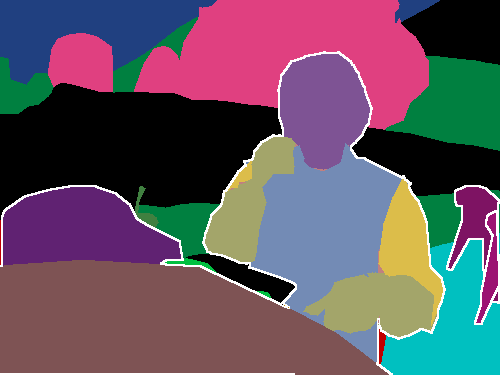}
		\includegraphics[height=0.084\textheight]{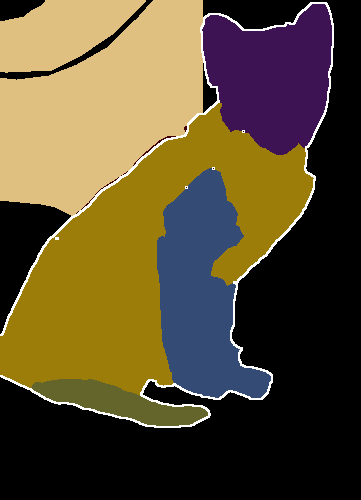}
		\includegraphics[height=0.084\textheight]{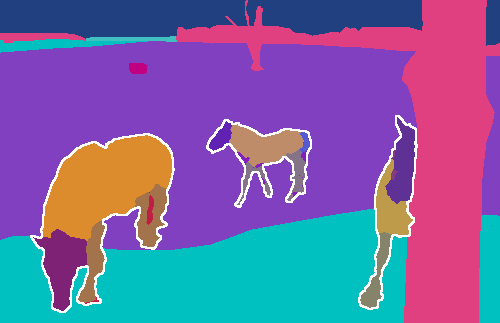}
		\includegraphics[height=0.084\textheight]{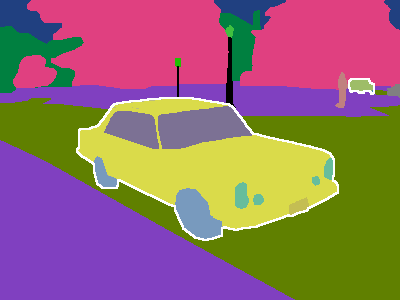}
		\includegraphics[height=0.084\textheight]{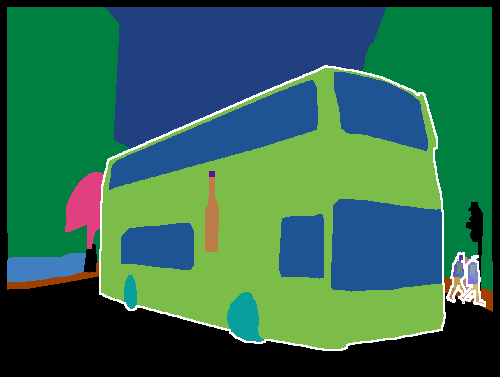}
		\includegraphics[height=0.084\textheight]{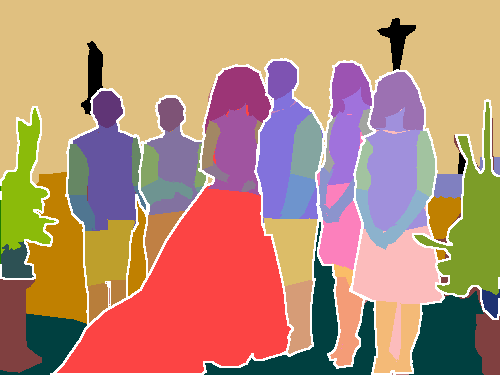}\\

		\includegraphics[height=0.084\textheight]{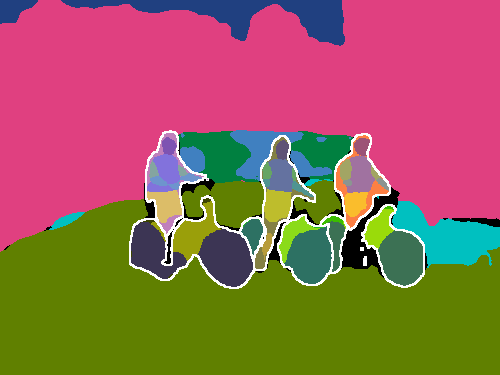}
		\includegraphics[height=0.084\textheight]{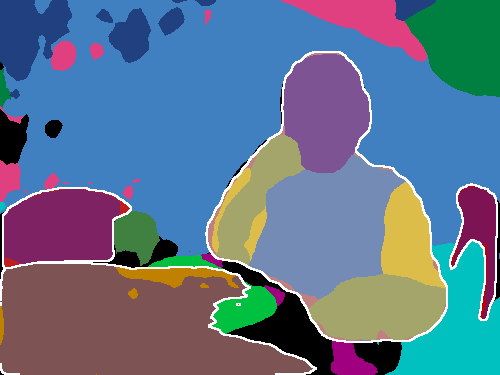}
		\includegraphics[height=0.084\textheight]{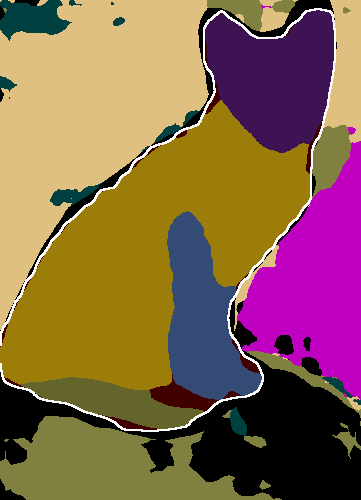}
		\includegraphics[height=0.084\textheight]{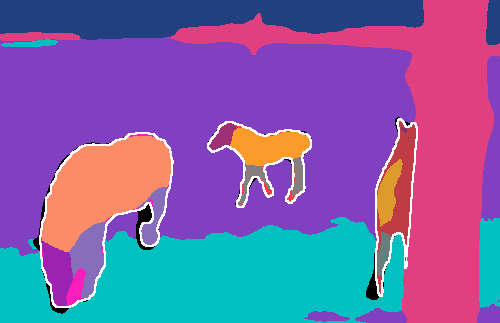}
		\includegraphics[height=0.084\textheight]{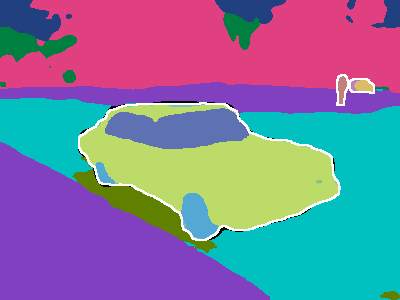}
		\includegraphics[height=0.084\textheight]{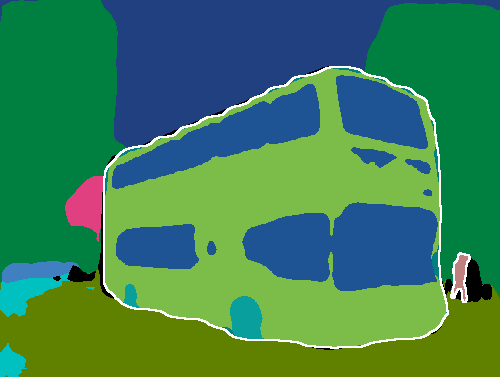}
		\includegraphics[height=0.084\textheight]{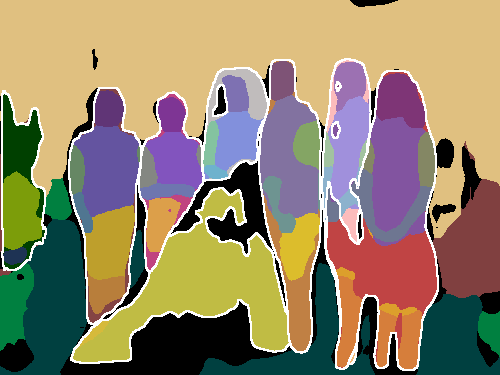}\\
		
		\vspace{10pt}
		
    	\includegraphics[height=0.085\textheight]{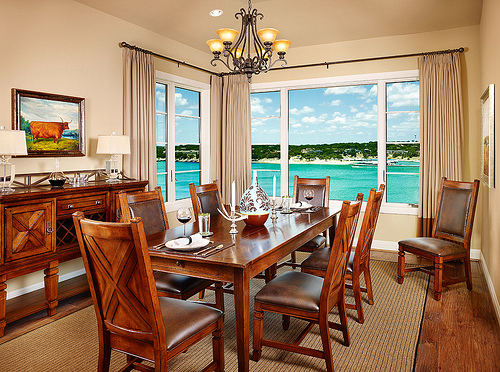}
    	\includegraphics[height=0.085\textheight]{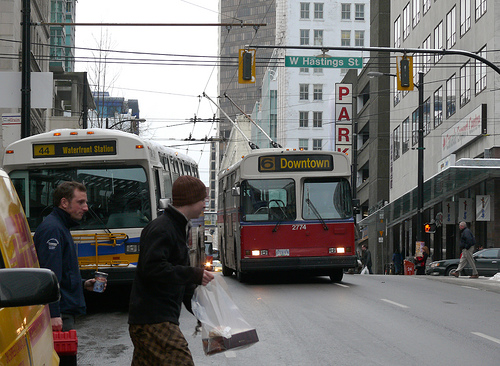}
    	\includegraphics[height=0.085\textheight]{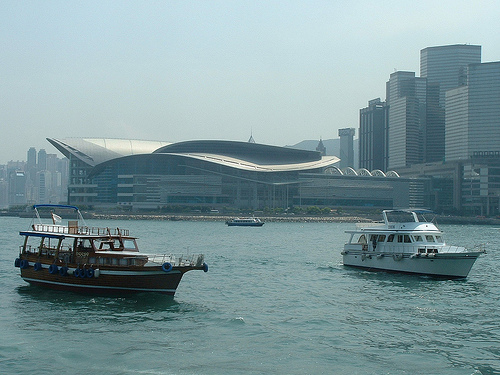}
    	\includegraphics[height=0.085\textheight]{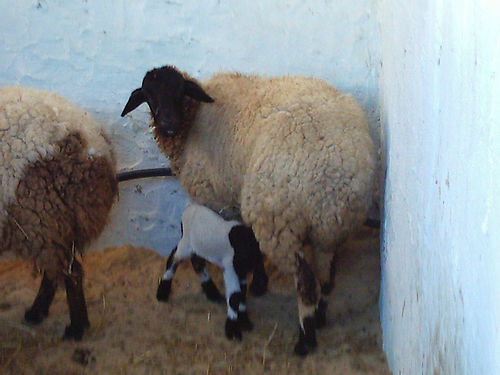}
    	\includegraphics[height=0.085\textheight]{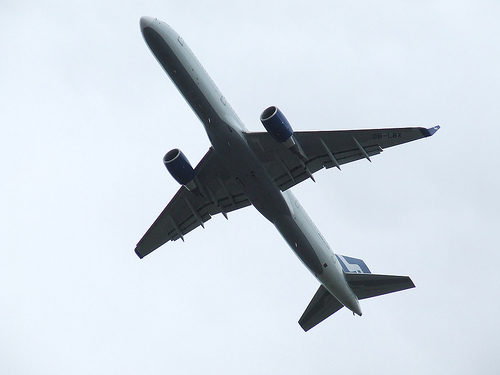}
    	\includegraphics[height=0.085\textheight]{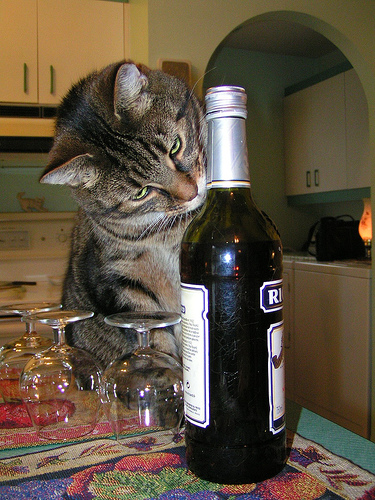}
    	\includegraphics[height=0.085\textheight]{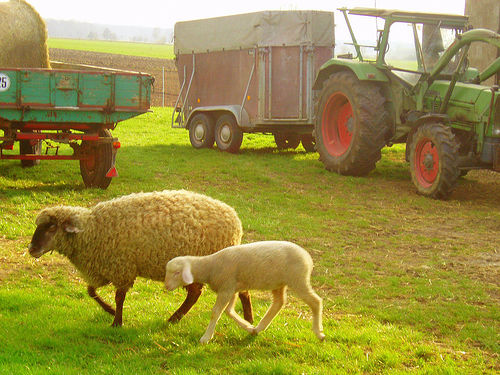}\\

		\includegraphics[height=0.085\textheight]{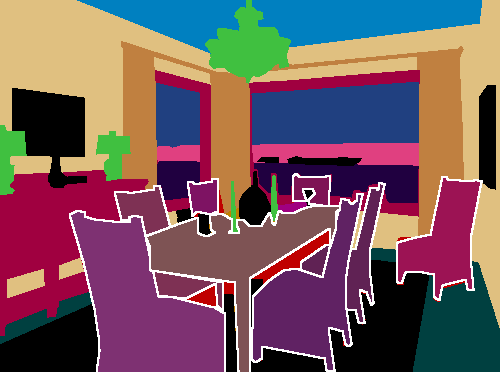}
		\includegraphics[height=0.085\textheight]{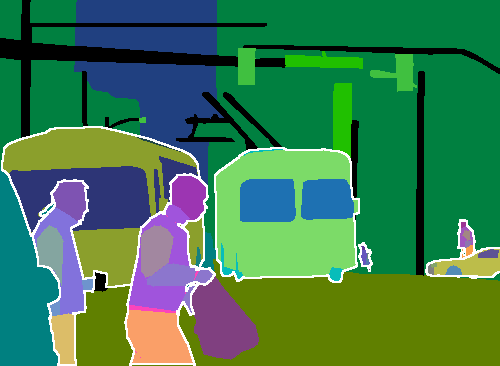}
		\includegraphics[height=0.085\textheight]{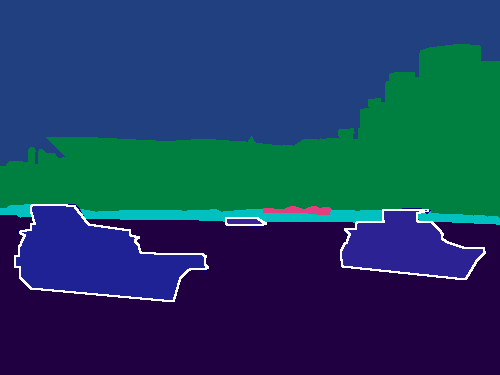}
		\includegraphics[height=0.085\textheight]{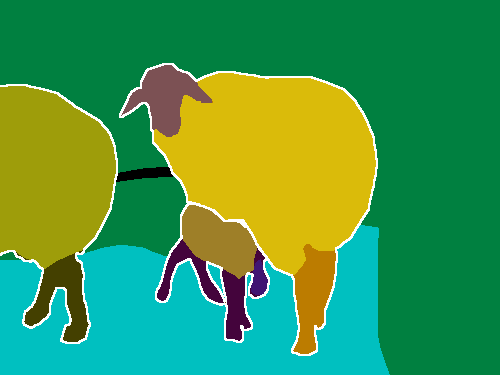}
		\includegraphics[height=0.085\textheight]{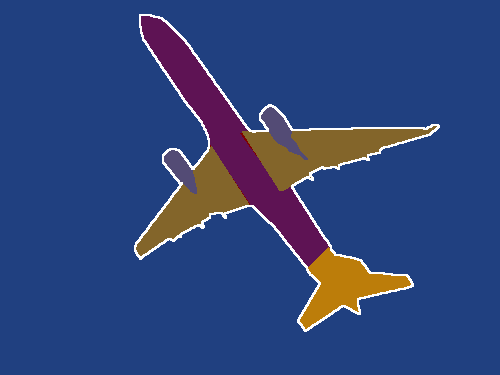}
		\includegraphics[height=0.085\textheight]{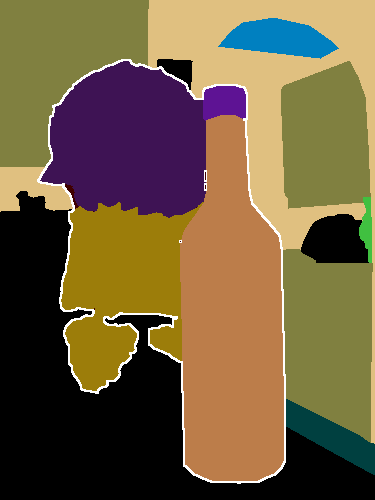}
		\includegraphics[height=0.085\textheight]{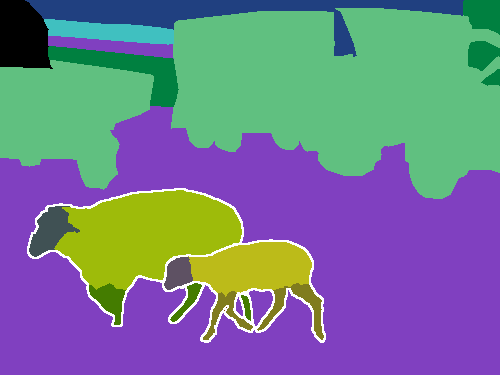}\\

		\includegraphics[height=0.085\textheight]{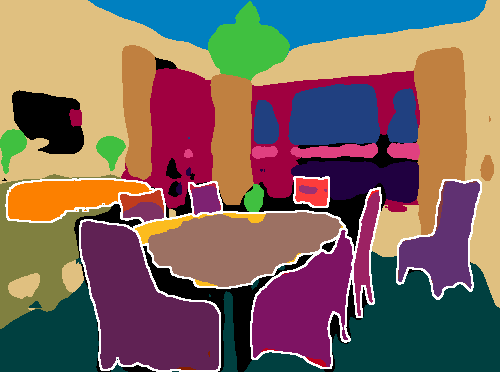}
		\includegraphics[height=0.085\textheight]{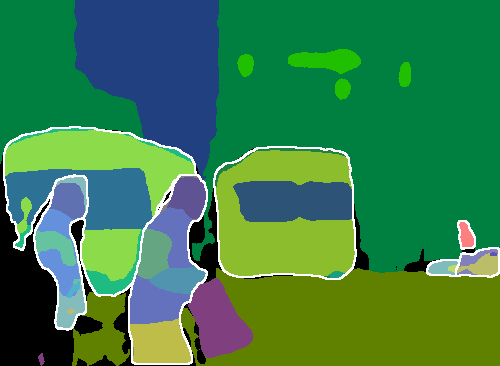}
		\includegraphics[height=0.085\textheight]{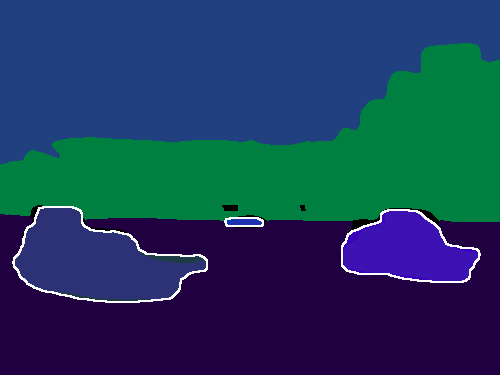}
		\includegraphics[height=0.085\textheight]{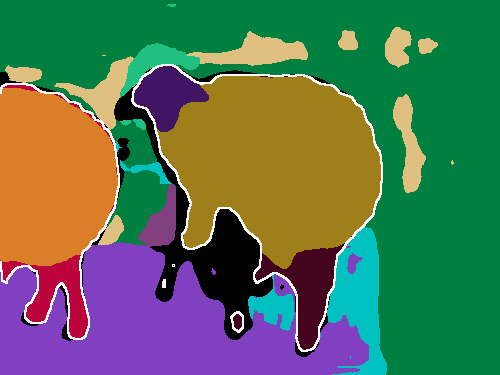}
		\includegraphics[height=0.085\textheight]{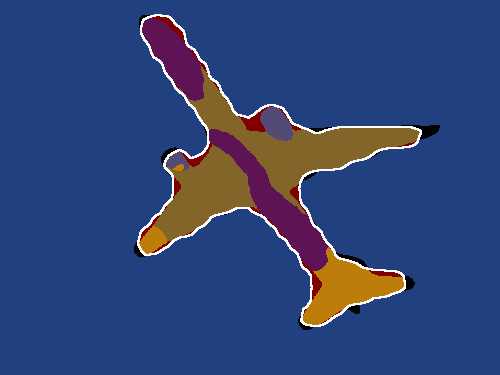}
		\includegraphics[height=0.085\textheight]{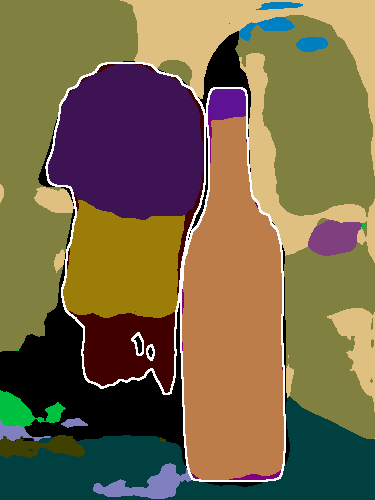}
		\includegraphics[height=0.085\textheight]{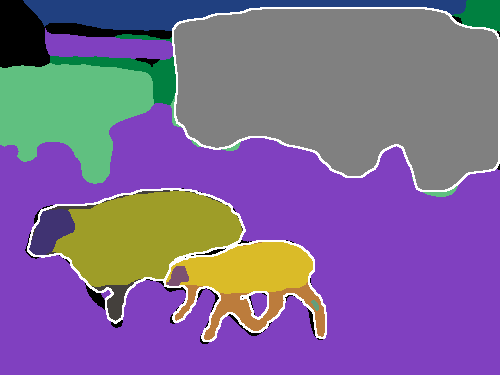}\\
	\caption{Qualitative examples for the highest-scoring part-aware panoptic segmentation baseline on PASCAL Panoptic Parts (DeepLabv3-ResNeSt269 \& DetectoRS \& BSANet~\cite{chen2017deeplabv3, zhang2020resnest,qiao2020detectors,Zhao2019BSANet}). On each of the three rows, we show the input images (top), ground truth (middle) and predictions (bottom).}
\label{fig:ppp_examples_supp}
\end{figure*}

We provide more qualitative examples of the highest scoring baselines on CPP and PPP in Figure \ref{fig:cpp_examples_supp} and \ref{fig:ppp_examples_supp}, respectively.

\end{document}